\documentclass{article}

\PassOptionsToPackage{numbers,compress}{natbib}

\usepackage[preprint]{neurips_2026}

\usepackage{wrapfig}
\usepackage{multicol}
\usepackage{subcaption}
\usepackage{booktabs}
\usepackage{appendix}
\usepackage{multirow}
\usepackage{amsmath}
\usepackage{amsthm}

\usepackage{graphicx} 
\graphicspath{{figs/}}
\usepackage[utf8]{inputenc} 
\usepackage[T1]{fontenc}    
\usepackage{hyperref}       
\usepackage{url}            
\usepackage{amsfonts}       
\usepackage{nicefrac}       
\usepackage{microtype}      
\usepackage{xcolor}         
\usepackage{enumitem}
\usepackage{bm}
\usepackage{bbm}
\usepackage{booktabs}
\usepackage{mathrsfs}  
\newcommand{\std}[1]{_{\pm\scriptscriptstyle #1}}

\usepackage{float}

\newcommand{\ind}{\perp\!\!\!\!\perp} 
\newcommand{\E}{\mathbb{E}}

\DeclareMathOperator*{\argmin}{arg\,min}
\newtheorem{theorem}{Theorem}
\newtheorem{lemma}[theorem]{Lemma}
\newtheorem{proposition}[theorem]{Proposition}

\newcommand{\Acal}{\mathcal A}
\newcommand{\Hcal}{\mathcal H}
\newcommand{\Fone}{\mathcal F_1}
\newcommand{\Ftwo}{\mathcal F_2}
\newcommand{\Relu}{\mathcal R}
\newcommand{\ReluB}{\mathcal R_B}

\newcommand{\X}{\mathcal X}

\newcommand{\R}{\mathbb{R}}
\newcommand{\N}{\mathcal{N}}
\newcommand{\cF}{\mathcal{F}}

\newcommand{\cH}{\mathcal{H}}
\newcommand{\cW}{\mathcal{W}}
\newcommand{\cA}{\mathcal{A}}
\newcommand{\cR}{\mathcal{R}}

\newcommand{\ol}{\overline}
\newcommand{\wh}{\widehat}

\newcommand{\eps}{\varepsilon}
\newcommand{\norm}[1]{\left\lVert #1 \right\rVert}
\newcommand{\abs}[1]{\left\lvert #1 \right\rvert}

\title{Estimating Heterogeneous Causal Effect on Networks via Orthogonal Learning} 
\author{%
  Yuanchen Wu \\
  Department of Statistics \\
  The Pennsylvania State University \\
  \texttt{yqw5734@psu.edu} \\
  \And
  Yubai Yuan \\
  Department of Statistics \\
  The Pennsylvania State University \\
  \texttt{yvy5509@psu.edu} \\
}
\begin{document}
\maketitle
\begin{abstract}

Estimating causal effects on networks is challenging because treatments may affect both treated units and their neighbors, while network homophily induces dependence and confounding. These challenges are amplified when causal effects are heterogeneous across units and edges. We propose a two-stage orthogonal learning framework for estimating heterogeneous direct and spillover effects on networks. The first stage uses graph neural networks to estimate nuisance components that capture complex dependence on covariates and network structure. The second stage residualizes these nuisance components and estimates causal effects through an interpretable attention-based interference model, yielding edge-level spillover estimates as well as node- and population-level summaries. Neyman orthogonalization and cross-fitting reduce sensitivity to first-stage estimation error, so nuisance errors enter only at higher order. We further develop a bootstrap-based uncertainty quantification procedure for the estimated spillover matrix, enabling pointwise and simultaneous inference for heterogeneous edge- and node-level effects. Experiments show that our method improves heterogeneous effect estimation while supporting interpretable downstream analyses such as influential-neighbor detection and spillover-sign recovery.
\end{abstract}

\section{Introduction}\label{Intro}

Causal inference on networks studies how one unit's treatment affects not only its own outcome but also the outcomes of others. This setting arises naturally in social media, epidemiology, and other applications with large-scale networked data and rich individual features. Unlike standard causal inference, network interference induces dependence across connected units, making methods based on independence assumptions generally inapplicable.

Two challenges are central. \textit{First}, causal effects under network interference are inherently heterogeneous: node features and pairwise connection strengths vary widely, so a unit's response to its neighbors depends on both individual traits and relational structure. Capturing this heterogeneity is essential for identifying causal effects and understanding diverse interaction patterns. \textit{Second}, network structure introduces complex confounding. Connected units often exhibit correlated behaviors due not only to causal spillovers but also to shared traits, latent dependencies, and homophily. Separating spillover effects from non-causal network associations therefore requires adjustment for high-dimensional features and latent relational structure.

\paragraph{Running example: political polarization.}
Social media campaigns use targeted ads (treatment) to influence voter turnout (outcome) \citep{bond2012experiment}. These campaigns generate both \textit{direct} effects, by encouraging ad recipients to vote, and \textit{spillover} effects as messages are reshared through social networks.

\begin{wrapfigure}{r}{0.53\textwidth}
\vspace{-15pt}
\centering
  \includegraphics[width=0.5\textwidth]{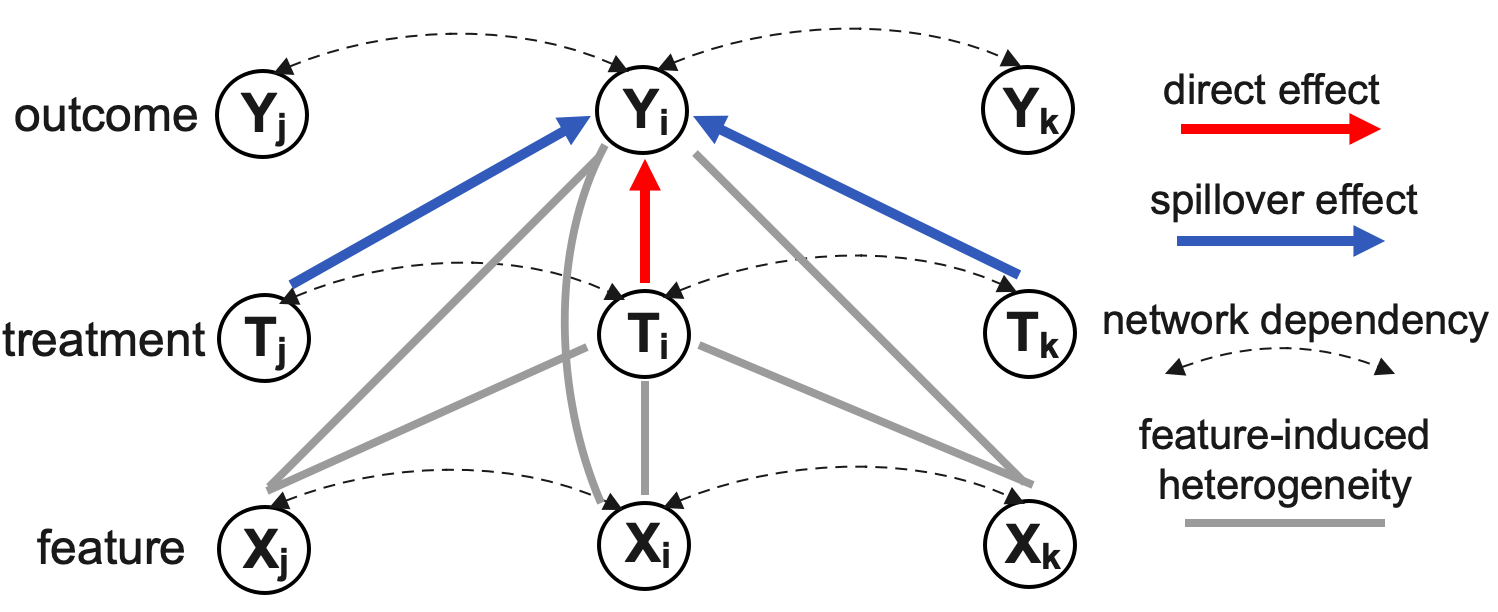} 
  \caption{Causal diagram for an ego unit $i$ on network where units $j$ and $k$ are two neighbors of $i$.}
  \label{fig:causal}
\vspace{-8pt}
\end{wrapfigure}

Spillover magnitudes may vary by ideological alignment and socioeconomic status, and their signs may differ across ideological positions, reflecting \textit{political polarization and echo chambers} \citep{bail2018exposure}. At the same time, \textit{network homophily} clusters voters with similar turnout patterns, so targeted ad exposure may overlap with naturally active groups. This complicates the separation of causal impacts from baseline voting behavior. Figure~\ref{fig:causal} illustrates the causal relations among outcome, treatment, and networked individual features.
This example motivates moving beyond average effects toward heterogeneous spillovers across individuals and their connections. Causal estimands under network interference can be organized by granularity: \emph{population-level}, \emph{node-level}, and \emph{edge-level} effects. Population-level estimands summarize average effects, whereas node- and edge-level estimands capture heterogeneity across units and connected pairs. Estimating effects at this finer resolution requires flexible modeling and weaker assumptions on the interference structure, making both estimation and \emph{uncertainty quantification} substantially more challenging.

Existing work has largely followed two directions. Classical interference methods, beginning with \citep{hudgens2008toward} and \citep{tchetgen2012causal} and further developed by \citep{exposure2017}, \citep{forastiere2021identification}, \citep{leung2020treatment}, \citep{leung2022causal}, \citep{ogburn2024causal}, and \citep{bargagli2025heterogeneous}, establish identification, estimation, and uncertainty quantification for population-level \emph{average causal effects}, typically under known exposure mappings. While foundational, these methods rely on structural assumptions that are often restrictive for \emph{heterogeneous spillover estimation} between connected units.

Recent \emph{graph neural network-based} approaches model interference more flexibly using high-dimensional covariates and complex network structure \citep{guo2020learning,ma2021causal,jiang2022estimating,cristali2022using,huang2023modeling,lin2023estimating,wu2025causal}. However, they typically encode spillovers through black-box latent representations, making pairwise \emph{edge-level} effects difficult to interpret and limiting valid \emph{uncertainty quantification} for heterogeneous causal estimands. Recent semiparametric and doubly robust network methods \citep{chen2024doubly,khatami2024graph} move toward more principled estimation and inference under weaker assumptions, but still primarily target lower-dimensional average network effects.

Motivated by this gap, our contribution is twofold. \textit{First}, we propose an orthogonal learning framework for heterogeneous causal effect estimation on networks, combining flexible graph-based nuisance modeling with an interpretable attention-based interference model for pairwise spillovers. This design allows direct and spillover effects to vary with sender features, receiver features, and local network structure, while remaining robust to first-stage nuisance estimation error. \textit{Second}, we develop a bootstrap-based procedure for uncertainty quantification of the estimated spillover structure, extending inference beyond population-level summaries to heterogeneous pairwise edge-level spillover effects.
\section{Problem setup}\label{Section2}
We work within the potential outcomes framework for causal inference under interference. We observe a single network with unit set \(\bm{V}=\{1,\ldots,n\}\), adjacency matrix \(\bm{A}\in\{0,1\}^{n\times n}\), unit-level covariates \(\bm{X}=(X_1^\top,\ldots,X_n^\top)^\top \in \mathbb{R}^{n\times p}\), treatment assignments \(\bm{T}=(T_1,\ldots,T_n)\), and observed outcomes \(\bm{Y}=(Y_1,\ldots,Y_n)\), where each \(X_i\in\mathcal X\) is a \(p\)-dimensional feature vector. Equivalently, the observed data are \((X_i,T_i,Y_i)_{i=1}^n\) together with the network structure \(\bm{A}\).

Let \(\bm{t}=(t_1,\ldots,t_n)\in\{0,1\}^n\) denote a generic treatment assignment vector, where \(t_i\) is the treatment assigned to unit \(i\). Under interference, the potential outcome of unit \(i\) is denoted by \(Y_i(\bm{t})\), allowing the outcome of unit \(i\) to depend on the entire treatment assignment vector rather than only on its own treatment. We write \(\bm{t}_{-i}\) for the treatment assignments of all units other than \(i\).

\paragraph{Neighborhood notation.} Let $\mathrm{dist}(i,j)$ denote the shortest-path distance between units $i$ and $j$ in $\bm A$. For $K\ge 0$, define
$\mathcal N_K(i):=\{j\in \bm V: 1\le \mathrm{dist}(i,j)\le K\}$ and $\bar{\mathcal N}_K(i):=\mathcal N_K(i)\cup\{i\}$. Let $d_K(i):=|\mathcal N_K(i)|$ and $\bar d_K(i):=|\bar{\mathcal N}_K(i)|$. For any node set $S\subseteq \bm V$, write $\bm t_S:=(t_j)_{j\in S}$ and $\bm X_S:=(X_j)_{j\in S}$, and define $\bar{\mathcal N}_K(S):=\bigcup_{i\in S}\bar{\mathcal N}_K(i)$.

\paragraph{Target estimands.}
We define the individual direct, spillover, and total effects as $\mathrm{IDE}_i := \E[Y_i(t_i=1,\bm t_{-i}=\bm 0)-Y_i(t_i=0,\bm t_{-i}=\bm 0)\mid \bm X,\bm A]$, $\mathrm{ISE}_i := \E[Y_i(t_i=0,\bm t_{-i}=\bm 1)-Y_i(t_i=0,\bm t_{-i}=\bm 0)\mid \bm X,\bm A]$, and $\mathrm{ITE}_i := \E[Y_i(t_i=1,\bm t_{-i}=\bm 1)-Y_i(t_i=0,\bm t_{-i}=\bm 0)\mid \bm X,\bm A]$. These are node-level causal quantities that depend on each unit's features and position in the network. Their population-level averages are $\mathrm{ADE}:=\frac{1}{n}\sum_{i=1}^n \mathrm{IDE}_i$, $\mathrm{ASE}:=\frac{1}{n}\sum_{i=1}^n \mathrm{ISE}_i$, and $\mathrm{ATE}:=\frac{1}{n}\sum_{i=1}^n \mathrm{ITE}_i$.
\section{Identification under local additive interference}

To identify the node-level and population-level estimands in Section~2 from a single observed network, we first introduce two identification assumptions. 


\noindent\textbf{Assumption 1 (Local interference).}
For every unit $i$ and treatment assignment vector $\bm t$, $\E\!\left[Y_i(\bm t)\mid \bm X,\bm A\right]
=
\E\!\left[Y_i(\bm t_{\bar{\mathcal N}_1(i)})\mid \bm X_{\bar{\mathcal N}_K(i)},\bm A\right].$

\noindent\textbf{Assumption 2 (Neighborhood unconfoundedness and overlap).}
For every unit $i$ and every local treatment vector $\bm s \in \{0,1\}^{\bar d_1(i)}$, $Y_i(\bm s) \ind \bm T_{\bar{\mathcal N}_1(i)} \mid \bm X_{\bar{\mathcal N}_K(i)}, \bm A,
\qquad
0 < \Pr\!\big(\bm T_{\bar{\mathcal N}_1(i)}=\bm s \mid \bm X_{\bar{\mathcal N}_K(i)}, \bm A\big)<1.$

Assumptions~1 and 2 separate the scope of causal interference from the scope of adjustment:
outcomes depend on treatments within the closed one-hop neighborhood, while
confounding adjustment may use covariates from a larger $K$-hop neighborhood.
Together, they identify the node- and population-level estimands from the observed network.

\noindent\textbf{Assumption 3 (Additive local interference).}
For each unit $i$, there exist functions $\{g_{ij}(t_j,\bm X_{\bar{\mathcal N}_K(i)},\bm A)\}_{j\in \bar{\mathcal N}_1(i)}$ such that $\E\!\left[Y_i(\bm t_{\bar{\mathcal N}_1(i)}) \mid \bm X_{\bar{\mathcal N}_K(i)},\bm A\right]
=
\sum_{j\in\bar{\mathcal N}_1(i)} g_{ij}\!\left(t_j,\bm X_{\bar{\mathcal N}_K(i)},\bm A\right).$

Assumption~3 is the structural restriction specific to our framework, allowing us to introduce \textit{edge-level} pairwise spillover components. It imposes a first-order additive structure: the ego unit contributes a direct-effect component, while each neighbor contributes a context-dependent spillover component. This yields an interpretable edge-level decomposition of the node- and population-level estimands. 

We refer to Appendix~\ref{dicuss_assumption3} for a detailed discussion of this assumption, including its role as a first-order approximation to more general interference functions, its tradeoff between flexibility and identifiability, and possible extensions to higher-order treatment interactions.


\begin{theorem}
For unit $i$, define
\begingroup
\small
\[
\tau_i :=
g_{ii}\!\left(1,\bm X_{\bar{\mathcal N}_K(i)},\bm A\right)
-
g_{ii}\!\left(0,\bm X_{\bar{\mathcal N}_K(i)},\bm A\right),\;
\tau_{ij}
:=
g_{ij}\!\left(1,\bm X_{\bar{\mathcal N}_K(i)},\bm A\right)
-
g_{ij}\!\left(0,\bm X_{\bar{\mathcal N}_K(i)},\bm A\right),
\; j\in\mathcal N_1(i).
\]
\endgroup

Under Assumptions~1--3, the unit-level total, direct, and spillover effects are identifiable from the observed law of $\bigl(Y_i,\bm T_{\bar{\mathcal N}_1(i)},\bm X_{\bar{\mathcal N}_K(i)},\bm A\bigr).$ In particular,
\begin{align}
\label{def_1}
\mathrm{ITE}_i = \tau_i + \sum_{j\in\mathcal N_1(i)} \tau_{ij},
\qquad
\mathrm{IDE}_i = \tau_i,
\qquad
\mathrm{ISE}_i = \sum_{j\in\mathcal N_1(i)} \tau_{ij}.
\end{align}
Furthermore, for any two treatment assignments $\bm t$ and $\bm t'$,
\[
\E\!\left[Y_i(\bm t)-Y_i(\bm t') \mid \bm X,\bm A\right]
=
\tau_i(t_i-t_i') +  \sum_{j\in\mathcal N_1(i)} \tau_{ij}(t_j-t_j').
\]
\end{theorem}
\section{Orthogonal estimation of heterogeneous spillovers}\label{sec:method}
Theorem~1 motivates estimating an augmented effect matrix
$\Gamma\in\mathbb R^{n\times n}$, with $\Gamma_{ii}:=\tau_i$,
$\Gamma_{ij}:=\tau_{ij}$ for $j\in\mathcal N_1(i)$, and $\Gamma_{ij}:=0$
otherwise. The node- and population-level estimands are then linear functionals
of $\Gamma$, so we reduce causal effect estimation to estimating $\Gamma$. In this section, we develop an orthogonal two-stage estimator for $\bm{\Gamma}$ on a single dependent network.

\begin{figure}[t!]
  \centering
  \includegraphics[width=\textwidth]{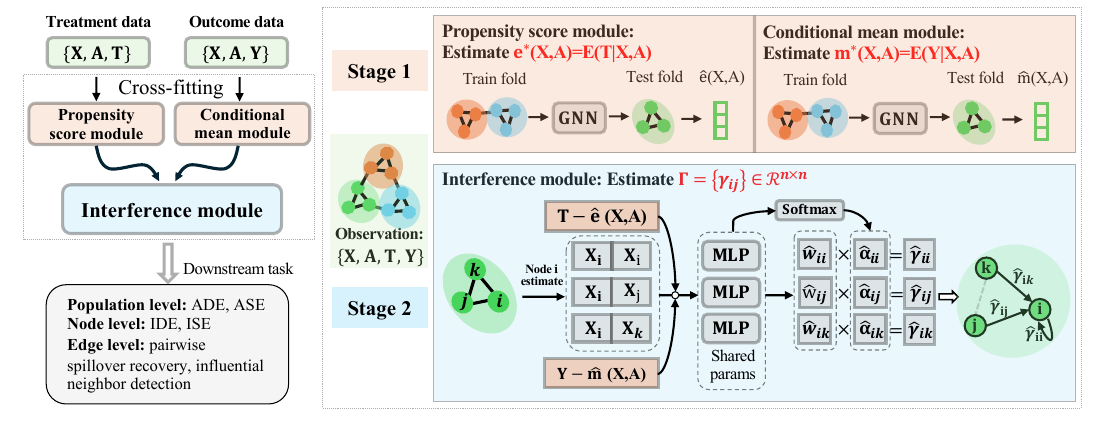} 
  \caption{The proposed two-stage orthogonal learning framework for estimating direct and spillover effects.}
  \label{fig:workflow}
\end{figure}

\subsection{Orthogonal moment formulation}



Under Assumptions~1--3, the outcome model admits the additive representation
\begin{align}
\label{eq_1}
Y_i
=
\mu_i^0
+
T_i\tau_i
+
\sum_{j\in \mathcal{N}_1(i)} T_j\tau_{ij}
+
\epsilon_i,
\qquad
\mu_i^0
:=
\E\!\left[
Y_i(\bm 0_{\bar{\mathcal N}_1(i)})
\mid
\bm X_{\bar{\mathcal N}_K(i)},\bm A
\right],
\end{align}
where
$\E(\epsilon_i\mid \bm T_{\bar{\mathcal N}_1(i)},\bm X_{\bar{\mathcal N}_K(i)},\bm A)=0$. To derive the orthogonal learning formulation for estimating \(\bm{\Gamma}\), we first take the expectation of both sides of equation~(\ref{eq_1}) with respect to \((\bm{T}, \bm{Y})\), conditional on \(\bm{X}\) and \(\bm{A}\), and then subtract the conditional outcome from both sides of (\ref{eq_1}): 
{\small
\begin{equation}\label{eq_3}
Y_i - \bm{E}(Y_i \mid \bm X,\bm A)
=
\big(T_i - \bm{P}(T_i =1 \mid \bm X,\bm A)\big)\tau_i
+
\sum_{j\in \mathcal N_1(i)}
\big(T_j - \bm{P}(T_j =1 \mid \bm X,\bm A)\big)\tau_{ij}
+\epsilon_i .
\end{equation}
}
The orthogonal learning formulation in equation~(\ref{eq_3}) separates the estimation of nuisance components—specifically, the conditional outcome mean \(m := \mathbb{E}(Y \mid \bm{X}, \bm{A})\) and the propensity score \(e := \mathbb{P}(T = 1 \mid \bm{X}, \bm{A})\)—from the estimation of \(\bm{\Gamma}\). These nuisance functions are relevant for identifying the target causal estimands, but errors in estimating \(m\) and \(e\) can introduce bias when estimating \(\bm{\Gamma}\). The orthogonal loss design in (\ref{eq_3}) generalizes the R-Learner framework \citep{nie2021quasi} and enables estimation of \(\bm{\Gamma}\) with generalization error comparable to that of an oracle estimator with known \(m\) and \(e\). Given that \(m\) and \(e\) may have complex dependencies on \(\bm{X}\) and \(\bm{A}\), we propose using expressive graph neural networks to estimate \(\hat{m}\) and \(\hat{e}\), which are then plugged into equation~(\ref{eq_3}) to estimate \(\bm{\Gamma}\). An overview of the proposed framework for network interference is illustrated in Figure~\ref{fig:workflow}.


Orthogonal learning also relies on a cross-fitting procedure, where nuisance and target components are estimated on two independent subsets of the data to avoid overfitting bias. Unlike the i.i.d. setting, observations \(\{T_i, Y_i\}_{i=1}^n\) are typically dependent due to the network structure. This necessitates a refined cross-fitting strategy along with additional assumptions.


\textbf{Assumption 4.}
For any two node subsets $\mathcal S_1,\mathcal S_2\subseteq V$ satisfying
$\mathrm{dist}(i,j)>2K$ for all $i\in\mathcal S_1$ and $j\in\mathcal S_2$,
\textup{(1)}
$(\bm Y_{\mathcal S_1},\bm T_{\bar{\mathcal N}_1(\mathcal S_1)})
\ind
(\bm Y_{\mathcal S_2},\bm T_{\bar{\mathcal N}_1(\mathcal S_2)})
\mid
\bm X_{\bar{\mathcal N}_K(\mathcal S_1)\cup\bar{\mathcal N}_K(\mathcal S_2)},\bm A$;
\textup{(2)}
$\bm X_{\bar{\mathcal N}_K(\mathcal S_1)}
\ind
\bm X_{\bar{\mathcal N}_K(\mathcal S_2)}\mid\bm A$;
and \textup{(3)}
$\E[\epsilon_i\mid \bm T_{\bar{\mathcal N}_1(i)},\bm X_{\bar{\mathcal N}_K(i)},\bm A]=0$
for all $i\in V$.

\noindent Assumption 4 imposes constraints on the scope of network dependence, enabling estimation from a single snapshot of the network. Similar assumptions have been used in \citep{ogburn2024causal, leung2020treatment} to ensure consistent ATE estimation. In the following, we introduce the cross-fitting procedure under network dependence.




\underline{\textit{Data Splitting}}:
Randomly select a subset of units $\bm V_1\subset\bm V$, and define
$
\bm V_2
:=
\{j\in\bm V:\mathrm{dist}(j,i)>2K \text{ for all } i\in\bm V_1\}.
$
For $s\in\{1,2\}$, let $\bm V_{-1}:=\bm V_2$ and $\bm V_{-2}:=\bm V_1$, and define
$
\widetilde{\bm V}_s
:=
\{j\in\bar{\mathcal N}_1(i):i\in\bm V_s\}.$ Denote $\tilde n:=|\bm V_1|+|\bm V_2|$.

\underline{\textit{Stage 1}}:
For each held-out fold $s\in\{1,2\}$, train and tune GNN-based nuisance
estimators on the opposite closed-neighborhood fold $\widetilde{\bm V}_{-s}$.
Specifically, fit the outcome nuisance $\hat m^{(-s)}$ using
$(\{Y_j\}_{j\in\widetilde{\bm V}_{-s}},\bm X,\bm A)$ and the propensity
nuisance $\hat e^{(-s)}$ using
$(\{T_j\}_{j\in\widetilde{\bm V}_{-s}},\bm X,\bm A)$.


\underline{\textit{Stage 2}}:
For each unit $i\in\bm V_1\cup\bm V_2$, let $s_i$ denote the fold containing
$i$. Estimate the interference coefficients by solving
{\small
\begin{align}\label{sp_est}
\hat\theta
=
\arg\min_{\theta\in\Theta}
\sum_{i\in \bm V_1\cup \bm V_2}
\left(
Y_i-\hat m_i^{(-s_i)}
-(T_i-\hat e_i^{(-s_i)})\tau_i(\theta)
-\sum_{j\in\mathcal N_1(i)}(T_j-\hat e_j^{(-s_i)})\tau_{ij}(\theta)
\right)^2,
\;
\hat\Gamma=\Gamma_{\hat\theta}.
\end{align}
}


\subsection{Attention-based network interference modeling}

The interference matrix $\bm{\Gamma}$ should capture heterogeneous and asymmetric
spillover effects across neighboring units. We parameterize $\bm{\Gamma}$ using an
attention-based interference model inspired by graph attention
\citep{velivckovic2017graph}. For each unit $i\in\bm V$, we model the direct
effect as $\tau_i=\tilde W(X_i)$, where
$\tilde W=\text{MLP-ReLU}:\mathbb R^p\to\mathbb R$. For each neighbor
$j\in\mathcal N_1(i)$, we define
\begin{align}\label{GAT}
    \tau_{ij}
    =
    \alpha_{ij} w_{ij},
    \;
    w_{ij}:=W(X_i,X_j),
    \;
    \alpha_{ij}:=\bm\alpha(w_{ij},\bm w_i),
    \;
    \alpha_{ij}\ge 0,\quad
    \sum_{j\in\mathcal N_1(i)}\alpha_{ij}=1 .
\end{align}
Here $\bm w_i$ collects $\{w_{ij}:j\in\mathcal N_1(i)\}$, and
$W=\text{MLP-ReLU}:\mathbb R^{2p}\to\mathbb R$ is applied to the concatenated
features of $(X_i,X_j)$. We instantiate the attention weights by
$\alpha_{ij}=(\mathrm{softmax}(\beta|\bm w_i|))_j$, where $|\beta|$ is a bounded
learnable temperature parameter. This parameterization captures both the sign
and magnitude of heterogeneous spillovers, allows asymmetric influence
$(\tau_{ij}\neq\tau_{ji})$, and interpolates between diffuse and concentrated
neighborhood influence as $\beta$ varies. It also generalizes several exposure
mappings used in prior work~\citep{aronow2017estimating}.

\subsection{Theoretical analysis}
According to the interference model, causal effect estimation is primarily determined by the influence function \(W\). Therefore, we analyze the convergence of the induced effect estimates $\hat\tau_i$ and $\hat\tau_{ij}$ obtained from the attention-based interference model.

Assume that width, depth, and weight magnitude of the ReLU neural networks   $\tilde{W}$ and $W$ in section 4 are bounded by $\mathcal{W},\mathcal{L},\Lambda$, and $\max_{i,j}\{|\tau_i|, |\tau_{ij}|\}\leq B$.

\textbf{Assumption 5.}
With probability at least $1-\delta$, the estimation of the conditional mean and propensity score satisfies the following property:
{\small
\begin{align}\label{nuis}
 \left(\frac1n\sum_{i=1}^n
  \mathbb{E}(\hat{m}_i-m_i^\star)^4 \right)^{1/4} = \mathcal{O}(r_{m}) ;\;\;\;
 \left(\frac{1}{\sum_{i=1}^n d_1(i)}\sum_{i=1}^n
  \sum_{j\in \mathcal{N}_1(i)}
  \mathbb{E}(\hat{e}_{j}-e_{j}^\star)^4\right)^{1/4} = \mathcal{O}(r_{e}) .
\end{align}
}

Assumption 5 requires that the prediction errors of the nuisance estimators can be controlled. In our work, we use graph neural networks as the estimators, whose generalization properties have been studied in \citep{garg2020generalization, tang2023towards, vasileiou2025survey}. The \(L_4\) norm condition can be relaxed to an \(L_2\) norm under the assumption that \(\bm{X}\) follows a sub-Gaussian distribution.

\textbf{Assumption 6.} Consider the loss function
$l(\tau_{i\cdot};m,e)=$$\left(Y_i-m_i-(T_i-e_i)\tau_i-\sum_{j\in \mathcal N_1(i)}(T_j-e_j)\tau_{ij}\right)^2$.
There exists $L_l>0$ such that, for any $\tau_{i\cdot}$ and $\tau'_{i\cdot}$,
$|l(\tau_{i\cdot};m,e)-l(\tau'_{i\cdot};m,e)|\le L_l\|\tau_{i\cdot}-\tau'_{i\cdot}\|_\infty$ and $l(\tau_{i\cdot};m,e)\le M$ for all $i\in V$.

Assumption~6 is a regularity condition requiring the loss function to be Lipschitz continuous and uniformly bounded.

\begin{theorem}
    Under Assumption 4, 5, and 6, and assuming for any distinct $j,k \in \bar{{\mathcal{N}}}_1(i)$ there exists constant $0\leq\rho < 1$ such that
$|\text{corr}(T_j, T_k \mid \bm{X}_{\bar{\mathcal{N}}_K(i)})| \leq \rho$. Denote $\{\tau^{\star}_i, \tau^{\star}_{ij} \}$ as true interference coefficients. With probability at least $1-\delta$, we have
\begin{align*}
 \frac1n\sum_{i=1}^n \| \hat{\tau}_{i\cdot} -  {\tau}^{\star}_{i\cdot} \|^2_2
   =  \mathcal{O}\Big\{  \frac{d_{K,n}M\mathcal{W}\mathcal{L}^{1/2} }{\delta\sqrt{\tilde{n}(1-\rho)}}\Delta
  + \frac{B^2 (1+d_{1,n})^2}{(1-\rho)^{3/2}}(r_{m}^4+r_{e}^4) \Big\},
\end{align*}
where $\| \hat{\tau}_{i\cdot} -  {\tau}^{\star}_{i\cdot} \|^2_2 = (\hat{\tau}_{i} -  {\tau}^{\star}_{i})^2+\sum_{j\in \mathcal{N}_1(i)}(\hat{\tau}_{ij} -  {\tau}^{\star}_{ij})^2$, and $\Delta := \sqrt{ \log\left(
  \frac{\bar\beta B L_l \max\{B,(\mathcal{W}\Lambda)^{\mathcal{L}} \}}{M}
  \right)
  }$, and $d_{K,n}: = \max_{i\in V}d_K(i)$.
\end{theorem}

Theorem~2 decomposes the estimation error of $\hat{\tau}$ into a second-stage term of order $\mathcal O(\tilde n^{-1/2})$ and a nuisance-induced term of order $\mathcal O(r_m^4+r_e^4)$, up to network-degree, dependence, and model-complexity factors. Thus, nuisance errors enter only at higher order: if $r_m,r_e=\mathcal O(\tilde n^{-1/8})$, then $r_m^4+r_e^4=\mathcal O(\tilde n^{-1/2})$, and $\hat{\tau}$ attains the same rate as the oracle estimator using the true nuisance functions $m$ and $e$.

\section{Uncertainty quantification}
\label{sec:uq}

Prior work on causal inference under network interference has primarily focused on population-level inference. In this section, we adapt multiplier-bootstrap ideas from the empirical process literature
\citep{chernozhukov2013gaussian,chernozhukov2014gaussian}
to quantify uncertainty for edge- and node-level causal estimands.

Theorem~2 implies that, after orthogonalization and cross-fitting, first-stage
nuisance error enters the second-stage estimator only at higher order. We
therefore condition on the cross-fitted nuisance estimates used in
\eqref{sp_est} and quantify uncertainty by perturbing the second-stage loss.
Let $\ell_i(\theta)$ denote the squared loss contribution of unit $i$ in
\eqref{sp_est}, and define
$L_{\tilde n}(\theta)=\tilde n^{-1}\sum_{i\in\bm V_1\cup\bm V_2}\ell_i(\theta)$,
with $\hat\theta\in\arg\min_{\theta\in\Theta}L_{\tilde n}(\theta)$ and
$\hat\Gamma=\Gamma_{\hat\theta}$.

For $b=1,\ldots,B$, draw iid multipliers
$\{\xi_i^{(b)}:i\in\bm V_1\cup\bm V_2\}$ with
$\E[\xi_i^{(b)}]=1$ and $\mathrm{Var}(\xi_i^{(b)})=1$, and refit the
second-stage model using
\[
L_{\tilde n}^{(b)}(\theta)
=
\frac{1}{\tilde n}
\sum_{i\in\bm V_1\cup\bm V_2}\xi_i^{(b)}\ell_i(\theta),
\qquad
\hat\theta^{(b)}
\in
\arg\min_{\theta\in\Theta}L_{\tilde n}^{(b)}(\theta),
\qquad
\hat\Gamma^{(b)}=\Gamma_{\hat\theta^{(b)}}.
\]

Given bootstrap draws $\{\hat\Gamma^{(b)}\}_{b=1}^B$, let
$
\hat s_{ij}
=
\left\{
\frac{1}{B-1}
\sum_{b=1}^B
\left[
(\hat\tau_{ij}^{(b)}-\hat\tau_{ij})
-
\frac{1}{B}\sum_{b'=1}^B(\hat\tau_{ij}^{(b')}-\hat\tau_{ij})
\right]^2
\right\}^{1/2}.
$
A pointwise confidence interval for $\tau_{ij}$ is
\[
\hat\tau_{ij}
\pm
\hat s_{ij}\,
q_{1-\alpha}\!\left(
\frac{|\hat\tau_{ij}^{(b)}-\hat\tau_{ij}|}{\hat s_{ij}}
\right).
\]
For simultaneous inference over an edge set
$\mathcal M\subseteq\{1,\dots,n\}^2$, we use the studentized max-deviation
critical value
\[
c_{1-\alpha}(\mathcal M)
=
q_{1-\alpha}\!\left(
\max_{(i,j)\in\mathcal M}
\frac{|\hat\tau_{ij}^{(b)}-\hat\tau_{ij}|}{\hat s_{ij}}
\right),
\qquad
\hat\tau_{ij}\pm c_{1-\alpha}(\mathcal M)\hat s_{ij},
\quad (i,j)\in\mathcal M.
\]
Similarly, node-level intervals can be constructed by applying the same
bootstrap draws to the corresponding linear functionals of $\Gamma$, as defined
in~\ref{def_1}, and using the analogous bootstrap standard errors for those
functionals.
\paragraph{One-step approximation to the weighted bootstrap.}
Exact bootstrap refitting can be expensive because the second-stage interference
model is parameterized by an MLP. We therefore approximate each weighted refit
by a one-step linearization. For a generic weight vector
$\omega=(\omega_i)_{i\in\bm V_1\cup\bm V_2}$, set
$L_{\tilde n}(\theta,\omega)=\tilde n^{-1}
\sum_{i\in\bm V_1\cup\bm V_2}\omega_i\ell_i(\theta)$,
$g_{\tilde n}(\theta,\omega)=\nabla_\theta L_{\tilde n}(\theta,\omega)$, and
$H_{\tilde n}(\theta,\omega)=\nabla_\theta^2 L_{\tilde n}(\theta,\omega)$.
If $\hat\theta(\omega)$ satisfies
$g_{\tilde n}(\hat\theta(\omega),\omega)=0$, then a first-order expansion around
$\hat\theta=\hat\theta(\mathbf 1)$ gives
\begin{align}\label{one_step}
   \hat\theta_{\mathrm{IJ}}(\omega)
=
\hat\theta
-
H_{\tilde n}(\hat\theta,\mathbf 1)^{-1}
\left\{
g_{\tilde n}(\hat\theta,\omega)
-
g_{\tilde n}(\hat\theta,\mathbf 1)
\right\}. 
\end{align}

Thus, for bootstrap replicate $b$, we set
$\hat\theta_{\mathrm{IJ}}^{(b)}=\hat\theta_{\mathrm{IJ}}(\xi^{(b)})$ and
$\hat\Gamma_{\mathrm{IJ}}^{(b)}=\Gamma_{\hat\theta_{\mathrm{IJ}}^{(b)}}$.

\section{Experiment}
In this section, we evaluate our method on three fronts: benchmark comparisons against baselines on two network datasets for causal effect estimation (Section~\ref{semi}), uncertainty quantification for heterogeneous spillover effects through confidence interval construction (Section~\ref{sec:bootstrap-ci-exp}), and interpretability analysis for practical edge-level spillover applications (Section~\ref{sec:edge-interpretability}).

\subsection{Benchmark comparisons on real networks}\label{semi}
\textbf{Data generation and setup.}
We evaluate our method in a semi-synthetic setting using two real social network datasets, BlogCatalog (BC) and Flickr, where node features $\bm{X}$ and network structure $\bm{A}$ are provided by the dataset. We then generate each node’s binary treatment $T_i$ and outcome $Y_i$ 
following \cite{ma2021causal,chen2024doubly,wu2025causal}   
from its own covariate $X_i$ and its one‑hop neighbors’ covariates $X_{\mathcal N_1(i)}$:

\begin{align*}T_i 
\sim 
\operatorname{Bernoulli}\!\left(
\sigma\!\left(f_T(\bm X_{\bar{\mathcal N}_1(i)})\right)
\right),\;
Y_i=
f_0(\bm X_{\bar{\mathcal N}_1(i)})
+
T_i\tau_i^\star
+
\sum_{j\in\mathcal N_1(i)}T_j\tau_{ij}^\star
+
\epsilon_i .
\end{align*}
where $f_T$ and $f_0$ are summary functions, $\sigma(\cdot)$ is the sigmoid function, and $\epsilon_i$ is random noise. The ground-truth effects $\tau_i^\star$ and $\tau_{ij}^\star$ are generated from node features, pairwise kernels, and local attention weights, with details provided in Appendix~\ref{Experiment_setup}.

To capture different interference patterns, we consider three choices of the pairwise function $W(\cdot,\cdot)$: 
(1) \textbf{Cosine} and \textbf{RBF}, which model heterogeneous interference based on both $X_i$ and $X_j$; 
(2) \textbf{One-way}, where $W(X_i,X_j)=f(X_j)$ depends only on the treated neighbor's features; and 
(3) \textbf{Homo}, where $W(\cdot,\cdot)$ is constant and interference is homogeneous. 
Settings (2) and (3) match the network causal models considered in prior work~\cite{ma2021causal,chen2024doubly,wu2025causal}, while setting (1) introduces more heterogeneous and nonlinear spillover patterns. We further vary the attention temperature $\beta$ to interpolate between uniform spillovers ($\beta=0$) and sparse spillovers concentrated on influential neighbors ($\beta=10$). In the main text, we report results for setting (1) on Flickr in Table~\ref{Flickr}; results on BC and for settings (2) and (3) are provided in Appendix~\ref{BC_result} and Appendix~\ref{heter_homo_result}, respectively.


\paragraph{Estimands and evaluation metrics.}
We evaluate the node-level and population-level estimands defined in Section~\ref{Section2}. For a population-level estimand $\psi\in\{\mathrm{ADE},\mathrm{ASE}\}$, we report
mean absolute error (MAE), defined as $|\hat\psi-\psi|$. For a node-level estimand $\psi_i\in\{\mathrm{IDE}_i,\mathrm{ISE}_i\}$, we report precision in estimating heterogeneous effects (PEHE), defined as $\sqrt{n^{-1}\sum_{i=1}^n(\hat\psi_i-\psi_i)^2}$.


\textbf{Baselines.} \textbf{CFR}~\cite{CFR} is a widely-used neural network model for heterogeneous treatment effect estimation, and we adapt it to network data
by incorporating neighborhood treatment and feature summaries as additional inputs. \textbf{NetEst}~\cite{jiang2022estimating}, \textbf{ND}~\cite{ND}, and \textbf{Caugamer}~\cite{wu2025causal} are GNN-based methods that learn balanced node representations to control for network confounding in estimating causal effects. \textbf{GDML}~\cite{khatami2024graph} and \textbf{Tnet}~\cite{chen2024doubly} both construct doubly robust estimators for ADE and ASE. \textbf{EdgeConv}~\cite{dgcnn} is a graph convolutional network with heterogeneous neighbor weights, which we adapt to causal effect estimation. It also serves as an ablation in section \ref{sec:edge-interpretability} that models interference heterogeneity without adjusting for network confounding.

\begin{table}[t!]
\centering
\scriptsize
\caption{\small Causal estimation performance on the Flickr network using Cosine and RBF kernels as interference function with varying temperatures (Temps.) $\beta\in\{0,1,5,10\}$. We highlight best performance in bold and the second-best with underline among  
the proposed method (\textbf{Proposed}\textsubscript{est}) and all baselines.
We also report results of our method with known nuisance (\textbf{Proposed}\textsubscript{oracle}).}
\label{Flickr}
\resizebox{\textwidth}{!}{
\begin{tabular}{|cccccccccc|cc|}
\hline
Interference & Temp.\ & Effect &
CFR & EdgeConv & ND & Netest & Tnet & Caugamer & GDML &
$\textrm{Proposed}_{\mathrm{est}}$ &
$\textrm{Proposed}_{\mathrm{oracle}}$\\
\hline
\multirow{16}{*}{\textbf{Cosine}}
  & \multirow{4}{*}{0}
      & {\tiny ADE} & {\tiny $0.1958\std{0.062}$} & {\tiny $0.1715\std{0.034}$} & {\tiny $0.2831\std{0.042}$} & {\tiny $0.1975\std{0.063}$} & {\tiny $0.1170\std{0.078}$} & {\tiny $0.0955\std{0.093}$} & {\tiny $\mathbf{0.0007}\std{0.009}$} & {\tiny $\underline{0.0120}\std{0.006}$} & {\tiny $0.0012\std{0.005}$} \\
  & & {\tiny ASE} & {\tiny $0.0364\std{0.019}$} & {\tiny $0.0867\std{0.015}$} & {\tiny $0.0558\std{0.038}$} & {\tiny $0.1396\std{0.047}$} & {\tiny $\mathbf{0.0038}\std{0.004}$} & {\tiny $0.1073\std{0.090}$} & {\tiny $0.0499\std{0.073}$} & {\tiny $\underline{0.0046}\std{0.009}$} & {\tiny $0.0112\std{0.016}$} \\
  & & {\tiny IDE} & {\tiny $0.3296\std{0.058}$} & {\tiny $0.2603\std{0.039}$} & {\tiny $0.3903\std{0.044}$} & {\tiny $0.3274\std{0.061}$} & {\tiny $0.3110\std{0.043}$} & {\tiny $0.3202\std{0.052}$} & {\tiny $\mathbf{0.0088}\std{0.009}$} & {\tiny $\underline{0.0277}\std{0.008}$} & {\tiny $0.0190\std{0.005}$} \\
  & & {\tiny ISE} & {\tiny $0.2724\std{0.012}$} & {\tiny $\underline{0.2102}\std{0.005}$} & {\tiny $0.3080\std{0.017}$} & {\tiny $0.3190\std{0.022}$} & {\tiny $0.3061\std{0.026}$} & {\tiny $0.3121\std{0.035}$} & {\tiny $0.2862\std{0.018}$} & {\tiny $\mathbf{0.0409}\std{0.004}$} & {\tiny $0.0381\std{0.010}$} \\
\cline{2-12}
  & \multirow{4}{*}{1}
      & {\tiny ADE} & {\tiny $0.2299\std{0.086}$} & {\tiny $0.2051\std{0.031}$} & {\tiny $0.2968\std{0.058}$} & {\tiny $0.2101\std{0.074}$} & {\tiny $0.1541\std{0.103}$} & {\tiny $0.0812\std{0.103}$} & {\tiny $\underline{0.0230}\std{0.005}$} & {\tiny $\mathbf{0.0080}\std{0.005}$} & {\tiny $0.0039\std{0.004}$} \\
  & & {\tiny ASE} & {\tiny $0.0409\std{0.022}$} & {\tiny $0.1259\std{0.020}$} & {\tiny $0.0450\std{0.040}$} & {\tiny $0.1221\std{0.056}$} & {\tiny $\mathbf{0.0024}\std{0.001}$} & {\tiny $0.1658\std{0.109}$} & {\tiny $0.0399\std{0.068}$} & {\tiny $\underline{0.0041}\std{0.013}$} & {\tiny $0.0033\std{0.005}$} \\
  & & {\tiny IDE} & {\tiny $0.3577\std{0.065}$} & {\tiny $0.2868\std{0.034}$} & {\tiny $0.4035\std{0.055}$} & {\tiny $0.3374\std{0.064}$} & {\tiny $0.3407\std{0.055}$} & {\tiny $0.3404\std{0.040}$} & {\tiny $\underline{0.0763}\std{0.009}$} & {\tiny $\mathbf{0.0252}\std{0.008}$} & {\tiny $0.0158\std{0.005}$} \\
  & & {\tiny ISE} & {\tiny $0.2925\std{0.024}$} & {\tiny $\underline{0.2405}\std{0.010}$} & {\tiny $0.3192\std{0.005}$} & {\tiny $0.3498\std{0.024}$} & {\tiny $0.3160\std{0.022}$} & {\tiny $0.3541\std{0.048}$} & {\tiny $0.2819\std{0.018}$} & {\tiny $\mathbf{0.0499}\std{0.005}$} & {\tiny $0.0362\std{0.006}$} \\
\cline{2-12}
  & \multirow{4}{*}{5}
      & {\tiny ADE} & {\tiny $0.1701\std{0.090}$} & {\tiny $0.3342\std{0.036}$} & {\tiny $0.1711\std{0.077}$} & {\tiny $0.0896\std{0.069}$} & {\tiny $0.2841\std{0.173}$} & {\tiny $\underline{0.0758}\std{0.059}$} & {\tiny $0.1043\std{0.022}$} & {\tiny $\mathbf{0.0028}\std{0.002}$} & {\tiny $0.0018\std{0.005}$} \\
  & & {\tiny ASE} & {\tiny $0.1620\std{0.076}$} & {\tiny $0.1431\std{0.048}$} & {\tiny $0.1723\std{0.112}$} & {\tiny $0.1182\std{0.080}$} & {\tiny $0.0313\std{0.037}$} & {\tiny $0.2039\std{0.177}$} & {\tiny $\underline{0.0282}\std{0.074}$} & {\tiny $\mathbf{0.0012}\std{0.005}$} & {\tiny $0.0046\std{0.005}$} \\
  & & {\tiny IDE} & {\tiny $0.3847\std{0.048}$} & {\tiny $0.4297\std{0.032}$} & {\tiny $0.3790\std{0.039}$} & {\tiny $0.2999\std{0.037}$} & {\tiny $0.4852\std{0.094}$} & {\tiny $0.3922\std{0.127}$} & {\tiny $\underline{0.2974}\std{0.024}$} & {\tiny $\mathbf{0.0282}\std{0.005}$} & {\tiny $0.0184\std{0.005}$} \\
  & & {\tiny ISE} & {\tiny $0.4102\std{0.042}$} & {\tiny $0.3093\std{0.030}$} & {\tiny $0.4085\std{0.051}$} & {\tiny $0.4462\std{0.022}$} & {\tiny $0.3863\std{0.039}$} & {\tiny $0.4334\std{0.083}$} & {\tiny $\underline{0.2643}\std{0.023}$} & {\tiny $\mathbf{0.0295}\std{0.004}$} & {\tiny $0.0178\std{0.005}$} \\
\cline{2-12}
  & \multirow{4}{*}{10}
      & {\tiny ADE} & {\tiny $0.1558\std{0.064}$} & {\tiny $0.3398\std{0.044}$} & {\tiny $0.1689\std{0.093}$} & {\tiny $0.0785\std{0.047}$} & {\tiny $0.3014\std{0.171}$} & {\tiny $\underline{0.0249}\std{0.028}$} & {\tiny $0.1124\std{0.022}$} & {\tiny $\mathbf{0.0063}\std{0.008}$} & {\tiny $0.0017\std{0.006}$} \\
  & & {\tiny ASE} & {\tiny $0.1357\std{0.098}$} & {\tiny $0.1450\std{0.054}$} & {\tiny $0.1836\std{0.088}$} & {\tiny $0.0906\std{0.064}$} & {\tiny $\underline{0.0278}\std{0.044}$} & {\tiny $0.1875\std{0.166}$} & {\tiny $0.0435\std{0.098}$} & {\tiny $\mathbf{0.0029}\std{0.004}$} & {\tiny $0.0048\std{0.006}$} \\
  & & {\tiny IDE} & {\tiny $0.3824\std{0.033}$} & {\tiny $0.4382\std{0.048}$} & {\tiny $0.3834\std{0.050}$} & {\tiny $\underline{0.2945}\std{0.016}$} & {\tiny $0.4986\std{0.091}$} & {\tiny $0.3106\std{0.047}$} & {\tiny $0.3188\std{0.024}$} & {\tiny $\mathbf{0.0293}\std{0.009}$} & {\tiny $0.0200\std{0.006}$} \\
  & & {\tiny ISE} & {\tiny $0.4276\std{0.053}$} & {\tiny $0.3091\std{0.020}$} & {\tiny $0.4218\std{0.046}$} & {\tiny $0.4384\std{0.015}$} & {\tiny $0.3934\std{0.039}$} & {\tiny $0.4467\std{0.064}$} & {\tiny $\underline{0.2838}\std{0.023}$} & {\tiny $\mathbf{0.0317}\std{0.003}$} & {\tiny $0.0204\std{0.005}$} \\
\cline{1-12}
\multirow{16}{*}{\textbf{RBF}}
  & \multirow{4}{*}{0}
      & {\tiny ADE} & {\tiny $0.2116\std{0.069}$} & {\tiny $0.1680\std{0.028}$} & {\tiny $0.2758\std{0.077}$} & {\tiny $0.2213\std{0.073}$} & {\tiny $0.1142\std{0.076}$} & {\tiny $\underline{0.0134}\std{0.009}$} & {\tiny $\mathbf{0.0013}\std{0.004}$} & {\tiny $0.0155\std{0.007}$} & {\tiny $0.0007\std{0.006}$} \\
  & & {\tiny ASE} & {\tiny $0.0410\std{0.035}$} & {\tiny $0.1378\std{0.023}$} & {\tiny $0.0442\std{0.029}$} & {\tiny $0.1535\std{0.039}$} & {\tiny $\mathbf{0.0032}\std{0.002}$} & {\tiny $0.1455\std{0.081}$} & {\tiny $0.0506\std{0.060}$} & {\tiny $\underline{0.0044}\std{0.010}$} & {\tiny $0.0067\std{0.011}$} \\
  & & {\tiny IDE} & {\tiny $0.3360\std{0.063}$} & {\tiny $0.2552\std{0.032}$} & {\tiny $0.3834\std{0.076}$} & {\tiny $0.3448\std{0.073}$} & {\tiny $0.3095\std{0.044}$} & {\tiny $0.3057\std{0.037}$} & {\tiny $\mathbf{0.0034}\std{0.005}$} & {\tiny $\underline{0.0327}\std{0.009}$} & {\tiny $0.0164\std{0.005}$} \\
  & & {\tiny ISE} & {\tiny $0.2287\std{0.013}$} & {\tiny $0.2119\std{0.013}$} & {\tiny $0.2439\std{0.011}$} & {\tiny $0.3030\std{0.027}$} & {\tiny $0.2492\std{0.021}$} & {\tiny $0.2694\std{0.030}$} & {\tiny $\underline{0.2111}\std{0.022}$} & {\tiny $\mathbf{0.0378}\std{0.005}$} & {\tiny $0.0281\std{0.008}$} \\
\cline{2-12}
  & \multirow{4}{*}{1}
      & {\tiny ADE} & {\tiny $0.2188\std{0.072}$} & {\tiny $0.2181\std{0.031}$} & {\tiny $0.2623\std{0.040}$} & {\tiny $0.2176\std{0.073}$} & {\tiny $0.1720\std{0.111}$} & {\tiny $0.0597\std{0.079}$} & {\tiny $\underline{0.0339}\std{0.008}$} & {\tiny $\mathbf{0.0090}\std{0.005}$} & {\tiny $0.0058\std{0.005}$} \\
  & & {\tiny ASE} & {\tiny $0.0626\std{0.038}$} & {\tiny $0.1573\std{0.030}$} & {\tiny $0.0814\std{0.049}$} & {\tiny $0.1611\std{0.045}$} & {\tiny $\underline{0.0097}\std{0.020}$} & {\tiny $0.1602\std{0.147}$} & {\tiny $0.0289\std{0.052}$} & {\tiny $\mathbf{0.0038}\std{0.008}$} & {\tiny $0.0018\std{0.014}$} \\
  & & {\tiny IDE} & {\tiny $0.3535\std{0.061}$} & {\tiny $0.2954\std{0.031}$} & {\tiny $0.3793\std{0.044}$} & {\tiny $0.3426\std{0.061}$} & {\tiny $0.1708\std{0.033}$} & {\tiny $0.3242\std{0.068}$} & {\tiny $\underline{0.0595}\std{0.015}$} & {\tiny $\mathbf{0.0405}\std{0.007}$} & {\tiny $0.0281\std{0.004}$} \\
  & & {\tiny ISE} & {\tiny $0.2903\std{0.015}$} & {\tiny $0.2918\std{0.020}$} & {\tiny $0.2755\std{0.021}$} & {\tiny $0.3145\std{0.077}$} & {\tiny $\underline{0.2478}\std{0.005}$} & {\tiny $0.3301\std{0.063}$} & {\tiny $0.2871\std{0.060}$} & {\tiny $\mathbf{0.0800}\std{0.020}$} & {\tiny $0.0575\std{0.008}$} \\
\cline{2-12}
  & \multirow{4}{*}{5}
      & {\tiny ADE} & {\tiny $0.1701\std{0.090}$} & {\tiny $0.3342\std{0.036}$} & {\tiny $0.1711\std{0.077}$} & {\tiny $0.0896\std{0.069}$} & {\tiny $0.2841\std{0.173}$} & {\tiny $\underline{0.0758}\std{0.059}$} & {\tiny $0.1043\std{0.022}$} & {\tiny $\mathbf{0.0028}\std{0.002}$} & {\tiny $0.0018\std{0.005}$} \\
  & & {\tiny ASE} & {\tiny $0.1620\std{0.076}$} & {\tiny $0.1431\std{0.048}$} & {\tiny $0.1723\std{0.112}$} & {\tiny $0.1182\std{0.080}$} & {\tiny $0.0313\std{0.037}$} & {\tiny $0.2039\std{0.177}$} & {\tiny $\underline{0.0282}\std{0.074}$} & {\tiny $\mathbf{0.0012}\std{0.005}$} & {\tiny $0.0046\std{0.005}$} \\
  & & {\tiny IDE} & {\tiny $0.3847\std{0.048}$} & {\tiny $0.4297\std{0.032}$} & {\tiny $0.3790\std{0.039}$} & {\tiny $0.2999\std{0.037}$} & {\tiny $0.4852\std{0.094}$} & {\tiny $0.3922\std{0.127}$} & {\tiny $\underline{0.2974}\std{0.024}$} & {\tiny $\mathbf{0.0282}\std{0.005}$} & {\tiny $0.0184\std{0.005}$} \\
  & & {\tiny ISE} & {\tiny $0.4102\std{0.042}$} & {\tiny $0.3093\std{0.030}$} & {\tiny $0.4085\std{0.051}$} & {\tiny $0.4462\std{0.022}$} & {\tiny $0.3863\std{0.039}$} & {\tiny $0.4334\std{0.083}$} & {\tiny $\underline{0.2643}\std{0.023}$} & {\tiny $\mathbf{0.0295}\std{0.004}$} & {\tiny $0.0178\std{0.005}$} \\
\cline{2-12}
  & \multirow{4}{*}{10}
      & {\tiny ADE} & {\tiny $0.1558\std{0.064}$} & {\tiny $0.3398\std{0.044}$} & {\tiny $0.1689\std{0.093}$} & {\tiny $0.0785\std{0.047}$} & {\tiny $0.3014\std{0.171}$} & {\tiny $\underline{0.0249}\std{0.028}$} & {\tiny $0.1124\std{0.022}$} & {\tiny $\mathbf{0.0063}\std{0.008}$} & {\tiny $0.0017\std{0.006}$} \\
  & & {\tiny ASE} & {\tiny $0.1357\std{0.098}$} & {\tiny $0.1450\std{0.054}$} & {\tiny $0.1836\std{0.088}$} & {\tiny $0.0906\std{0.064}$} & {\tiny $\underline{0.0278}\std{0.044}$} & {\tiny $0.1875\std{0.166}$} & {\tiny $0.0435\std{0.098}$} & {\tiny $\mathbf{0.0029}\std{0.004}$} & {\tiny $0.0048\std{0.006}$} \\
  & & {\tiny IDE} & {\tiny $0.3824\std{0.033}$} & {\tiny $0.4382\std{0.048}$} & {\tiny $0.3834\std{0.050}$} & {\tiny $\underline{0.2945}\std{0.016}$} & {\tiny $0.4986\std{0.091}$} & {\tiny $0.3106\std{0.047}$} & {\tiny $0.3188\std{0.024}$} & {\tiny $\mathbf{0.0293}\std{0.009}$} & {\tiny $0.0200\std{0.006}$} \\
  & & {\tiny ISE} & {\tiny $0.4276\std{0.053}$} & {\tiny $0.3091\std{0.020}$} & {\tiny $0.4218\std{0.046}$} & {\tiny $0.4384\std{0.015}$} & {\tiny $0.3934\std{0.039}$} & {\tiny $0.4467\std{0.064}$} & {\tiny $\underline{0.2838}\std{0.023}$} & {\tiny $\mathbf{0.0317}\std{0.003}$} & {\tiny $0.0204\std{0.005}$} \\
\hline
\end{tabular}
}
\end{table}

\textbf{Evaluations.}
Table~\ref{Flickr} reports out-of-sample estimation accuracy on the semi-synthetic Flickr dataset under \textbf{Cosine} and \textbf{RBF} interference kernels. The proposed estimator (\textbf{Proposed}$_{\text{est}}$) is competitive on population-level metrics (ADE and ASE), closely matching DBML and TNet, while substantially improving node-level estimates (IDE and ISE). The gains increase with the temperature $\beta$, as spillovers become more concentrated on a few influential neighbors. We also compare with the oracle estimator (\textbf{Proposed}$_{\text{oracle}}$), which uses the true nuisance components. The small gap between the estimated and oracle versions supports the effectiveness of the orthogonal design in mitigating nuisance-estimation bias. 
\paragraph{Additional Experiments.} In Appendix \ref{app:interaction_mispeci}, we evaluate more complex spillover patterns in which effects depend jointly on a unit's own treatment and its neighbors' treatments, testing robustness to model misspecification. Moreover, we consider a more general spillover definition based on gradual neighborhood interventions (Appendix~\ref{app:bernoulli-z}). Finally, we apply our method to a real air-pollution study (Appendix~\ref{app:real_world_dapsm}).

\subsection{Synthetic network experiments for heterogeneous spillovers}\label{SBM}

The benchmark experiments evaluate standard node- and population-level estimands. We now use synthetic networks to assess finer-grained edge-level behavior. Motivated by the \textit{political polarization} example in Section~\ref{Intro}, we simulate SBM networks with three latent communities representing different ideological groups. We use \(n=3000\) nodes for training and \(n=1000\) for evaluation. Node features are drawn from community-specific distributions, and the raw pairwise influences $\{w_{ij}\}$ are constructed from node embeddings using a cosine kernel. This design allows within- and between-community effects to vary in both magnitude and sign, thereby mimicking heterogeneous political influence under polarization.\citep{bail2018exposure}.

\subsubsection{Confidence interval construction}\label{sec:bootstrap-ci-exp}

We evaluate the calibration of the proposed second-stage bootstrap for heterogeneous spillover effects. Specifically, we compare confidence intervals under two nuisance regimes: \textbf{oracle} nuisance functions and GNN-\textbf{estimated} nuisance functions. For each regime, we construct bootstrap intervals for both edge-level effects $\tau_{ij}$ and node-level spillover effects $\mathrm{ISE}_i$. For each target, we report pointwise coverage, simultaneous uniform coverage, and average interval length at nominal levels $90\%$ and $95\%$.

\begin{wrapfigure}{r}{0.53\textwidth}
\vspace{-14pt}
\centering
\begin{subfigure}{0.48\linewidth}
    \centering
    \includegraphics[width=\linewidth]{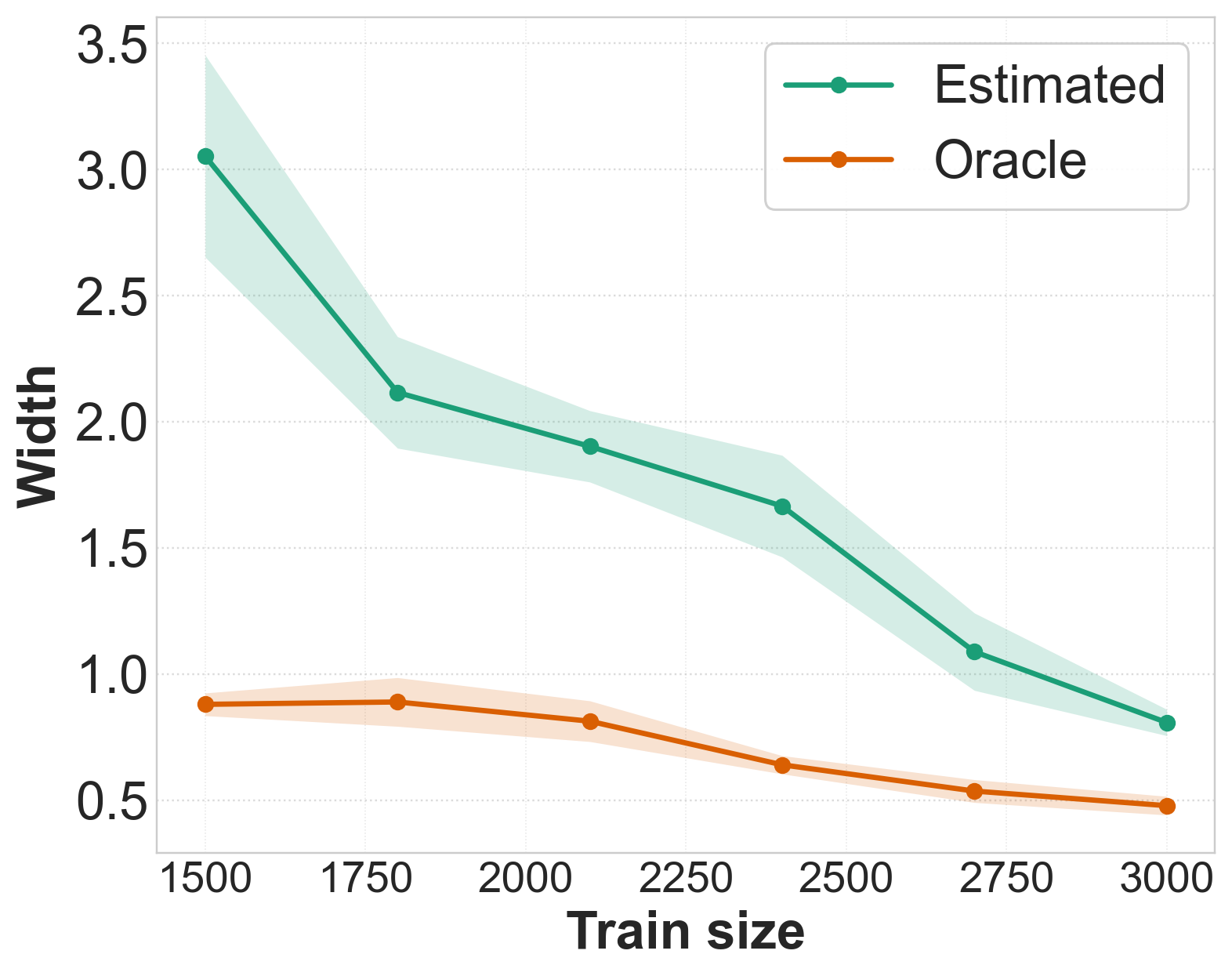}
    \caption{}
    \label{fig:ci-width-train}
\end{subfigure}
\hfill
\begin{subfigure}{0.48\linewidth}
    \centering
    \includegraphics[width=\linewidth]{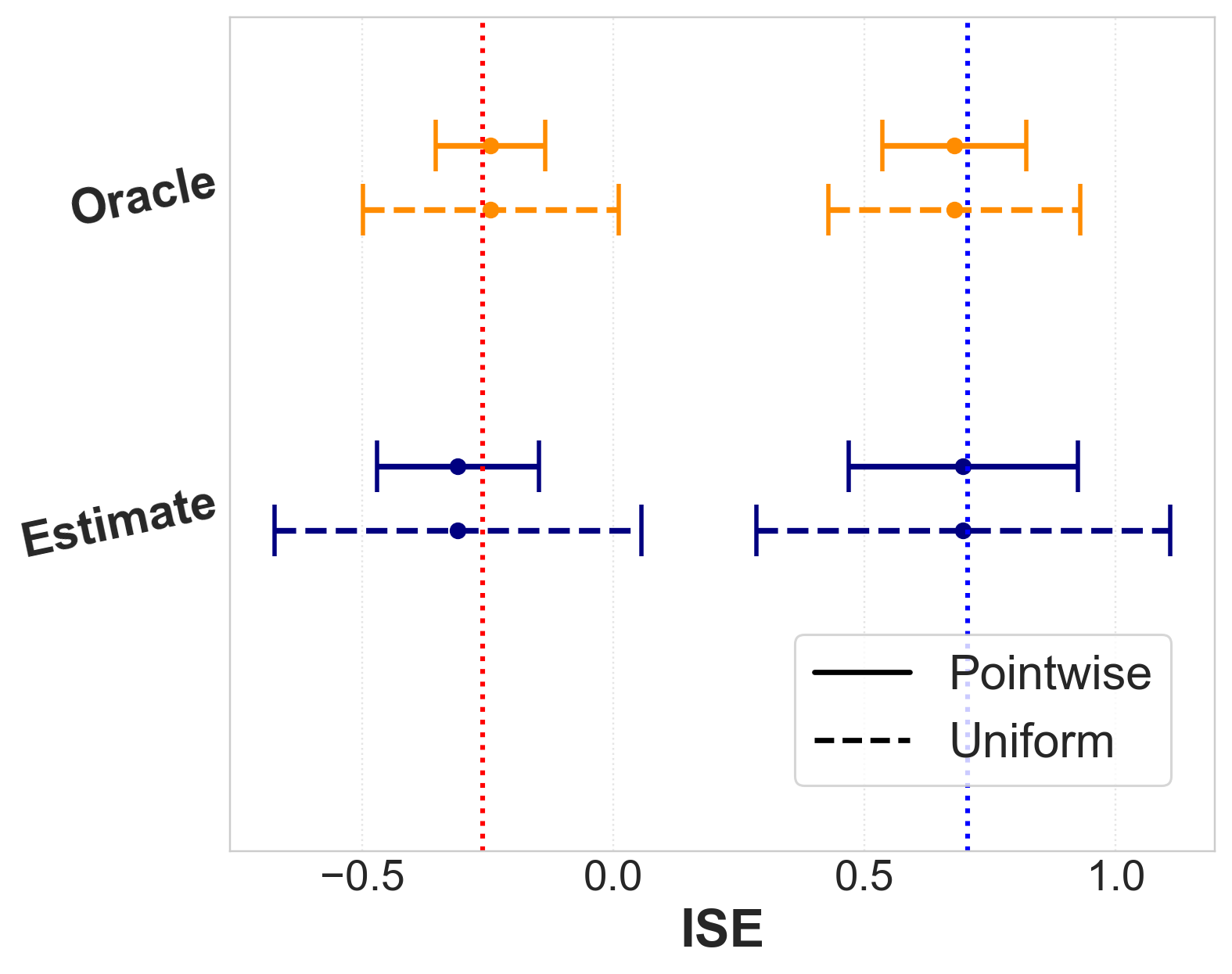}
    \caption{}
    \label{fig:ci-node-example}
\end{subfigure}
\caption{
\textbf{Node-level confidence intervals for ISE at nominal level $90\%$.}
\textit{Left:} Average uniform interval width decreases with training size, and estimated nuisances produce wider intervals than oracle nuisances.
\textit{Right:} Pointwise and uniform intervals for one negative-ISE node and one positive-ISE node under oracle and estimated nuisances.
}
\label{fig:two_diagrams}
\vspace{-8pt}
\end{wrapfigure}

\paragraph{Results.} 
Table~\ref{tab:bootstrap-ci} shows that the bootstrap intervals achieve coverage close to the nominal levels for both $\tau_{ij}$ and $\mathrm{ISE}_i$. Intervals are generally wider under GNN-estimated nuisances than under oracle nuisances, reflecting the finite-sample cost of first-stage estimation. As shown in Figure~\ref{fig:ci-width-train}, this gap decreases as the training size $n$ increases. Uniform intervals are also wider than pointwise intervals, consistent with their stronger joint-coverage target over the evaluation edge set. Figure~\ref{fig:ci-node-example} further illustrates that, for representative positive- and negative-spillover nodes, both pointwise and uniform intervals cover the true $\mathrm{ISE}$ values.

\begin{table}[t]
\centering
\caption{
Multiplier bootstrap confidence interval calibration on the SBM experiment.
Coverage is reported at nominal level $1-\alpha$; interval lengths are averaged over the same evaluation mask.
We use $B=200$ bootstrap draws and report results averaged over 200 repeated runs.
}
\label{tab:bootstrap-ci}
\scriptsize
\setlength{\tabcolsep}{3.5pt}
\renewcommand{\arraystretch}{0.92}
\resizebox{\textwidth}{!}{
\begin{tabular}{llcccccccc}
\toprule
& & \multicolumn{4}{c}{$90\%$} & \multicolumn{4}{c}{$95\%$} \\
\cmidrule(lr){3-6} \cmidrule(lr){7-10}
Target & Nuisance
& Point Cov. & Unif. Cov. & Point Len. & Unif. Len.
& Point Cov. & Unif. Cov. & Point Len. & Unif. Len.\\
\midrule
$\tau_{ij}$      & Oracle    & 0.898 & 0.895 & 0.315 & 0.721 & 0.953 & 0.955 & 0.333 & 0.798 \\
$\tau_{ij}$      & Estimated & 0.905 & 0.890 & 0.473 & 1.236 & 0.946 & 0.940 & 0.555 & 1.290 \\
$\mathrm{ISE}_i$ & Oracle    & 0.893 & 0.885 & 0.192 & 0.455 & 0.948 & 0.945 & 0.232 & 0.495 \\
$\mathrm{ISE}_i$ & Estimated & 0.908 & 0.900 & 0.325 & 0.729 & 0.944 & 0.940 & 0.407 & 0.839 \\
\bottomrule
\end{tabular}
}
\end{table}

\subsubsection{Causal estimation interpretability}
\label{sec:edge-interpretability}
In this section, we evaluate whether the estimated edge-level effects recover interpretable spillover structure, focusing on tasks that are useful for downstream analysis.

\begin{wrapfigure}{r}{0.52\textwidth}
\vspace{-10pt}
\centering
\begin{subfigure}[t]{0.26\textwidth}
    \centering
    \includegraphics[width=\linewidth]{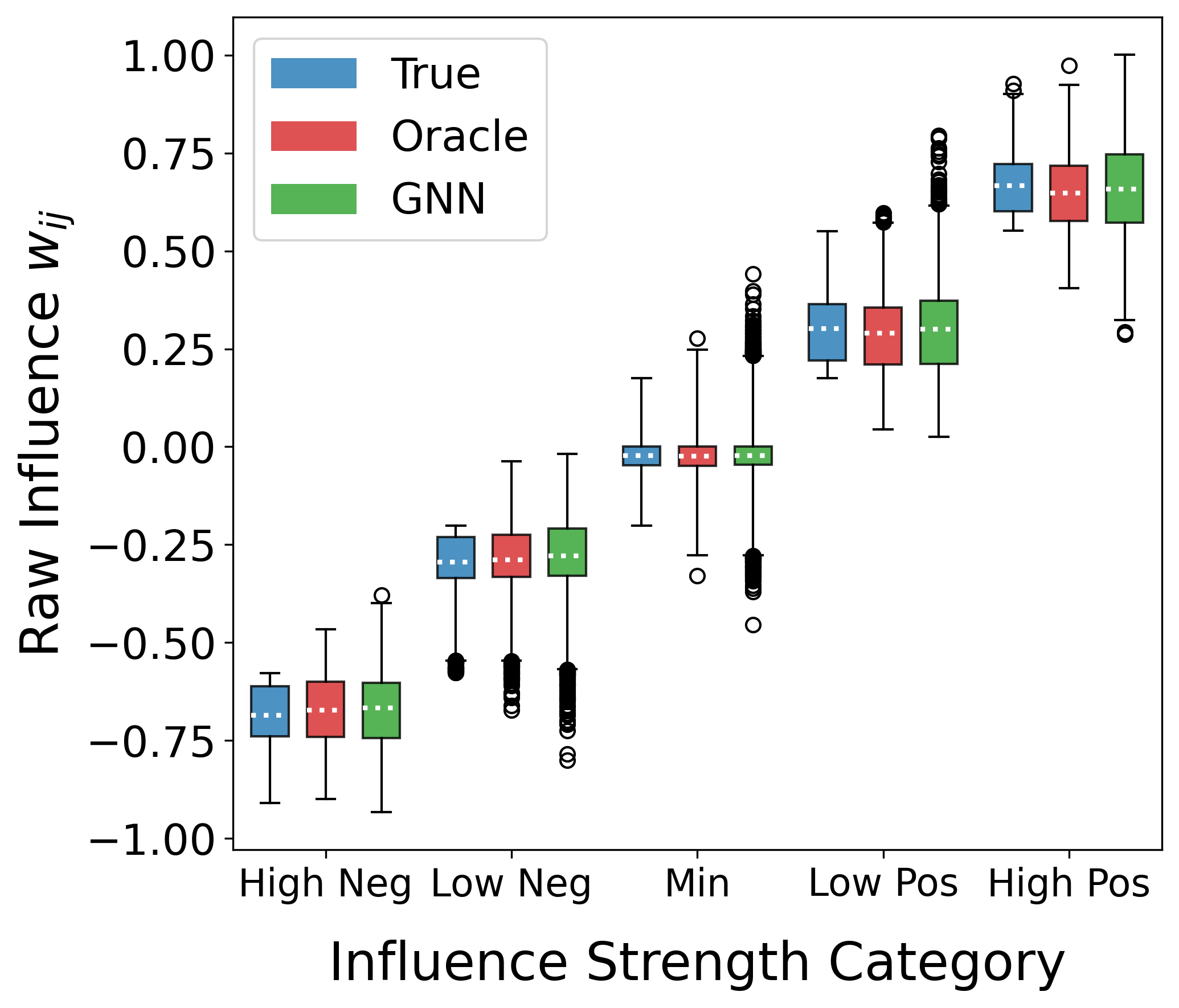}
    \caption{}
    \label{grouped}
\end{subfigure}
\hfill
\begin{subfigure}[t]{0.22\textwidth}
    \centering
    \includegraphics[width=\linewidth]{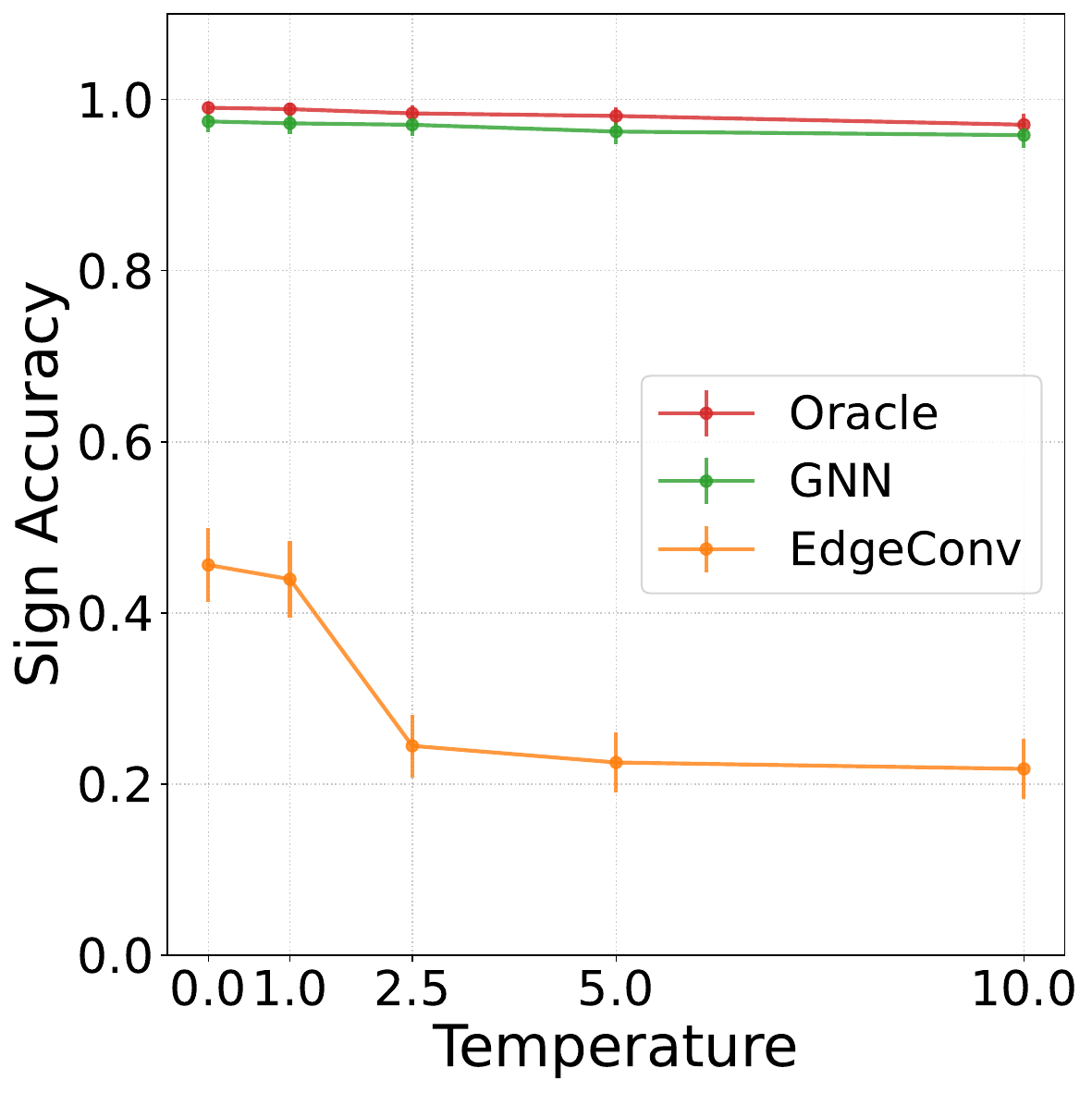}
    \caption{}
    \label{sign}
\end{subfigure}

\vspace{4pt}

\begin{subfigure}[t]{0.22\textwidth}
    \centering
    \includegraphics[width=\linewidth]{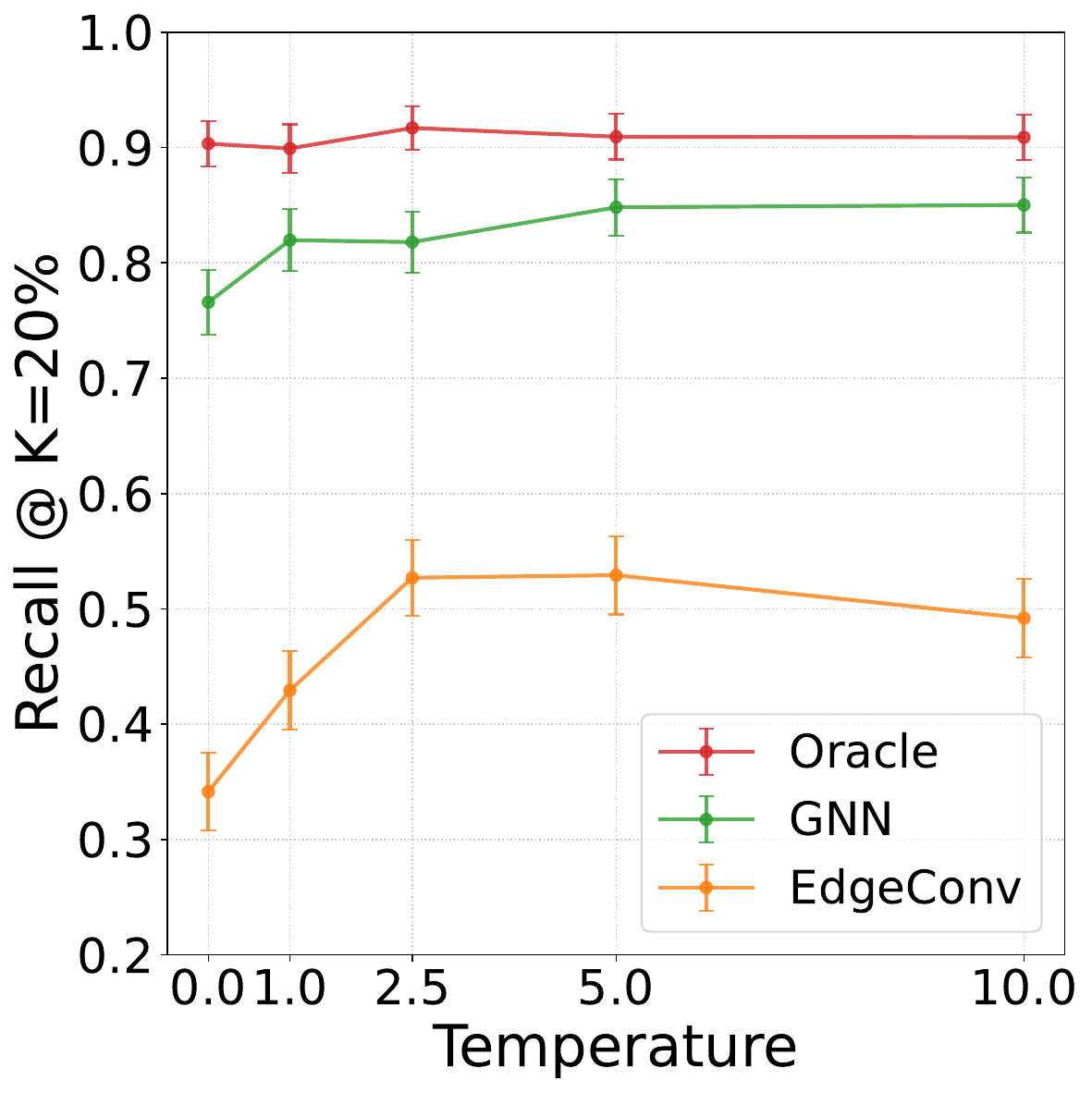}
    \caption{}
    \label{recall}
\end{subfigure}
\hfill
\begin{subfigure}[t]{0.22\textwidth}
    \centering
    \includegraphics[width=\linewidth]{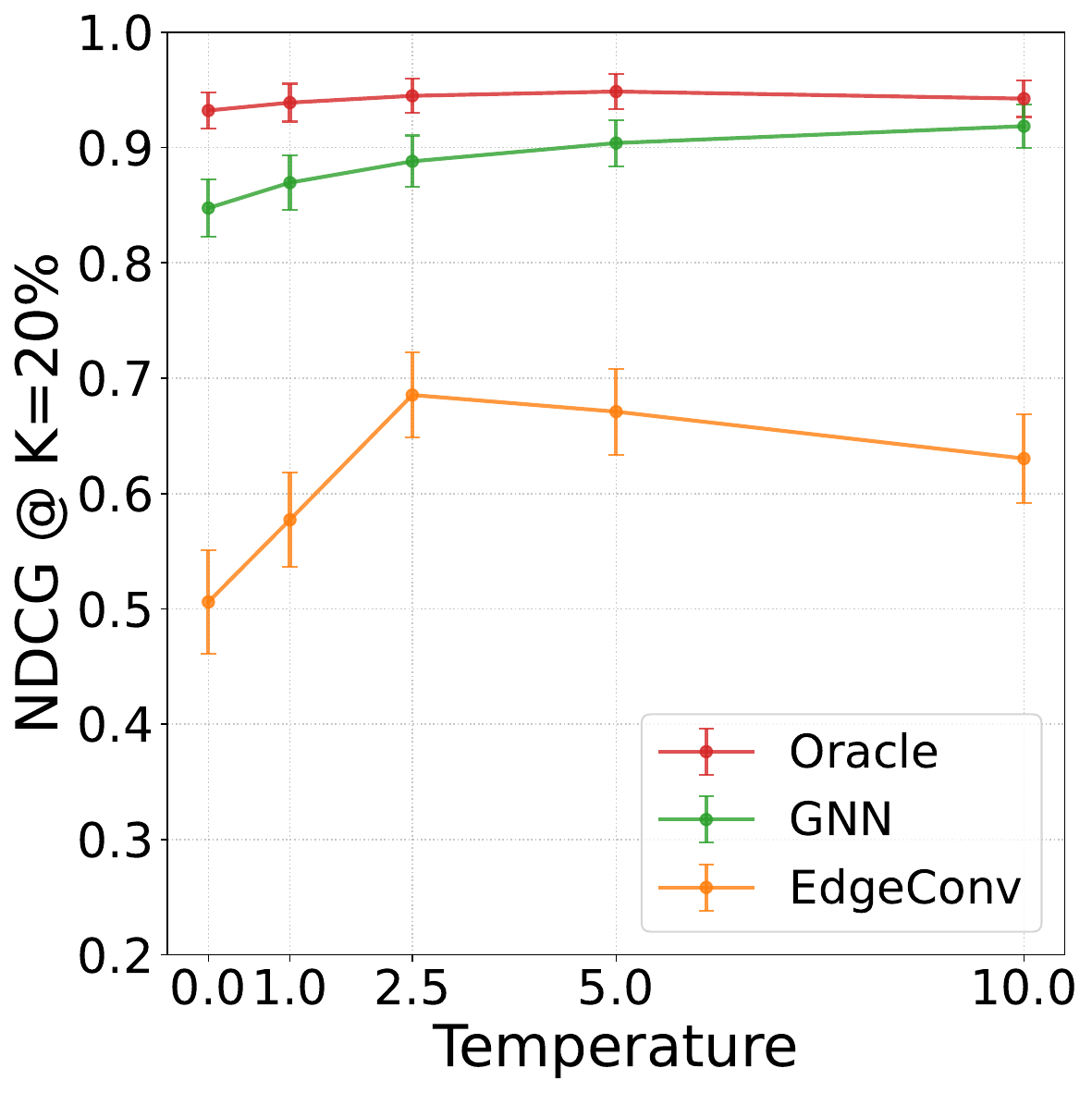}
    \caption{}
    \label{ndcg}
\end{subfigure}

\vspace{4pt}
\caption{\small Edge-level interference estimation. (a) Pairwise influence recovery. (b) Spillover sign recovery. (c,d) Influential neighbors detection.}
\label{Ablation}
\vspace{-10pt}
\end{wrapfigure}

\textbf{Pairwise influence recovery.} 
We group the true raw pairwise score \{\(w_{ij}\)\} into five groups based on their signs and magnitudes to reflect different ideological interaction strength. Figure~\ref{grouped} compares the group-specific distributions of (1) true \(w_{ij}\) (\textbf{True}), (2) estimated \(\hat{w}_{ij}\) using the true nuisance parameters \(m^\star, e^\star\) (\textbf{Oracle}), and (3) estimated \(\hat{w}_{ij}\) using cross-fitted GNN-based nuisance estimators \(\hat{m}, \hat{e}\) (\textbf{GNN}). Our method accurately identifies five categories distinguished by intensity and sign of pairwise influence. Within each category, our estimator closely matches the true medians and quantiles of $\{{w}_{ij}\}$, while exhibiting slightly larger variance due to expected nuisance-estimation noise.

\textbf{Influential neighbors detection.} For each ego node \( i \), we evaluate our method's performance in identifying its influential neighbors, defined as $i$'s top 20\% of neighbors ranked by the magnitude of their interference coefficients \(\{|\tau_{ij}|\}_{j\in\mathcal{N}_1(i)}\). We measure performance using two complementary metrics: (1) \textbf{Recall@K=20\%}: how well the model’s predicted top 20\% neighbors \textit{cover} the true top 20\%; (2) \textbf{NDCG@K=20\%}: how well the model \textit{ranks} its predicted top 20\% neighbors by rewarding higher placement of neighbors with larger true $\{|\tau_{ij}|\}$. The formal definitions of the two metrics are in Appendix \ref{appendix_SBM}. In addition to Oracle and GNN, we include \textbf{EdgeConv} as a baseline to illustrate performance without adjusting for network confounding. 

Figure~\ref{recall} and \ref{ndcg} show that our method consistently recovers over 80\% of the true top-20\% neighbors of ego nodes on average, and perform closely to oracle estimation. Our method improves slightly with increasing temperature $\beta$ due to that higher $\beta$ amplifies the magnitude of spillovers from top neighbors hence making them easier to identify. In contrast, EdgeConv performs substantially worse.

\textbf{Spillover sign recovery.} We evaluate the performance in recovering the spillover signs for each node $i$'s one-hop neighbors using the metric $\frac{1}{|\mathcal{N}_1(i)|}\sum_{j\in \mathcal{N}_1(i)}\mathbf{1}\bigl[\text{sign}(\hat{w}_{ij}) =\text{sign}({w}_{ij})\big]$. Figure~\ref{sign} shows the average estimation accuracy across all nodes. Our method achieves near-perfect recovery, significantly outperforming EdgeConv. 

     

\section{Conclusion}

In this paper, we propose an orthogonal learning framework for estimating heterogeneous causal effects on networks. Our method combines graph-based nuisance estimation with an interpretable attention-based interference model, enabling the estimation of direct, spillover, and edge-level effects under network confounding. We further develop a bootstrap-based procedure for uncertainty quantification of heterogeneous spillover effects. Experiments on semi-synthetic and synthetic networks show that our method improves node- and edge-level estimation while supporting fine-grained analyses for practical downstream tasks. 


\bibliographystyle{plain}
\bibliography{neurips_2025}

@misc{asuSocialComputingData,
  title        = {Social Computing Data Repository},
  author       = {{Data Mining and Machine Learning Laboratory, Arizona State University}},
  year         = {2010},
  howpublished = {\url{https://datasets.syr.edu/pages/datasets.html}},
  note         = {Accessed: 2026-05-04}
}

@inproceedings{huang2017label,
  title     = {Label Informed Attributed Network Embedding},
  author    = {Huang, Xiao and Li, Jundong and Hu, Xia},
  booktitle = {Proceedings of the Tenth ACM International Conference on Web Search and Data Mining},
  series    = {WSDM '17},
  pages     = {731--739},
  year      = {2017},
  publisher = {ACM},
  doi       = {10.1145/3018661.3018667}
}

@article{papadogeorgou2019causal,
  title={Causal inference with interfering units for cluster and population level treatment allocation programs},
  author={Papadogeorgou, Georgia and Mealli, Fabrizia and Zigler, Corwin M.},
  journal={Biometrics},
  volume={75},
  number={3},
  pages={778--787},
  year={2019},
  doi={10.1111/biom.13049}
}

@article{papadogeorgou2019adjusting,
  title={Adjusting for unmeasured spatial confounding with distance adjusted propensity score matching},
  author={Papadogeorgou, Georgia and Choirat, Christine and Zigler, Corwin M.},
  journal={Biostatistics},
  volume={20},
  number={2},
  pages={256--272},
  year={2019},
  doi={10.1093/biostatistics/kxx074}
}

@article{chernozhukov2013gaussian,
  title={Gaussian approximations and multiplier bootstrap for maxima of sums of high-dimensional random vectors},
  author={Chernozhukov, Victor and Chetverikov, Denis and Kato, Kengo},
  journal={The Annals of Statistics},
  volume={41},
  number={6},
  pages={2786--2819},
  year={2013},
  doi={10.1214/13-AOS1161}
}

@article{hudgens2008toward,
  title={Toward causal inference with interference},
  author={Hudgens, Michael G. and Halloran, M. Elizabeth},
  journal={Journal of the American Statistical Association},
  volume={103},
  number={482},
  pages={832--842},
  year={2008}
}

@article{aronow2017estimating,
  title={Estimating average causal effects under general interference, with application to a social network experiment},
  author={Aronow, Peter M. and Samii, Cyrus},
  journal={The Annals of Applied Statistics},
  volume={11},
  number={4},
  pages={1912--1947},
  year={2017}
}

@article{leung2022causal,
  title={Causal inference under approximate neighborhood interference},
  author={Leung, Michael P.},
  journal={Econometrica},
  volume={90},
  number={1},
  pages={267--293},
  year={2022}
}

@article{zhao2024dwr,
  title   = {Learning Individual Treatment Effects under Heterogeneous Interference in Networks},
  author  = {Zhao, Ziyu and Bai, Yuqi and Xiong, Ruoxuan and Cao, Qingyu and Ma, Chao and Jiang, Ning and Wu, Fei and Kuang, Kun},
  journal = {ACM Transactions on Knowledge Discovery from Data},
  volume  = {18},
  number  = {8},
  pages   = {199:1--199:21},
  year    = {2024},
  doi     = {10.1145/3673761}
}

@inproceedings{fan2026idenet,
  title     = {Identifiable Disentangled Representation Learning for Causal Inference under Network Interference},
  author    = {Fan, Di and Gao, Chuanhou},
  booktitle = {Proceedings of the Nineteenth ACM International Conference on Web Search and Data Mining},
  pages     = {142--152},
  year      = {2026},
  publisher = {ACM},
  doi       = {10.1145/3773966.3777967}
}

@inproceedings{lin2023hinite,
  title     = {Estimating Treatment Effects Under Heterogeneous Interference},
  author    = {Lin, Xiaofeng and Zhang, Guoxi and Lu, Xiaotian and Bao, Han and Takeuchi, Koh and Kashima, Hisashi},
  booktitle = {Machine Learning and Knowledge Discovery in Databases: Research Track},
  pages     = {576--592},
  year      = {2023},
  publisher = {Springer Nature Switzerland},
  doi       = {10.1007/978-3-031-43412-9_34}
}

@article{liu2023ndr,
  title   = {Nonparametric Doubly Robust Estimation of Causal Effect on Networks in Observational Studies},
  author  = {Liu, Jie and Ye, Fangjuan and Yang, Yang},
  journal = {Stat},
  volume  = {12},
  number  = {1},
  pages   = {e549},
  year    = {2023},
  doi     = {10.1002/sta4.549}
}

@article{huang2024spnet,
  title   = {Modeling Interference for Individual Treatment Effect Estimation from Networked Observational Data},
  author  = {Huang, Qiang and Ma, Jing and Li, Jundong and Guo, Ruocheng and Sun, Huiyan and Chang, Yi},
  journal = {ACM Transactions on Knowledge Discovery from Data},
  volume  = {18},
  number  = {3},
  pages   = {48:1--48:21},
  year    = {2024},
  doi     = {10.1145/3628449}
}

@article{box1986bayesian,
  title={An analysis for unreplicated fractional factorials},
  author={Box, George E. P. and Meyer, R. Daniel},
  journal={Technometrics},
  volume={28},
  number={1},
  pages={11--18},
  year={1986}
}

@article{yuan2007efficient,
  title={An efficient variable selection approach for analyzing designed experiments},
  author={Yuan, Ming and Joseph, V. Roshan and Lin, Ying},
  journal={Technometrics},
  volume={49},
  number={4},
  pages={430--439},
  year={2007}
}

@article{chernozhukov2014gaussian,
  title={Gaussian approximation of suprema of empirical processes},
  author={Chernozhukov, Victor and Chetverikov, Denis and Kato, Kengo},
  journal={The Annals of Statistics},
  volume={42},
  number={4},
  pages={1564--1597},
  year={2014},
  doi={10.1214/14-AOS1230}
}

@article{bail2018exposure,
  title={Exposure to opposing views on social media can increase political polarization},
  author={Bail, Christopher A and Argyle, Laura P and Brown, Taylor W and Bumpus, John P and Chen, Haohan and Hunzaker, M Brooke and Lee, Jaemin and Mann, Marcus and Merhout, Friedolin and Volfovsky, Alexander},
  journal={Proceedings of the National Academy of Sciences},
  volume={115},
  number={37},
  pages={9216--9221},
  year={2018},
  publisher={National Acad Sciences}
}

@article{bond2012experiment,
  title={A 61-million-person experiment in social influence and political mobilization},
  author={Bond, Robert M and Fariss, Christopher J and Jones, Jason J and Kramer, Adam DI and Marlow, Cameron and Settle, Jaime E and Fowler, James H},
  journal={Nature},
  volume={489},
  number={7415},
  pages={295--298},
  year={2012},
  publisher={Nature Publishing Group}
}

@article{forastiere2021identification,
  title={Identification and estimation of treatment and interference effects in observational studies on networks},
  author={Forastiere, Laura and Airoldi, Edoardo M and Mealli, Fabrizia},
  journal={Journal of the American Statistical Association},
  volume={116},
  number={534},
  pages={901--918},
  year={2021},
  publisher={Taylor \& Francis}
}

@article{exposure2017,
  title={Estimating average causal effects under general interference, with application to a social network experiment},
  author={Peter M. Aronow and Cyrus Samii},
  journal={The Annals of Applied Statistic},
  volume={11},
  number={4},
  pages={1912-1947},
  year={2017},
  publisher={Wiley Online Library}
}

@inproceedings{ma2021causal,
  title={Causal Inference under Networked Interference and Intervention Policy Enhancement},
  author={Ma, Yunpu and Tresp, Volker},
  booktitle={Proceedings of The 24th International Conference on Artificial Intelligence and Statistics},
  pages={3700--3708},
  year={2021},
  organization={PMLR}
}

@inproceedings{chen2024doubly,
  title={Doubly Robust Causal Effect Estimation under Networked Interference via Targeted Learning},
  author={Chen, Weilin and Cai, Ruichu and Yang, Zeqin and Qiao, Jie and Yan, Yuguang and Li, Zijian and Hao, Zhifeng},
  booktitle={Proceedings of the 41st International Conference on Machine Learning},
  pages={6457--6485},
  year={2024},
  organization={PMLR}
}

@inproceedings{jiang2022estimating,
  title={Estimating Causal Effects on Networked Observational Data via Representation Learning},
  author={Jiang, Song and Li, Yaliang and Gao, Jing and Zhang, Aidong},
  booktitle={Proceedings of the 31st ACM International Conference on Information \& Knowledge Management},
  pages={6457--6466},
  year={2022},
  organization={ACM}
}

@article{nie2021quasi,
  title={Quasi-oracle estimation of heterogeneous treatment effects},
  author={Nie, Xinkun and Wager, Stefan},
  journal={Biometrika},
  volume={108},
  number={2},
  pages={299--319},
  year={2021},
  publisher={Oxford University Press}
}

@article{khatami2024graph,
  title={Graph Machine Learning Based Doubly Robust Estimator for Network Causal Effects},
  author={Khatami, Seyedeh Baharan and Parikh, Harsh and Chen, Haowei and Roy, Sudeepa and Salimi, Babak},
  journal={arXiv preprint arXiv:2403.11332},
  year={2024}
}

@inproceedings{guo2020learning,
  title={Learning Individual Causal Effects from Networked Observational Data},
  author={Guo, Ruocheng and Li, Jundong and Liu, Huan},
  booktitle={Proceedings of the 13th International Conference on Web Search and Data Mining},
  pages={232--240},
  year={2020},
  organization={ACM}
}

@article{ogburn2024causal,
  title={Causal inference for social network data},
  author={Ogburn, Elizabeth L and Sofrygin, Oleg and Diaz, Ivan and Van der Laan, Mark J},
  journal={Journal of the American Statistical Association},
  volume={119},
  number={545},
  pages={597--611},
  year={2024},
  publisher={Taylor \& Francis}
}

@article{leung2020treatment,
  title={Treatment and spillover effects under network interference},
  author={Leung, Michael P},
  journal={Review of Economics and Statistics},
  volume={102},
  number={2},
  pages={368--380},
  year={2020},
  publisher={MIT Press One Rogers Street, Cambridge, MA 02142-1209, USA journals-info~…}
}

@article{cristali2022using,
  title={Using embeddings for causal estimation of peer influence in social networks},
  author={Cristali, Irina and Veitch, Victor},
  journal={Advances in Neural Information Processing Systems},
  volume={35},
  pages={15616--15628},
  year={2022}
}

@article{velivckovic2017graph,
  title={Graph attention networks},
  author={Veli{\v{c}}kovi{\'c}, Petar and Cucurull, Guillem and Casanova, Arantxa and Romero, Adriana and Lio, Pietro and Bengio, Yoshua},
  journal={arXiv preprint arXiv:1710.10903},
  year={2017}
}

@article{tchetgen2012causal,
  title={On causal inference in the presence of interference},
  author={Tchetgen, Eric J Tchetgen and VanderWeele, Tyler J},
  journal={Statistical methods in medical research},
  volume={21},
  number={1},
  pages={55--75},
  year={2012},
  publisher={SAGE Publications Sage UK: London, England}
}

@inproceedings{lin2023estimating,
  title={Estimating treatment effects under heterogeneous interference},
  author={Lin, Xiaofeng and Zhang, Guoxi and Lu, Xiaotian and Bao, Han and Takeuchi, Koh and Kashima, Hisashi},
  booktitle={Joint European Conference on Machine Learning and Knowledge Discovery in Databases},
  pages={576--592},
  year={2023},
  organization={Springer}
}

@article{huang2023modeling,
  title={Modeling Interference for Individual Treatment Effect Estimation from Networked Observational Data},
  author={Huang, Qiang and Ma, Jing and Li, Jundong and Guo, Ruocheng and Sun, Huiyan and Chang, Yi},
  journal={ACM Transactions on Knowledge Discovery from Data},
  volume={18},
  number={3},
  pages={1--21},
  year={2023},
  publisher={ACM New York, NY}
}

@inproceedings{wu2025causal,
  title={Causal Graph Transformer for Treatment Effect Estimation Under Unknown Interference},
  author={Wu, Anpeng and Qiu, Haiyi and Chen, Zhengming and Li, Zijian and Xiong, Ruoxuan and Wu, Fei and Zhang, Kun},
  booktitle={Proceedings of the 13th International Conference on Learning Representations (ICLR)},
  year={2025},
  url={https://openreview.net/forum?id=foQ4AeEGG7}
}

@article{foster2023orthogonal,
  title={Orthogonal statistical learning},
  author={Foster, Dylan J and Syrgkanis, Vasilis},
  journal={The Annals of Statistics},
  volume={51},
  number={3},
  pages={879--908},
  year={2023},
  publisher={Institute of Mathematical Statistics}
}

@article{bargagli2025heterogeneous,
  title={Heterogeneous treatment and spillover effects under clustered network interference},
  author={Bargagli-Stoffi, Falco J and Tort{\`u}, Costanza and Forastiere, Laura},
  journal={The Annals of Applied Statistics},
  volume={19},
  number={1},
  pages={28--55},
  year={2025},
  publisher={Institute of Mathematical Statistics}
}

@article{dgcnn,
  title={Dynamic Graph CNN for Learning on Point Clouds},
  author={Wang, Yue and Sun, Yongbin and Liu, Ziwei and Sarma, Sanjay E. and Bronstein, Michael M. and Solomon, Justin M.},
  journal={ACM Transactions on Graphics (TOG)},
  volume={38},
  number={5},
  pages={146},
  year={2019},
  publisher={ACM}
}

@article{CFR,
  title={Generalization Bounds and Representation Learning for Estimation of Potential Outcomes and Causal Effects},
  author={Johansson, Fredrik D. and Shalit, Uri and Kallus, Nathan and Sontag, David},
  journal={Journal of Machine Learning Research},
  volume={23},
  number={166},
  pages={1--48},
  year={2022},
  publisher={JMLR.org}
}

@inproceedings{ND,
  title={Learning Individual Causal Effects from Networked Observational Data},
  author={Guo, Ruocheng and Li, Jundong and Liu, Huan},
  booktitle={Proceedings of the 13th International Conference on Web Search and Data Mining (WSDM)},
  pages={232--240},
  year={2020},
  publisher={ACM}
}

@article{karypis1998fast,
  title={A fast and high quality multilevel scheme for partitioning irregular graphs},
  author={Karypis, George and Kumar, Vipin},
  journal={SIAM Journal on Scientific Computing},
  volume={20},
  number={1},
  pages={359--392},
  year={1998},
  publisher={SIAM}
}

@inproceedings{tang2023towards,
  title={Towards understanding generalization of graph neural networks},
  author={Tang, Huayi and Liu, Yong},
  booktitle={International Conference on Machine Learning},
  pages={33674--33719},
  year={2023},
  organization={PMLR}
}

@inproceedings{garg2020generalization,
  title={Generalization and representational limits of graph neural networks},
  author={Garg, Vikas and Jegelka, Stefanie and Jaakkola, Tommi},
  booktitle={International conference on machine learning},
  pages={3419--3430},
  year={2020},
  organization={PMLR}
}

@article{vasileiou2025survey,
  title={Survey on Generalization Theory for Graph Neural Networks},
  author={Vasileiou, Antonis and Jegelka, Stefanie and Levie, Ron and Morris, Christopher},
  journal={arXiv preprint arXiv:2503.15650},
  year={2025}
}

@article{lee2019stable,
  title={Stable limit theorems for empirical processes under conditional neighborhood dependence},
  author={Lee, Ji Hyung and Song, Kyungchul},
  journal={Bernoulli},
  volume={25},
  number={2},
  pages={1189--1224},
  year={2019},
  publisher={JSTOR}
}

@article{ou2024covering,
  title={Covering numbers for deep relu networks with applications to function approximation and nonparametric regression},
  author={Ou, Weigutian and B{\"o}lcskei, Helmut},
  journal={arXiv preprint arXiv:2410.06378},
  year={2024}
}

\clearpage
\section*{\centering Appendix Contents}
\addcontentsline{toc}{section}{Appendix Contents}

\vspace{1.5em}

\begin{center}
\Large
\renewcommand{\arraystretch}{1.35}
\begin{tabular}{ll}
\textbf{A} & Experiment Setup \\
\textbf{B} & Related Work \\
\textbf{C} & Additional Experiments \\
\textbf{D} & Discussion of the Additive Local Interference Assumption \\

\textbf{E} & Technical Preliminaries and Notation \\
\textbf{F} & Proof of Identification Results (Theorem~1) \\
\textbf{G} & Proof of the Orthogonal Estimation Bound (Theorem~2) \\
\textbf{H} & Empirical Process Control under Conditional Neighborhood Dependence \\
\textbf{I} & Entropy Bound for the Attention-Based ReLU Interference Class \\
\end{tabular}
\end{center}

\clearpage
\appendix

\section{Experiment setup}\label{Experiment_setup}
\subsection{Data generation process}
We describe the generation of the binary treatment $T_i$ and observed outcome
$Y_i$ used in the semi-synthetic experiments. Consistent with the main text, we
generate
\[
T_i\sim
\operatorname{Bernoulli}\!\left(
\sigma\!\left(f_T(\bm X_{\bar{\mathcal N}_1(i)})\right)
\right),
\qquad
Y_i
=
f_0(\bm X_{\bar{\mathcal N}_1(i)})
+
T_i\tau_i^\star
+
\sum_{j\in\mathcal N_1(i)}T_j\tau_{ij}^\star
+
\epsilon_i .
\]
Here $\sigma(\cdot)$ is the sigmoid function and $\epsilon_i$ is random noise.
Let $\bar{\bm X}_i=|\mathcal N_1(i)|^{-1}\sum_{j\in\mathcal N_1(i)}\bm X_j$
denote the average neighbor feature vector. We use
\[
f_T(\bm X_{\bar{\mathcal N}_1(i)})
=
\bm X_i^\top \bm w_1+\bar{\bm X}_i^\top \bm v_1,
\qquad
f_0(\bm X_{\bar{\mathcal N}_1(i)})
=
\bm X_i^\top \bm w_2+\bar{\bm X}_i^\top \bm v_2,
\]
where $\bm w_1,\bm v_1,\bm w_2,\bm v_2\sim\mathcal N(\bm 0,I)$.

We construct the ground-truth effects from raw scores $r_{ij}$ and attention
weights. For the semi-synthetic real-network experiments in Section~6.1, the
direct and spillover effects are generated using the same raw-score function:
$r_{ii}$ and $\{r_{ij}:j\in\mathcal N_1(i)\}$ are normalized together over the
closed one-hop neighborhood. Specifically,
\[
\alpha_{ij}
=
\frac{\exp(\beta |r_{ij}|)}
{\sum_{k\in\bar{\mathcal N}_1(i)}\exp(\beta |r_{ik}|)},
\qquad
\tau_i^\star=\alpha_{ii}r_{ii},
\qquad
\tau_{ij}^\star=\alpha_{ij}r_{ij},\quad j\in\mathcal N_1(i).
\]
This mirrors the benchmark setting where direct and spillover components are
generated from a shared effect mechanism.

For the synthetic edge-level experiments in Section~6.2, we generate the direct
and spillover components separately. The direct effect is produced by a node-level
function $r_{ii}=W_{\mathrm{dir}}(\bm X_i)$, while the pairwise raw score is
$r_{ij}=W_{\mathrm{sp}}(\bm X_i,\bm X_j)$. The neighbor attention weights are then
normalized only over $\mathcal N_1(i)$:
\[
\alpha_{ij}
=
\frac{\exp(\beta |r_{ij}|)}
{\sum_{k\in\mathcal N_1(i)}\exp(\beta |r_{ik}|)},
\qquad
\tau_i^\star=r_{ii},
\qquad
\tau_{ij}^\star=\alpha_{ij}r_{ij},\quad j\in\mathcal N_1(i).
\]
In this setting, $W_{\mathrm{dir}}$ and $W_{\mathrm{sp}}$ are implemented as
separate MLPs, so the ego direct effect is not normalized together with neighbor
spillovers.

For the real-network experiments, we consider the following choices of the
pairwise raw-score function $r_{ij}=W(\bm X_i,\bm X_j)$:
\[
\begin{aligned}
\textsc{RBF:}\quad
&r_{ij}=\exp\!\left(-\frac{1}{2}\|\bm X_i-\bm X_j\|_2^2\right),\\
\textsc{Cosine:}\quad
&r_{ij}=\frac{\bm X_i^\top\bm X_j}{\|\bm X_i\|_2\|\bm X_j\|_2},\\
\textsc{Homo:}\quad
&r_{ij}=0.9\ \text{for }i\neq j,\qquad r_{ii}=1,\\
\textsc{One-way:}\quad
&r_{ij}=h(\bm X_j),
\end{aligned}
\]
where $h(\cdot)$ is a nonlinear scalar function of the treated neighbor's
features. Thus, the RBF and Cosine settings depend on both the focal node and the
neighbor, the One-way setting depends only on the neighbor, and the Homo setting
corresponds to homogeneous spillover strength.

\subsection{Details of the model structure and fitting procedure}

\paragraph{Nuisance model architectures.}
Our propensity and conditional-mean modules use a GAT-style aggregator with flexible dot-product attention to capture one-hop neighbor interactions. We first project the node features \(X\in\mathbb{R}^{N\times d}\) into a hidden representation \(H\) through a learnable linear layer. We then compute raw attention scores as
\[
E=\mathrm{LeakyReLU}\!\left(a_{\mathrm{scale}}HH^\top\right),
\]
mask the scores using the normalized adjacency matrix \(A_{\mathrm{norm}}\), and apply a row-wise softmax to obtain attention coefficients \(\alpha\). The neighborhood representation is then computed by
\[
Z=\alpha H.
\]
For each node, we concatenate its original features \(X\) with the aggregated features \(Z\), forming a \(2d\)-dimensional representation. This representation is passed through an MLP with one hidden layer of 64 units and ReLU activation, followed by a final linear projection to produce either the scalar propensity estimate or the scalar conditional-mean estimate. This shared architecture integrates local node attributes and relational context in a single end-to-end module.

\paragraph{Interference model architecture.}
We specify the interference model to be consistent with the outcome-generation process in both the semi-synthetic and synthetic experiments. Specifically, we use a two-layer MLP with 64 hidden units per layer and ReLU activations to estimate the raw pairwise influence function \(W(\cdot)\). We also include a learnable parameter \(\hat{\beta}\) to capture the temperature parameter in the attention weights. 

\paragraph{Sample splitting via graph partitioning.} Theoretically, the cross-fitting procedure requires two independent subsets of units, \(\bm V_1\) and \(\bm V_2\), separated by a margin in graph distance. This involves discarding all responses \(\{Y_i\}\) located in between, potentially reducing training efficiency. In practice, we replace the margin split with a balanced graph partitioning step using METIS algorithm \cite{karypis1998fast}. We specify $S\ge2$ roughly equal-sized parts $\{\bm V_s\}_{s=1}^S$ that minimize edge-cuts between clusters. In steps~2--3, we perform cross-fitting by training nuisance components on \(\bm V_s\) and estimating the interference model on the complement \(\bigcup_{s' \neq s}\bm V_{s'}\), iterating through all subsets \(s\). Data splitting via graph partitioning leverages all observed outcomes \(\{Y_i\}\) and ensures balanced sample sizes for nuisance and causal-effect estimation. Empirically, this approach scales effectively and provides stable performance in our experiments. We report results with \(S = 5\). Additional results with varying numbers of partitions are provided in \ref{sec:partition}.

\paragraph{Compute.} The experimental model-fitting procedure does not involve large-scale matrix computations, excessively large models, or very large sample sizes, so all experiments were run locally on a MacBook with GPU support.

\subsection{More on real network experiment}
\paragraph{Datasets} We use two widely used attributed social network benchmarks, BlogCatalog (BC) and Flickr \citep{asuSocialComputingData,huang2017label}. BlogCatalog (BC) is an online blogging platform where each user is represented as a node in a social network, with edges indicating social links between users. Node features are bag-of-words vectors derived from keywords in user profiles. Flickr is a photo-sharing network in which nodes represent users and edges capture social relationships based on shared metadata. User features consist of tag-based indicators reflecting individual interests.

BlogCatalog contains \(n=5196\) nodes with 8189-dimensional features, while Flickr contains \(n=7192\) nodes with 12047-dimensional features. For both datasets, we apply Linear Discriminant Analysis (LDA) to reduce the feature dimension to \(d=5\), and evaluate performance using equal-sized training, validation, and test splits of the network.

\subsection{More on synthetic network experiment}\label{appendix_SBM}
\paragraph{Network generation.}
We generate synthetic networks using a stochastic block model (SBM). For the training network, we use \(n=3000\), within-community connection probability \(p_{\mathrm{in}}=0.00333\), and between-community connection probability \(p_{\mathrm{out}}=0.001\). For the evaluation network, we use \(n=1000\), \(p_{\mathrm{in}}=0.01\), and \(p_{\mathrm{out}}=0.003\). Both networks follow the same data-generating process. Given each node's community label \(k\), we sample embeddings
\[
X_i \sim \mathcal{N}(\mu_k,\sigma^2 I),
\qquad
\sigma^2=0.25,
\]
where the unit-norm centroids \(\mu_k\in\mathbb{R}^d\), with \(d=5\), are chosen to satisfy
\[
\mu_1^\top \mu_2=-0.75,
\qquad
\mu_1^\top \mu_3=-0.30,
\qquad
\mu_2^\top \mu_3=-0.05.
\]
To generate outcomes, we set \(W(\cdot)\) to the cosine similarity and use the same summarization functions \(f_T, f_1, f_2\) as in Section~5.1. For the confidence-interval experiment in Section~\ref{sec:bootstrap-ci-exp}, we fix the attention temperature at \(\beta=0\).

\paragraph{Curvature approximation for Hessian computation.}
In practice, since $\hat\theta$ is obtained by stochastic optimization and the
second-stage Hessian used in the one-step approximation in \ref{one_step} may be ill-conditioned, we use the stabilized update
\[
\hat\theta_{\mathrm{IJ}}^{(b)}
=
\hat\theta
-
(\hat H+\lambda I)^{-1}
\left\{
\nabla_\theta L_{\tilde n}(\hat\theta,\xi^{(b)})
-
\nabla_\theta L_{\tilde n}(\hat\theta,\mathbf 1)
\right\},
\]
where $\lambda>0$ is a damping parameter. Rather than forming the full Hessian
explicitly, we use a Gauss--Newton curvature approximation for the residualized
squared loss. If $r(\theta)$ denotes the vector of second-stage residuals and
$J=\partial r(\theta)/\partial\theta$, then
\[
\hat H \approx \frac{2}{\tilde n}J^\top J .
\]
This approximation is positive semidefinite and is applied through
Hessian-vector products, so the damped linear system
$(\hat H+\lambda I)\delta=g$ can be solved matrix-free by conjugate gradient.
For numerical checks, we also compare against Lanczos low-rank and dense direct
solvers in small-scale settings. In the experiments in
Section~\ref{sec:bootstrap-ci-exp}, we use $\lambda=0.0004$ for oracle nuisance
and $\lambda=0.0006$ for estimated nuisance.

\paragraph{Multiplier weights.}
In all bootstrap experiments, we use normal multiplier weights. Specifically,
for each bootstrap replicate $b$, we draw independent weights
$\xi_i^{(b)}\sim N(1,1)$ for units $i\in\bm V_1\cup\bm V_2$. This choice
satisfies $\E[\xi_i^{(b)}]=1$ and $\mathrm{Var}(\xi_i^{(b)})=1$, matching the
weighted multiplier bootstrap formulation used in the main text.

\paragraph{Metrics definition.}

We provide formal definitions for the two metrics used in influential neighbour detection in section \ref{sec:edge-interpretability}.

\begin{itemize}
    \item Recall@K=20\%: For each ego node \(i\), let \(K_i = \lceil 0.2\,|\mathcal{N}_1(i)| \rceil\) be the number of top neighbours under consideration. Let \(\mathcal{T}_i^{(K)}\) denote the set of true top-\(K_i\) neighbours ranked by \(|\tau_{ij}|\), and \(\mathcal{P}_i^{(K)}\) the set of predicted top-\(K_i\) neighbours ranked by \(|\hat{\tau}_{ij}|\). The per node recall is defined as $\text{Recall}_i^{(K)} = \frac{|\mathcal{T}_i^{(K)} \cap \mathcal{P}_i^{(K)}|}{K_i}$
    and the overall recall @ 20\% is calculated by averaging across all nodes: $\mathrm{Recall@20}\% = \frac{1}{n} \sum_{i=1}^n \text{Recall}_i^{(K)}$.
    \item NDCG@K=20\%: To define NDCG@20\%, we first assign an exponential relevance score to each true top neighbour \(j \in \mathcal{T}_i^{(K)}\) based on its true rank: $
    \mathrm{rel}_i(j) = 2^{K_i - \mathrm{TrueRank}_i(j)} - 1$.  We then compute the discounted cumulative gain (DCG) over the model’s predicted top-\(K_i\) order \((j_{(1)}, \ldots, j_{(K_i)})\) as $\mathrm{DCG}_i = \sum_{r=1}^{K_i} \frac{\mathrm{rel}_i(j_{(r)})}{\log_2(r + 1)}$. The ideal DCG, denoted \(\mathrm{IDCG}_i\), is computed by summing the same relevance scores in descending order. The normalized DCG for node \(i\) is $\mathrm{NDCG}_i = \frac{\mathrm{DCG}_i}{\mathrm{IDCG}_i}, \quad \text{and} \quad \mathrm{NDCG@20\%} = \frac{1}{n} \sum_{i=1}^n \mathrm{NDCG}_i$.
\end{itemize}

\section{Related Work}
Causal inference under interference has developed along two broad lines. Classical statistical work studies identification and inference for treatment and spillover effects under exposure mappings, locality, or partial-interference assumptions, with primary emphasis on population- or cluster-level estimands \citep{hudgens2008toward,aronow2017estimating,forastiere2021identification,leung2020treatment,leung2022causal,ogburn2024causal,bargagli2025heterogeneous}. More recent representation-learning methods target observational network data with high-dimensional covariates and unknown interference. In our experimental suite, CFR~\citep{CFR} is adapted as a standard non-network representation baseline, ND~\citep{guo2020learning} and NetEst~\citep{jiang2022estimating} learn network-aware representations to mitigate hidden confounding, CauGramer~\citep{wu2025causal} uses a graph transformer to infer interference representations under unknown interference, GDML~\citep{khatami2024graph} and TNet~\citep{chen2024doubly} pursue doubly robust adjustment for direct and peer effects, and EdgeConv~\citep{dgcnn} provides a flexible graph architecture for heterogeneous neighbor weighting. Closely related recent methods include SPNet\citep{huang2024spnet}, which combines GCN-based confounder modeling with masked attention for interference, DWR\citep{zhao2024dwr}, which jointly learns interference attention weights and sample weights through bi-level optimization, HINITE\citep{fan2026idenet}, which models heterogeneous multi-view interference, IDENet, which uses identifiable disentangled representations under network interference, and NDR\citep{liu2023ndr}, which develops a nonparametric doubly robust estimator under general interference. Our work is positioned between the flexible but largely black-box representation-learning literature and the more principled doubly robust literature.

\section{Additional Experiments}
\subsection{Experiments on BC dataset}\label{BC_result}

Following the same data-generating process as in Table~\ref{Flickr} for the Flickr dataset, we report the corresponding results on the BC dataset in Table~\ref{BC}. The results closely mirror those reported for Flickr in the main text: our proposed estimator matches DBML and TNet on population-level metrics (ADE and ASE), while achieving substantially lower node-level errors (IDE and ISE). The gains become more pronounced at higher temperatures $\beta$. The consistently small gap between the proposed estimator and its oracle counterpart further suggests that the orthogonal learning framework is robust across different network structures.

\begin{table}[H]
\centering
\scriptsize
\caption{Out-of-sample results on the \textbf{BC} dataset using \textbf{Cosine} and \textbf{RBF} kernels with varying temperatures $\beta\in\{0,1,5,10\}$. The top result is highlighted in bold, and the runner-up is underlined.}
\label{BC}
\resizebox{\textwidth}{!}{
\begin{tabular}{|cccccccccc|cc|}
\hline
Interference & Temp.\ & Effect &
CFR & EdgeConv & ND & Netest & Tnet & Caugamer & GDML &
$\textrm{Proposed}_{\mathrm{est}}$ &
$\textrm{Proposed}_{\mathrm{oracle}}$\\
\hline
\multirow{16}{*}{\textbf{Cosine}}
  & \multirow{4}{*}{0}
      & {\tiny ADE} & {\tiny $0.1550\std{0.094}$} & {\tiny $0.0732\std{0.011}$} & {\tiny $0.1913\std{0.062}$} & {\tiny $0.0574\std{0.037}$} & {\tiny $0.0181\std{0.031}$} & {\tiny $0.0203\std{0.019}$} & {\tiny $\underline{0.0011}\std{0.017}$} & {\tiny $\mathbf{0.0004}\std{0.009}$} & {\tiny $0.0022\std{0.005}$} \\
  & & {\tiny ASE} & {\tiny $0.1005\std{0.049}$} & {\tiny $0.0992\std{0.028}$} & {\tiny $0.1116\std{0.061}$} & {\tiny $0.1126\std{0.134}$} & {\tiny $\mathbf{0.0057}\std{0.003}$} & {\tiny $0.2342\std{0.014}$} & {\tiny $0.1293\std{0.253}$} & {\tiny $\underline{0.0075}\std{0.029}$} & {\tiny $0.0122\std{0.015}$} \\
  & & {\tiny IDE} & {\tiny $0.2059\std{0.028}$} & {\tiny $0.1072\std{0.016}$} & {\tiny $0.2299\std{0.057}$} & {\tiny $0.1317\std{0.027}$} & {\tiny $0.1229\std{0.019}$} & {\tiny $0.1395\std{0.043}$} & {\tiny $\mathbf{0.0163}\std{0.011}$} & {\tiny $\underline{0.0238}\std{0.011}$} & {\tiny $0.0185\std{0.005}$} \\
  & & {\tiny ISE} & {\tiny $0.3327\std{0.021}$} & {\tiny $\underline{0.2501}\std{0.009}$} & {\tiny $0.3001\std{0.023}$} & {\tiny $0.3053\std{0.071}$} & {\tiny $0.2598\std{0.002}$} & {\tiny $0.3477\std{0.009}$} & {\tiny $0.3436\std{0.128}$} & {\tiny $\mathbf{0.0758}\std{0.024}$} & {\tiny $0.0568\std{0.010}$} \\
\cline{2-12}
  & \multirow{4}{*}{1}
      & {\tiny ADE} & {\tiny $0.2337\std{0.062}$} & {\tiny $0.1051\std{0.015}$} & {\tiny $0.2207\std{0.120}$} & {\tiny $0.0714\std{0.048}$} & {\tiny $0.0280\std{0.046}$} & {\tiny $0.0228\std{0.020}$} & {\tiny $\mathbf{0.0004}\std{0.014}$} & {\tiny $\underline{0.0018}\std{0.003}$} & {\tiny $0.0027\std{0.005}$} \\
  & & {\tiny ASE} & {\tiny $0.1056\std{0.057}$} & {\tiny $0.1635\std{0.014}$} & {\tiny $0.1369\std{0.105}$} & {\tiny $0.1919\std{0.164}$} & {\tiny $\mathbf{0.0058}\std{0.007}$} & {\tiny $0.3742\std{0.020}$} & {\tiny $0.0249\std{0.132}$} & {\tiny $\underline{0.0103}\std{0.019}$} & {\tiny $0.0102\std{0.021}$} \\
  & & {\tiny IDE} & {\tiny $0.2810\std{0.058}$} & {\tiny $0.1523\std{0.021}$} & {\tiny $0.2747\std{0.102}$} & {\tiny $0.1603\std{0.029}$} & {\tiny $0.1528\std{0.035}$} & {\tiny $0.1715\std{0.018}$} & {\tiny $\underline{0.0442}\std{0.009}$} & {\tiny $\mathbf{0.0341}\std{0.005}$} & {\tiny $0.0245\std{0.004}$} \\
  & & {\tiny ISE} & {\tiny $0.3572\std{0.009}$} & {\tiny $0.2917\std{0.009}$} & {\tiny $0.3290\std{0.037}$} & {\tiny $0.3598\std{0.112}$} & {\tiny $\underline{0.2405}\std{0.010}$} & {\tiny $0.4629\std{0.013}$} & {\tiny $0.2871\std{0.020}$} & {\tiny $\mathbf{0.0771}\std{0.006}$} & {\tiny $0.0625\std{0.006}$} \\
\cline{2-12}
  & \multirow{4}{*}{5}
      & {\tiny ADE} & {\tiny $0.2053\std{0.054}$} & {\tiny $0.2223\std{0.022}$} & {\tiny $0.1442\std{0.097}$} & {\tiny $0.0738\std{0.025}$} & {\tiny $0.0521\std{0.070}$} & {\tiny $0.0167\std{0.012}$} & {\tiny $\underline{0.0082}\std{0.026}$} & {\tiny $\mathbf{0.0062}\std{0.007}$} & {\tiny $0.0059\std{0.001}$} \\
  & & {\tiny ASE} & {\tiny $0.0351\std{0.018}$} & {\tiny $0.2331\std{0.030}$} & {\tiny $0.0903\std{0.123}$} & {\tiny $0.1914\std{0.144}$} & {\tiny $\underline{0.0116}\std{0.005}$} & {\tiny $0.5420\std{0.095}$} & {\tiny $0.0332\std{0.194}$} & {\tiny $\mathbf{0.0104}\std{0.023}$} & {\tiny $0.0087\std{0.010}$} \\
  & & {\tiny IDE} & {\tiny $0.3630\std{0.039}$} & {\tiny $0.3327\std{0.026}$} & {\tiny $0.3180\std{0.060}$} & {\tiny $0.2690\std{0.016}$} & {\tiny $0.3015\std{0.049}$} & {\tiny $0.2634\std{0.032}$} & {\tiny $\underline{0.2034}\std{0.023}$} & {\tiny $\mathbf{0.0522}\std{0.008}$} & {\tiny $0.0346\std{0.006}$} \\
  & & {\tiny ISE} & {\tiny $0.3593\std{0.012}$} & {\tiny $0.3854\std{0.022}$} & {\tiny $0.3399\std{0.054}$} & {\tiny $0.3936\std{0.071}$} & {\tiny $\underline{0.2939}\std{0.007}$} & {\tiny $0.6197\std{0.073}$} & {\tiny $0.2993\std{0.069}$} & {\tiny $\mathbf{0.0645}\std{0.010}$} & {\tiny $0.0496\std{0.010}$} \\
\cline{2-12}
  & \multirow{4}{*}{10}
      & {\tiny ADE} & {\tiny $0.1609\std{0.083}$} & {\tiny $0.2701\std{0.039}$} & {\tiny $0.1404\std{0.099}$} & {\tiny $\underline{0.0442}\std{0.039}$} & {\tiny $0.0505\std{0.069}$} & {\tiny $0.0348\std{0.039}$} & {\tiny $0.0139\std{0.031}$} & {\tiny $\mathbf{0.0007}\std{0.007}$} & {\tiny $0.0052\std{0.003}$} \\
  & & {\tiny ASE} & {\tiny $0.0454\std{0.056}$} & {\tiny $0.2415\std{0.039}$} & {\tiny $0.0805\std{0.077}$} & {\tiny $0.1620\std{0.124}$} & {\tiny $\mathbf{0.0034}\std{0.002}$} & {\tiny $0.4977\std{0.136}$} & {\tiny $0.0436\std{0.184}$} & {\tiny $\underline{0.0068}\std{0.023}$} & {\tiny $0.0067\std{0.005}$} \\
  & & {\tiny IDE} & {\tiny $0.3644\std{0.040}$} & {\tiny $0.3824\std{0.044}$} & {\tiny $0.3359\std{0.057}$} & {\tiny $0.2845\std{0.018}$} & {\tiny $0.3128\std{0.016}$} & {\tiny $0.3170\std{0.047}$} & {\tiny $\underline{0.2343}\std{0.022}$} & {\tiny $\mathbf{0.0583}\std{0.008}$} & {\tiny $0.0409\std{0.004}$} \\
  & & {\tiny ISE} & {\tiny $0.3596\std{0.031}$} & {\tiny $0.4012\std{0.018}$} & {\tiny $0.3247\std{0.029}$} & {\tiny $0.3911\std{0.054}$} & {\tiny $0.2958\std{0.007}$} & {\tiny $0.5908\std{0.099}$} & {\tiny $\underline{0.2953}\std{0.068}$} & {\tiny $\mathbf{0.0663}\std{0.005}$} & {\tiny $0.0434\std{0.006}$} \\
\cline{1-12}
\multirow{16}{*}{\textbf{RBF}}
  & \multirow{4}{*}{0}
      & {\tiny ADE} & {\tiny $0.1574\std{0.051}$} & {\tiny $0.0732\std{0.011}$} & {\tiny $0.1553\std{0.100}$} & {\tiny $0.0640\std{0.039}$} & {\tiny $0.0177\std{0.036}$} & {\tiny $0.0394\std{0.059}$} & {\tiny $\underline{0.0080}\std{0.020}$} & {\tiny $\mathbf{0.0076}\std{0.006}$} & {\tiny $0.0082\std{0.006}$} \\
  & & {\tiny ASE} & {\tiny $0.0843\std{0.028}$} & {\tiny $0.1302\std{0.019}$} & {\tiny $0.1427\std{0.109}$} & {\tiny $0.1270\std{0.138}$} & {\tiny $\underline{0.0041}\std{0.005}$} & {\tiny $0.2854\std{0.041}$} & {\tiny $0.0974\std{0.318}$} & {\tiny $\mathbf{0.0025}\std{0.021}$} & {\tiny $0.0073\std{0.020}$} \\
  & & {\tiny IDE} & {\tiny $0.2065\std{0.043}$} & {\tiny $0.1084\std{0.017}$} & {\tiny $0.2075\std{0.070}$} & {\tiny $0.1334\std{0.031}$} & {\tiny $0.1283\std{0.028}$} & {\tiny $0.1567\std{0.045}$} & {\tiny $\mathbf{0.0168}\std{0.021}$} & {\tiny $\underline{0.0236}\std{0.008}$} & {\tiny $0.0202\std{0.005}$} \\
  & & {\tiny ISE} & {\tiny $0.2614\std{0.012}$} & {\tiny $0.2455\std{0.016}$} & {\tiny $0.2811\std{0.054}$} & {\tiny $0.2808\std{0.080}$} & {\tiny $\underline{0.2173}\std{0.005}$} & {\tiny $0.3518\std{0.033}$} & {\tiny $0.3160\std{0.194}$} & {\tiny $\mathbf{0.0690}\std{0.010}$} & {\tiny $0.0560\std{0.013}$} \\
\cline{2-12}
  & \multirow{4}{*}{1}
      & {\tiny ADE} & {\tiny $0.2140\std{0.072}$} & {\tiny $0.1226\std{0.015}$} & {\tiny $0.1964\std{0.144}$} & {\tiny $0.0704\std{0.041}$} & {\tiny $0.0473\std{0.052}$} & {\tiny $\underline{0.0033}\std{0.004}$} & {\tiny $\mathbf{0.0000}\std{0.024}$} & {\tiny $0.0135\std{0.005}$} & {\tiny $0.0091\std{0.003}$} \\
  & & {\tiny ASE} & {\tiny $0.0514\std{0.027}$} & {\tiny $0.1921\std{0.026}$} & {\tiny $0.0842\std{0.070}$} & {\tiny $0.1426\std{0.139}$} & {\tiny $\mathbf{0.0147}\std{0.020}$} & {\tiny $0.3942\std{0.041}$} & {\tiny $0.1054\std{0.184}$} & {\tiny $\underline{0.0298}\std{0.028}$} & {\tiny $0.0213\std{0.008}$} \\
  & & {\tiny IDE} & {\tiny $0.2704\std{0.064}$} & {\tiny $0.1721\std{0.021}$} & {\tiny $0.2609\std{0.121}$} & {\tiny $0.1675\std{0.028}$} & {\tiny $0.1708\std{0.033}$} & {\tiny $0.1849\std{0.064}$} & {\tiny $\underline{0.0595}\std{0.015}$} & {\tiny $\mathbf{0.0405}\std{0.007}$} & {\tiny $0.0281\std{0.004}$} \\
  & & {\tiny ISE} & {\tiny $0.2903\std{0.015}$} & {\tiny $0.2918\std{0.020}$} & {\tiny $0.2755\std{0.021}$} & {\tiny $0.3145\std{0.077}$} & {\tiny $\underline{0.2478}\std{0.005}$} & {\tiny $0.4617\std{0.031}$} & {\tiny $0.2871\std{0.060}$} & {\tiny $\mathbf{0.0800}\std{0.020}$} & {\tiny $0.0575\std{0.008}$} \\
\cline{2-12}
  & \multirow{4}{*}{5}
      & {\tiny ADE} & {\tiny $0.2172\std{0.060}$} & {\tiny $0.2760\std{0.029}$} & {\tiny $0.1549\std{0.107}$} & {\tiny $0.0553\std{0.046}$} & {\tiny $0.0647\std{0.090}$} & {\tiny $0.0623\std{0.058}$} & {\tiny $\underline{0.0225}\std{0.024}$} & {\tiny $\mathbf{0.0034}\std{0.004}$} & {\tiny $0.0050\std{0.005}$} \\
  & & {\tiny ASE} & {\tiny $0.0502\std{0.073}$} & {\tiny $0.1951\std{0.034}$} & {\tiny $0.0416\std{0.026}$} & {\tiny $0.2077\std{0.129}$} & {\tiny $\mathbf{0.0067}\std{0.006}$} & {\tiny $0.5202\std{0.065}$} & {\tiny $0.0570\std{0.136}$} & {\tiny $\underline{0.0314}\std{0.014}$} & {\tiny $0.0194\std{0.006}$} \\
  & & {\tiny IDE} & {\tiny $0.3957\std{0.045}$} & {\tiny $0.3858\std{0.030}$} & {\tiny $0.3371\std{0.052}$} & {\tiny $0.2778\std{0.016}$} & {\tiny $0.3219\std{0.049}$} & {\tiny $0.3182\std{0.023}$} & {\tiny $\underline{0.2347}\std{0.021}$} & {\tiny $\mathbf{0.0572}\std{0.009}$} & {\tiny $0.0392\std{0.005}$} \\
  & & {\tiny ISE} & {\tiny $0.3625\std{0.032}$} & {\tiny $0.3617\std{0.018}$} & {\tiny $0.3133\std{0.013}$} & {\tiny $0.4039\std{0.056}$} & {\tiny $0.2981\std{0.009}$} & {\tiny $0.5943\std{0.057}$} & {\tiny $\underline{0.2885}\std{0.032}$} & {\tiny $\mathbf{0.0838}\std{0.015}$} & {\tiny $0.0458\std{0.007}$} \\
\cline{2-12}
  & \multirow{4}{*}{10}
      & {\tiny ADE} & {\tiny $0.2220\std{0.054}$} & {\tiny $0.3326\std{0.024}$} & {\tiny $0.1335\std{0.053}$} & {\tiny $\underline{0.0431}\std{0.032}$} & {\tiny $0.0840\std{0.098}$} & {\tiny $0.0473\std{0.092}$} & {\tiny $0.0461\std{0.026}$} & {\tiny $\mathbf{0.0034}\std{0.005}$} & {\tiny $0.0063\std{0.003}$} \\
  & & {\tiny ASE} & {\tiny $0.0346\std{0.033}$} & {\tiny $0.1717\std{0.029}$} & {\tiny $0.1520\std{0.171}$} & {\tiny $0.1277\std{0.126}$} & {\tiny $\mathbf{0.0066}\std{0.006}$} & {\tiny $0.5032\std{0.039}$} & {\tiny $0.0445\std{0.094}$} & {\tiny $\underline{0.0316}\std{0.010}$} & {\tiny $0.0129\std{0.018}$} \\
  & & {\tiny IDE} & {\tiny $0.3948\std{0.028}$} & {\tiny $0.4409\std{0.025}$} & {\tiny $0.3391\std{0.024}$} & {\tiny $0.2935\std{0.016}$} & {\tiny $0.3436\std{0.054}$} & {\tiny $0.3102\std{0.050}$} & {\tiny $\underline{0.2670}\std{0.023}$} & {\tiny $\mathbf{0.0698}\std{0.013}$} & {\tiny $0.0512\std{0.006}$} \\
  & & {\tiny ISE} & {\tiny $0.3561\std{0.016}$} & {\tiny $0.3576\std{0.012}$} & {\tiny $0.3746\std{0.095}$} & {\tiny $0.3822\std{0.047}$} & {\tiny $0.3091\std{0.015}$} & {\tiny $0.5904\std{0.027}$} & {\tiny $\underline{0.2839}\std{0.015}$} & {\tiny $\mathbf{0.1178}\std{0.069}$} & {\tiny $0.0627\std{0.009}$} \\
\hline
\end{tabular}
}
\end{table}
\subsection{Experiments on additional spillover patterns}\label{heter_homo_result}
Besides the Cosine and RBF interference kernels, we further test the flexibility of the MLP structure in capturing additional spillover patterns, as detailed in Section~6.1. Table~\ref{Homo_heter} reports the causal estimation accuracy under the \textbf{One-way} and \textbf{Homogeneous} interference models on both Flickr and BlogCatalog networks. Our method consistently achieves state-of-the-art performance across most settings, highlighting the flexibility and robustness of our estimator under the varied interference patterns and network confounding structures common in existing literature.
\begin{table}[t!]
\centering
\scriptsize
\caption{\small Causal estimation performance on the BC and Flickr networks 
using non-interaction and homogeneous interference function ($\beta=0$).}
\label{Homo_heter}
\resizebox{\textwidth}{!}{
\begin{tabular}{|cccccccccc|cc|}
\hline
Dataset & Interference & Effect &
CFR & EdgeConv & ND & Netest & Tnet & Caugamer & GDML &
$\textrm{Proposed}_{\mathrm{est}}$ &
$\textrm{Proposed}_{\mathrm{oracle}}$\\
\hline
\multirow{8}{*}{BC}
  & \multirow{4}{*}{\textbf{One-way}}
    & {\tiny ADE}
      & {\tiny $0.1254\std{0.094}$}
      & {\tiny $0.0131\std{0.004}$}
      & {\tiny $0.1426\std{0.081}$}
      & {\tiny $0.0520\std{0.010}$}
      & {\tiny $\underline{0.0015}\std{0.002}$}
      & {\tiny $0.0055\std{0.005}$}
      & {\tiny $\mathbf{0.0007}\std{0.015}$}
      & {\tiny $0.0031\std{0.005}$}
      & {\tiny $0.0018\std{0.004}$} \\
  & & {\tiny ASE}
      & {\tiny $0.0901\std{0.060}$}
      & {\tiny $0.1858\std{0.019}$}
      & {\tiny $0.1493\std{0.111}$}
      & {\tiny $0.2487\std{0.069}$}
      & {\tiny $\mathbf{0.0072}\std{0.006}$}
      & {\tiny $0.1823\std{0.119}$}
      & {\tiny $0.0721\std{0.077}$}
      & {\tiny $\underline{0.0151}\std{0.029}$}
      & {\tiny $0.0020\std{0.017}$} \\
  & & {\tiny IDE}
      & {\tiny $0.1940\std{0.066}$}
      & {\tiny $\underline{0.0583}\std{0.009}$}
      & {\tiny $0.2230\std{0.058}$}
      & {\tiny $0.1465\std{0.011}$}
      & {\tiny $0.1346\std{0.006}$}
      & {\tiny $0.1416\std{0.028}$}
      & {\tiny $0.1190\std{0.013}$}
      & {\tiny $\mathbf{0.0245}\std{0.008}$}
      & {\tiny $0.0198\std{0.006}$} \\
  & & {\tiny ISE}
      & {\tiny $0.4016\std{0.035}$}
      & {\tiny $\underline{0.3003}\std{0.018}$}
      & {\tiny $0.4307\std{0.066}$}
      & {\tiny $0.4147\std{0.056}$}
      & {\tiny $0.4224\std{0.016}$}
      & {\tiny $0.4268\std{0.071}$}
      & {\tiny $0.4096\std{0.025}$}
      & {\tiny $\mathbf{0.0706}\std{0.018}$}
      & {\tiny $0.0610\std{0.008}$} \\
\cline{2-12}
  & \multirow{4}{*}{\textbf{Homo}}
    & {\tiny ADE}
      & {\tiny $0.1150\std{0.042}$}
      & {\tiny $0.9119\std{0.019}$}
      & {\tiny $0.1388\std{0.033}$}
      & {\tiny $0.0235\std{0.019}$}
      & {\tiny $0.4504\std{0.152}$}
      & {\tiny $0.2553\std{0.072}$}
      & {\tiny $\underline{0.0064}\std{0.014}$}
      & {\tiny $\mathbf{0.0002}\std{0.007}$}
      & {\tiny $0.0130\std{0.040}$} \\
  & & {\tiny ASE}
      & {\tiny $0.0636\std{0.029}$}
      & {\tiny $0.1499\std{0.042}$}
      & {\tiny $0.1039\std{0.061}$}
      & {\tiny $0.0925\std{0.031}$}
      & {\tiny $\underline{0.0532}\std{0.104}$}
      & {\tiny $0.4226\std{0.076}$}
      & {\tiny $0.1353\std{0.151}$}
      & {\tiny $\mathbf{0.0031}\std{0.013}$}
      & {\tiny $0.0182\std{0.019}$} \\
  & & {\tiny IDE}
      & {\tiny $0.1704\std{0.034}$}
      & {\tiny $0.9166\std{0.017}$}
      & {\tiny $0.1700\std{0.020}$}
      & {\tiny $0.0969\std{0.007}$}
      & {\tiny $0.5192\std{0.107}$}
      & {\tiny $0.3076\std{0.060}$}
      & {\tiny $\mathbf{0.0132}\std{0.005}$}
      & {\tiny $\underline{0.0352}\std{0.011}$}
      & {\tiny $0.0900\std{0.025}$} \\
  & & {\tiny ISE}
      & {\tiny $\underline{0.1238}\std{0.019}$}
      & {\tiny $0.2528\std{0.020}$}
      & {\tiny $0.1456\std{0.049}$}
      & {\tiny $0.1319\std{0.025}$}
      & {\tiny $0.1692\std{0.051}$}
      & {\tiny $0.4253\std{0.072}$}
      & {\tiny $0.1514\std{0.134}$}
      & {\tiny $\mathbf{0.0385}\std{0.008}$}
      & {\tiny $0.0463\std{0.010}$} \\
\cline{1-12}
\multirow{8}{*}{Flickr}
  & \multirow{4}{*}{\textbf{One-way}}
    & {\tiny ADE}
      & {\tiny $0.0287\std{0.023}$}
      & {\tiny $0.0133\std{0.005}$}
      & {\tiny $0.1078\std{0.040}$}
      & {\tiny $0.0515\std{0.010}$}
      & {\tiny $\underline{0.0028}\std{0.002}$}
      & {\tiny $0.0451\std{0.064}$}
      & {\tiny $0.0152\std{0.019}$}
      & {\tiny $\mathbf{0.0010}\std{0.005}$}
      & {\tiny $0.0011\std{0.005}$} \\
  & & {\tiny ASE}
      & {\tiny $0.0989\std{0.061}$}
      & {\tiny $0.0109\std{0.006}$}
      & {\tiny $0.0956\std{0.065}$}
      & {\tiny $0.0735\std{0.031}$}
      & {\tiny $\underline{0.0042}\std{0.002}$}
      & {\tiny $0.0509\std{0.042}$}
      & {\tiny $0.0183\std{0.040}$}
      & {\tiny $\mathbf{0.0036}\std{0.016}$}
      & {\tiny $0.0024\std{0.008}$} \\
  & & {\tiny IDE}
      & {\tiny $0.2346\std{0.045}$}
      & {\tiny $\underline{0.1391}\std{0.028}$}
      & {\tiny $0.2940\std{0.028}$}
      & {\tiny $0.2595\std{0.051}$}
      & {\tiny $0.2630\std{0.041}$}
      & {\tiny $0.2654\std{0.043}$}
      & {\tiny $0.2454\std{0.048}$}
      & {\tiny $\mathbf{0.0201}\std{0.008}$}
      & {\tiny $0.0135\std{0.004}$} \\
  & & {\tiny ISE}
      & {\tiny $0.2888\std{0.031}$}
      & {\tiny $\underline{0.1665}\std{0.012}$}
      & {\tiny $0.2801\std{0.034}$}
      & {\tiny $0.2219\std{0.015}$}
      & {\tiny $0.2693\std{0.012}$}
      & {\tiny $0.2562\std{0.021}$}
      & {\tiny $0.2544\std{0.016}$}
      & {\tiny $\mathbf{0.0387}\std{0.008}$}
      & {\tiny $0.0257\std{0.003}$} \\
\cline{2-12}
  & \multirow{4}{*}{\textbf{Homo}}
    & {\tiny ADE}
      & {\tiny $0.0801\std{0.012}$}
      & {\tiny $0.7799\std{0.075}$}
      & {\tiny $0.1662\std{0.018}$}
      & {\tiny $0.1299\std{0.019}$}
      & {\tiny $0.6765\std{0.123}$}
      & {\tiny $0.1426\std{0.089}$}
      & {\tiny $\underline{0.0092}\std{0.022}$}
      & {\tiny $\mathbf{0.0001}\std{0.003}$}
      & {\tiny $0.0011\std{0.016}$} \\
  & & {\tiny ASE}
      & {\tiny $0.0586\std{0.018}$}
      & {\tiny $0.0775\std{0.025}$}
      & {\tiny $0.0695\std{0.037}$}
      & {\tiny $0.1699\std{0.039}$}
      & {\tiny $\underline{0.0212}\std{0.031}$}
      & {\tiny $0.2481\std{0.139}$}
      & {\tiny $0.1674\std{0.135}$}
      & {\tiny $\mathbf{0.0053}\std{0.010}$}
      & {\tiny $0.0049\std{0.017}$} \\
  & & {\tiny IDE}
      & {\tiny $0.1565\std{0.025}$}
      & {\tiny $0.8327\std{0.044}$}
      & {\tiny $0.1882\std{0.016}$}
      & {\tiny $0.1656\std{0.017}$}
      & {\tiny $0.6988\std{0.118}$}
      & {\tiny $0.2634\std{0.120}$}
      & {\tiny $\underline{0.0145}\std{0.018}$}
      & {\tiny $\mathbf{0.0087}\std{0.002}$}
      & {\tiny $0.0537\std{0.022}$} \\
  & & {\tiny ISE}
      & {\tiny $0.1780\std{0.025}$}
      & {\tiny $0.2462\std{0.042}$}
      & {\tiny $\underline{0.1736}\std{0.037}$}
      & {\tiny $0.2348\std{0.031}$}
      & {\tiny $0.2421\std{0.051}$}
      & {\tiny $0.3011\std{0.111}$}
      & {\tiny $0.2296\std{0.099}$}
      & {\tiny $\mathbf{0.0316}\std{0.014}$}
      & {\tiny $0.0490\std{0.013}$} \\
\hline
\end{tabular}
}
\end{table}
\subsection{Robustness to treatment-interaction misspecification}
\label{app:interaction_mispeci}

The main experiments use the first-order additive interference structure in
Assumption~3. To evaluate the robustness of this modeling choice, we further
consider an outcome-generating process with treatment interactions between the
ego treatment and neighbor treatments. Specifically, for each unit $i$, the
conditional outcome contains both first-order neighbor effects and second-order
interaction effects,
\[
\mathbb{E}\!\left[Y_i(t_{\bar N_1(i)}) \mid X_{\bar N_1(i)}, A\right]
=
g_i^{(0)}(X_{\bar N_1(i)},A)
+
t_i \tau_{ii}^{(1)}
+
\sum_{j\in N_1(i),\,j\neq i} t_j \tau_{ij}^{(1)}
+
\sum_{j\in N_1(i),\,j\neq i} t_i t_j \tau_{ij}^{(2)} .
\]

We compare two variants of our method. The
first variant, \textsc{Proposed-Main}, fits the original first-order model and is
therefore misspecified under this data-generating process. The second variant,
\textsc{Proposed-Int}, augments the second-stage orthogonal regression with
interaction residuals and fits both $\tau_{ij}^{(1)}$ and $\tau_{ij}^{(2)}$:
\[
\widehat \Gamma
=
\arg\min_{\{\tau_{ij}^{(1)},\tau_{ij}^{(2)}\}}
\sum_i
\left(
Y_i-\hat m_i
-
\sum_{j\in N_1(i)}
(T_j-\hat e_j)\tau_{ij}^{(1)}
-
\sum_{j\in N_1(i)}
(T_iT_j-\widehat e_{ij})\tau_{ij}^{(2)}
\right)^2 ,
\]
where $\hat m_i$, $\hat e_j$, and $\widehat e_{ij}$ are cross-fitted nuisance
estimates. This experiment is not intended to claim robustness to arbitrary
non-additive interference; rather, it tests whether the proposed framework can
be extended to low-order treatment interactions and whether the first-order
model remains useful under moderate misspecification.

\begin{table}[t]
\centering
\caption{
Robustness to treatment-interaction effects on Flickr dataset.}
\label{tab:interaction-robustness}
\small
\setlength{\tabcolsep}{4pt}
\begin{tabular}{lcccc}
\toprule
Method & ASE & ISE & ADE & IDE \\
\midrule
DRW~\citep{zhao2024dwr}
& $0.298 \pm 0.003$
& $0.346 \pm 0.003$
& $0.031 \pm 0.002$
& $0.032 \pm 0.002$ \\

IDENet~\citep{fan2026idenet}
& $0.383 \pm 0.003$
& $0.430 \pm 0.003$
& $0.051 \pm 0.004$
& $0.043 \pm 0.002$ \\

HINITE~\citep{lin2023hinite}
& $0.354 \pm 0.006$
& $0.405 \pm 0.006$
& $0.038 \pm 0.005$
& $0.039 \pm 0.005$ \\

NDR~\citep{liu2023ndr}
& $0.358 \pm 0.006$
& $0.409 \pm 0.006$
& $0.015 \pm 0.002$
& $0.022 \pm 0.002$ \\

SPNet~\citep{huang2024spnet}
& $0.365 \pm 0.003$
& $0.414 \pm 0.003$
& $0.042 \pm 0.005$
& $0.043 \pm 0.005$ \\
\midrule
\textsc{Proposed-Main}
& $0.098 \pm 0.098$
& $0.148 \pm 0.084$
& $0.028 \pm 0.031$
& $0.044 \pm 0.014$ \\

\textsc{Proposed-Int}
& $\mathbf{0.010 \pm 0.002}$
& $\mathbf{0.058 \pm 0.003}$
& $\mathbf{0.009 \pm 0.002}$
& $\mathbf{0.012 \pm 0.003}$ \\
\bottomrule
\end{tabular}
\end{table}

Table~\ref{tab:interaction-robustness} shows two patterns. First, when the
interaction terms are explicitly included, \textsc{Proposed-Int} gives the
lowest error across all four estimands, confirming that the orthogonal
second-stage formulation can be extended beyond the first-order additive model.
Second, even when the interaction terms are omitted, \textsc{Proposed-Main}
remains competitive for the spillover estimands and substantially improves over
the baselines on ASE and ISE. This suggests that the first-order estimator can
still recover useful marginal spillover information when the omitted interaction
component is moderate, although the correctly specified interaction model is
clearly preferable when such interactions are expected.

\subsection{\texorpdfstring{Generic Bernoulli-$z$ spillover estimand}{Generic Bernoulli-z spillover estimand}}
\label{app:bernoulli-z}

The main experiments evaluate spillover effects under the deterministic intervention
where all neighbors are treated. To assess whether the proposed method extends to
more general stochastic interventions, we consider a Bernoulli-$z$ spillover
estimand. For each unit $i$, let $b_{-i} = \{b_{-i,j}\}_{j \in N_1(i)\setminus i}$
denote independent draws with $b_{-i,j} \sim \mathrm{Bernoulli}(z)$. We define the
node-level spillover effect as
\[
\mathrm{ISE}_i(z)
=
\mathbb{E}\!\left[
Y_i(t_i = 0,\, t_{N_1(i)\setminus i} = b_{-i})
-
Y_i(t_{N_1(i)} = 0)
\;\middle|\;
X_{N_1(i)}, A
\right],
\]
and report the population-level average $\mathrm{ASE}(z) = \frac{1}{n}\sum_i \mathrm{ISE}_i(z)$.
This estimand interpolates between no spillover ($z=0$) and the fully treated
neighborhood ($z=1$), providing a more flexible characterization of peer effects.

We evaluate the proposed estimator (\textsc{Proposed-Est}) and its oracle variant
(\textsc{Proposed-Oracle}, using ground-truth nuisance functions) against GDML
under the interaction data-generating process described in
Appendix~\ref{app:interaction_mispeci}. All results are averaged over $20$ independent runs.

\begin{table}[t]
\centering
\caption{
Performance under Bernoulli-$z$ spillover interventions on Flickr dataset.
}
\label{tab:bernoulli-z}
\small
\setlength{\tabcolsep}{4pt}
\begin{tabular}{lcccc}
\toprule
 & $z=0.25$ & $z=0.50$ & $z=0.75$ & $z=1.00$ \\
\midrule
\multicolumn{5}{l}{\textbf{ASE}} \\
GDML
& $0.088 \pm 0.003$
& $0.083 \pm 0.004$
& $0.082 \pm 0.003$
& $0.081 \pm 0.003$ \\
\textsc{Proposed-Est}
& $0.006 \pm 0.001$
& $0.007 \pm 0.001$
& $0.006 \pm 0.001$
& $0.008 \pm 0.001$ \\
\textsc{Proposed-Oracle}
& $\mathbf{0.004 \pm 0.001}$
& $\mathbf{0.004 \pm 0.001}$
& $\mathbf{0.003 \pm 0.001}$
& $\mathbf{0.004 \pm 0.001}$ \\
\midrule
\multicolumn{5}{l}{\textbf{ISE}} \\
GDML
& $0.296 \pm 0.005$
& $0.280 \pm 0.005$
& $0.267 \pm 0.005$
& $0.259 \pm 0.004$ \\
\textsc{Proposed-Est}
& $0.061 \pm 0.002$
& $0.068 \pm 0.002$
& $0.069 \pm 0.002$
& $0.077 \pm 0.002$ \\
\textsc{Proposed-Oracle}
& $\mathbf{0.058 \pm 0.002}$
& $\mathbf{0.060 \pm 0.002}$
& $\mathbf{0.060 \pm 0.002}$
& $\mathbf{0.058 \pm 0.002}$ \\
\bottomrule
\end{tabular}
\end{table}

Table~\ref{tab:bernoulli-z} shows that the proposed estimator substantially
outperforms GDML across all values of $z$ for both $\mathrm{ASE}(z)$ and
$\mathrm{ISE}(z)$. The performance remains stable as $z$ varies, indicating that
the method is not sensitive to the specific intervention intensity. The oracle
results further suggest that the remaining gap is primarily due to nuisance
estimation error rather than limitations of the second-stage interference model.
These results support the applicability of the proposed framework to a broader
class of stochastic spillover interventions beyond the deterministic settings
considered in the main text.

\subsection{\texorpdfstring{Ablation study on the number of partitions \(K\)}{Ablation study on the number of partitions K}}\label{sec:partition}

\begin{figure} \centering \begin{subfigure}[b]{0.45\textwidth} \includegraphics[width=0.7\textwidth]{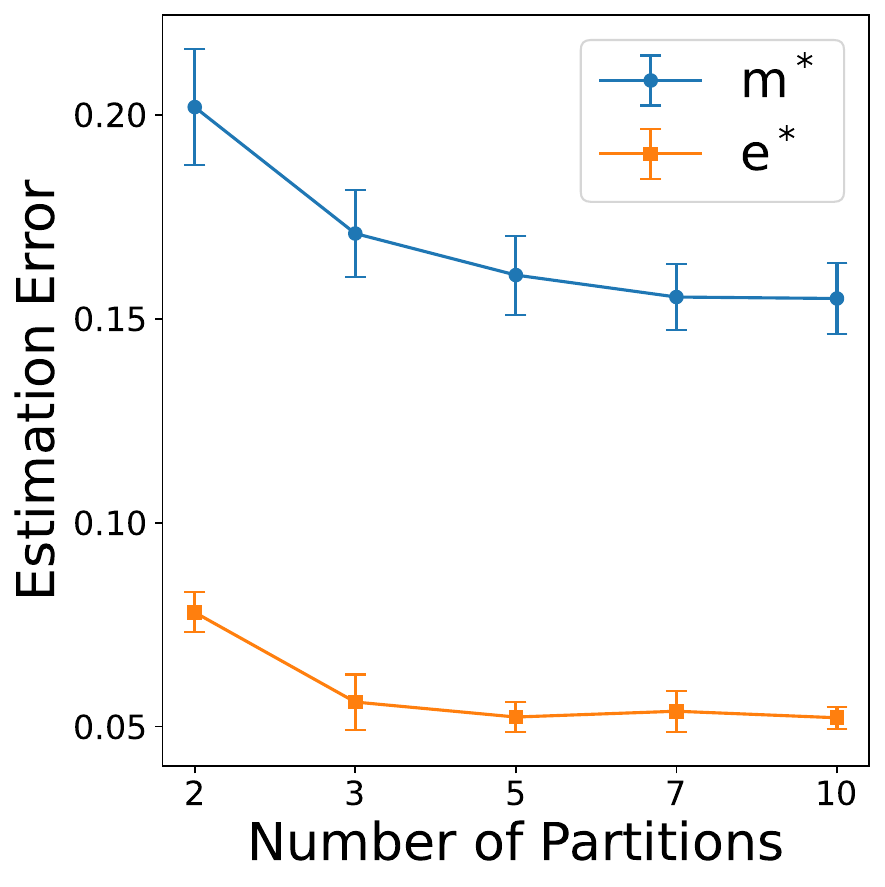} \caption{Nuisance estimate} \label{fig:plot1} \end{subfigure} \hfill \begin{subfigure}[b]{0.45\textwidth} \includegraphics[width=0.7\textwidth]{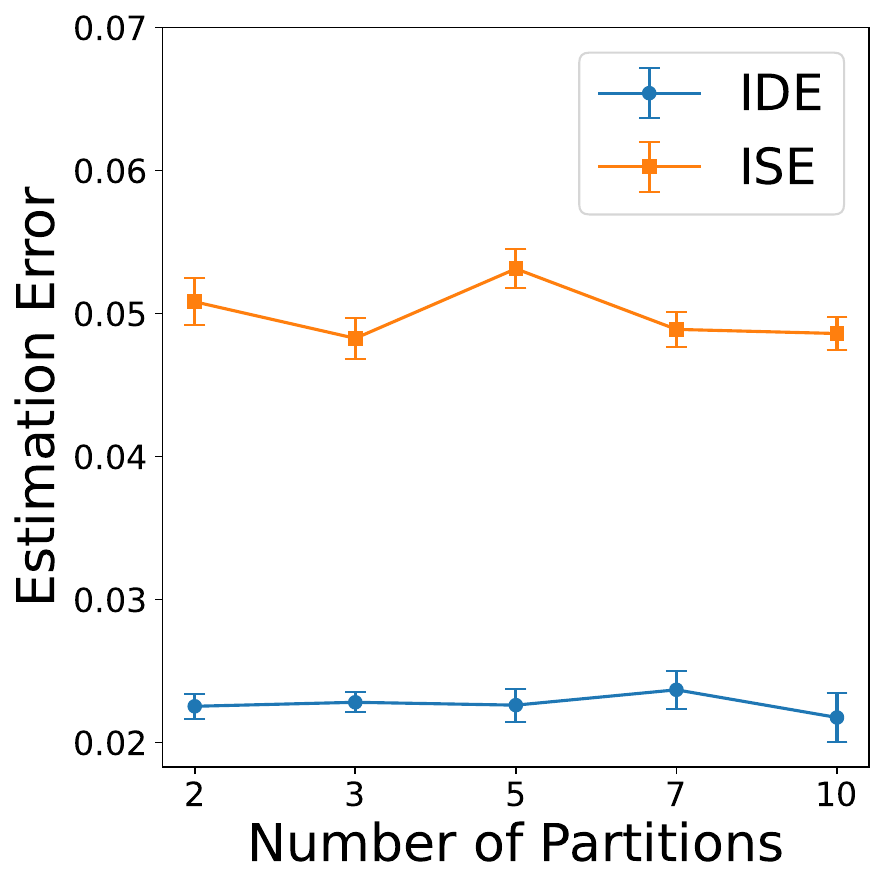} \caption{Nodel-level tasks} \label{fig:plot2} \end{subfigure} \caption{Effect of the number of partitions on performance using the Flickr dataset with the cosine kernel.} \label{Partitions} \end{figure}

We pick \(K = 5\) partitions for the main results and explore the effect of different partition counts using the METIS algorithm on the Flickr dataset. As shown in Figure~\ref{Partitions}, the accuracy of the nuisance estimates levels off around \(K = 5\), and when plugged into Stage 2, the downstream performance remains fairly stable across different values of \(K\). This pattern is in line with Theorem~2, which suggests that once the nuisance functions are estimated with reasonable accuracy, the final causal estimates are robust to how the data is split.

\subsection{Real-world DAPSm analysis}
\label{app:real_world_dapsm}

To complement the semi-synthetic experiments, we evaluate the proposed method on
a real-world air-pollution application based on the DAPSm power-plant study. In
this application, each node represents a power-generating facility, the treatment
indicates whether the facility installed selective catalytic reduction (SCR)
emission-control technology, and the outcome is ambient ozone concentration
around the facility, measured in parts per million (ppm). The dataset contains
$473$ facilities, with covariates describing power-plant characteristics, local
weather, and demographic characteristics of surrounding areas
\citep{papadogeorgou2019adjusting,papadogeorgou2019causal}.

This application naturally fits our network-interference setting. Although SCR is
installed at individual facilities, ozone pollution can be transported spatially,
so the ozone level around one facility may depend not only on its own treatment
status but also on nearby facilities' treatment status. We therefore construct a
network whose nodes are power plants and whose edges connect spatially nearby
plants. The direct effect corresponds to changing a plant's own SCR status, while
the indirect effect summarizes the spillover from SCR adoption among its
neighbors.

We consider interventions that increase the neighborhood SCR adoption share from
$0$ to $0.2$ and $0.4$. Under our sign convention, positive values indicate larger
estimated reductions in ambient ozone. Since ground-truth counterfactual effects
are unavailable in this observational setting, we report estimated direct and
indirect effect magnitudes rather than estimation errors.

\begin{table}[t]
\centering
\caption{
Real-world DAPSm analysis. We report estimated direct and indirect effect
magnitudes under increasing neighborhood SCR adoption share. Effects are measured
in parts per million (ppm).
}
\label{tab:dapsm-real}
\small
\setlength{\tabcolsep}{6pt}
\begin{tabular}{lccc}
\toprule
Effect & Adoption share $0$ & Adoption share $0.2$ & Adoption share $0.4$ \\
\midrule
Direct effect
& $0.032 \pm 0.009$
& $0.061 \pm 0.021$
& $0.044 \pm 0.011$ \\
Indirect effect
& $0.061 \pm 0.010$
& $0.073 \pm 0.010$
& $0.074 \pm 0.026$ \\
\bottomrule
\end{tabular}
\end{table}

Table~\ref{tab:dapsm-real} shows that both the estimated direct and indirect
effects are positive across the considered intervention levels. The direct-effect
estimates suggest that a facility's own SCR adoption is associated with lower
ambient ozone around that facility, while the indirect-effect estimates suggest
additional spillover reductions from neighboring facilities' SCR adoption. Because
this is a real-data analysis without known counterfactuals, these results should
be interpreted as an external case study rather than as an accuracy benchmark.

\section{Discussion of the additive local interference assumption}\label{dicuss_assumption3}
Assumption~3 is a structural restriction rather than a claim that all network spillovers are exactly additive. 
It rules out higher-order treatment interactions among neighbors, such as complementarity, substitution, 
threshold effects, or saturation effects that depend jointly on multiple treated neighbors. This restriction is 
therefore most appropriate when spillovers are primarily driven by first-order pairwise influences, or when 
higher-order interactions are weak relative to the main treatment and spillover effects, as shown is Appendix \ref{app:interaction_mispeci} and table \ref{tab:interaction-robustness}.

The role of Assumption~3 is to make a difficult high-dimensional interference problem statistically tractable. 
Without further structure, the conditional response surface
\[
  \mu_i(t_{\bar{\mathcal N}_1(i)})
  :=
  \mathbb E\!\left[
    Y_i(t_{\bar{\mathcal N}_1(i)})
    \mid
    X_{\bar{\mathcal N}_K(i)}, A
  \right]
\]
is an arbitrary function of the local treatment vector \(t_{\bar{\mathcal N}_1(i)}\). Since this vector has 
\(2^{\bar d_1(i)}\) possible configurations, identifying all local potential-outcome means would require 
overlap and sufficient observations for exponentially many treatment patterns. This is generally infeasible 
on a single observed network, especially when degrees are moderate or large. Similar dimensionality issues 
motivate the use of exposure mappings and neighborhood restrictions in the interference literature 
\citep{hudgens2008toward, aronow2017estimating, leung2022causal}. Assumption~3 reduces the local response 
surface to a sum of one-dimensional components, decreasing the effective complexity from exponential in 
\(\bar d_1(i)\) to linear in the number of neighbors. This reduction is what makes edge-level spillover 
estimation and uncertainty quantification feasible.

A useful way to interpret Assumption~3 is as a first-order approximation to a more general local interference 
function. For a fixed unit \(i\), write
\[
  h_i(t_{\bar{\mathcal N}_1(i)})
  =
  \mathbb E\!\left[
    Y_i(t_{\bar{\mathcal N}_1(i)})
    \mid
    X_{\bar{\mathcal N}_K(i)}, A
  \right].
\]
For binary treatments, \(h_i\) admits the multilinear expansion
\[
  h_i(t_{\bar{\mathcal N}_1(i)})
  =
  h_i(0)
  +
  \sum_{j\in \bar{\mathcal N}_1(i)}
  \Delta_j h_i(0) t_j
  +
  \sum_{j<k}
  \Delta_{jk}h_i(0)t_jt_k
  +
  \cdots,
\]
where \(\Delta_j h_i(0)\) is the first-order discrete difference and 
\(\Delta_{jk}h_i(0)\) is a second-order interaction contrast. Assumption~3 retains the first-order terms and 
absorbs heterogeneous dependence on covariates and network structure into the functions 
\(g_{ij}(t_j,X_{\bar{\mathcal N}_K(i)},A)\), while excluding the higher-order interaction terms. 
Equivalently, if one views \(h_i\) through a smooth extension to continuous treatment intensities, a Taylor 
expansion around a baseline treatment vector gives
\[
  h_i(t)
  =
  h_i(0)
  +
  \sum_{j\in \bar{\mathcal N}_1(i)}
  \frac{\partial h_i(0)}{\partial t_j} t_j
  +
  \frac12 t^\top H_i(\xi)t,
\]
where the final term captures second- and higher-order interactions. Thus, the additive model corresponds to 
the first-order part of the response surface. When the Hessian term is small, when treated neighbors are sparse, 
or when the main pairwise spillover effects dominate higher-order interactions, the additive approximation is 
expected to capture the leading causal variation. This is consistent with the effect-hierarchy principle in 
factorial and response-surface modeling, where lower-order effects are often treated as the primary effects and 
higher-order interactions are added only when supported by sufficient data \citep{box1986bayesian, 
yuan2007efficient}.

This assumption also reflects a tradeoff between expressiveness and interpretability. More flexible black-box 
interference models can encode arbitrary interactions among neighbors, but their learned representations are 
often difficult to interpret as pairwise causal spillovers and are hard to accompany with valid edge-level 
uncertainty quantification. In contrast, Assumption~3 yields identifiable signed pairwise effects 
\(\tau_{ij}\), node-level summaries such as \(\mathrm{ISE}_i=\sum_{j\in\mathcal N_1(i)}\tau_{ij}\), and 
population-level summaries such as \(\mathrm{ASE}\). These quantities are directly useful for downstream 
questions such as which neighbors are most influential and whether their spillovers are positive or negative.

When Assumption~3 is violated by strong treatment interactions, the estimated \(\tau_{ij}\) should be interpreted 
as first-order or marginal additive spillover effects rather than as a complete structural description of the 
network response surface. In this case, the omitted higher-order terms represent approximation error. Our 
framework can be extended to include such interactions by augmenting the second-stage model with terms such as
\[
  \sum_{j<k}
  (T_jT_k - e_{ijk})\,\tau_{ijk},
  \qquad
  e_{ijk}:=
  \mathbb E[T_jT_k\mid X_{\bar{\mathcal N}_K(i)},A],
\]
or ego-neighbor interaction terms such as
\[
  \sum_{j\in\mathcal N_1(i)}
  (T_iT_j-e_{ij})\,\tau^{(2)}_{ij}.
\]
The same orthogonalization principle applies after augmenting the nuisance functions to include the 
corresponding joint propensities. However, this extension increases the number of target parameters from 
\(O(d_1(i))\) to \(O(d_1(i)^2)\) or higher, requires stronger overlap over joint treatment configurations, and 
substantially increases the sample size needed for stable estimation and inference. For this reason, we use the 
additive model as a principled first-order specification and evaluate robustness to interaction misspecification 
in additional experiments.

\section{Technical Preliminaries and Notation}\label{app:technical-preliminaries}

\subsection{Basic notation}\label{app:basic-notation}
We define the \(L_p\) norm for \(p \geq 1\) as
\[  \|f\|_p = \left(\mathbb{E}_D[|f(X)|^p]\right)^{1/p},\]
where \(D\) is the distribution of \(X\). We write \(f(x)=\mathcal{O}(g(x))\) if there exists \(x_0\) such that
\[  |f(x)/g(x)| \le M  \qquad  \text{for all } x>x_0 .\]

\subsection{Fixed-network setup}\label{app:fixed-network-setup}
We consider a fixed network \(A_n\) on vertices \([n]=\{1,\ldots,n\}\). All probability statements below are conditional on the realized network \(A_n\). Let
\[  \N_r(i)=\{j\in[n]:\ell_A(i,j)\le r,\ j\ne i\},  \qquad  \ol{\N}_r(i)=\N_r(i)\cup\{i\},\]
where \(\ell_A(i,j)\) is the shortest-path distance in \(A_n\). Let
\[  d_i=\abs{\N_1(i)},\qquad  k_i=d_i+1,\qquad  k_{\max,n}=\max_{1\le i\le n} k_i .\]
The observed node-level data are
\[  D_i=(X_{\ol{\N}_1(i)},T_{\ol{\N}_1(i)},Y_i),\]
where \(X_i\in[0,1]^d\), \(T_i\in\{0,1\}\), and \(Y_i\in\R\).

\subsection{Residualized loss, risks, and estimators}\label{app:loss-risk-estimator}
We write
\[  R_{ij}(g)=T_j-e_{ij},\qquad j\in\ol{\N}_1(i),\]
where \(g=(m,e)\) denotes the nuisance functions,
\[  m_i=m_i(X_{\ol{\N}_{K}(i)}),  \qquad  e_{ij}=e_{ij}(X_{\ol{\N}_{K}(i)}).\]
The true nuisance is denoted
\[  g^\star=(m^\star,e^\star).\]
For each row \(i\), the interference coefficient vector is
\[  W_{i\cdot}=(W_{ij}:j\in\ol{\N}_1(i)).\]
The row-wise residualized square loss is
\begin{equation}\label{eq:loss}  \ell_i(W,g)  =  \frac12  \left[    Y_i-m_i-\sum_{j\in\ol{\N}_1(i)}    \{T_j-e_{ij}\}W_{ij}  \right]^2 .\end{equation}
The empirical and population risks are
\[  \wh L_n(W,g)=\frac1n\sum_{i=1}^n \ell_i(W,g),  \qquad  L_n(W,g)=\frac1n\sum_{i=1}^n \E[\ell_i(W,g)].\]
Here \(L_n\) is allowed to depend on \(n\), because the observations are network-dependent and the row dimensions \(k_i\) may vary with \(i\). Let
\[  W^\star\in\argmin_{W\in\cW_n} L_n(W,g^\star)\]
be a population minimizer. Given fixed nuisance estimates
\[  \wh g=(\wh m,\wh e),\]
the estimator of the interference coefficients is the plug-in empirical risk minimizer
\begin{equation}\label{eq:erm}  \wh W\in\argmin_{W\in\cW_n}\wh L_n(W,\wh g).\end{equation}

\subsection{Prediction and nuisance-error norms}\label{app:prediction-nuisance-norms}
For row \(i\), define the prediction norm
\begin{equation}\label{eq:theta-norm}  \norm{\Delta W_{i\cdot}}_{\Theta}^2  =  \E\left[  \left\{    \sum_{j\in\ol{\N}_1(i)}    (T_j-e_{ij}^\star)\Delta W_{ij}  \right\}^2  \right],\end{equation}
where \(\Delta W_{i\cdot}=W_{i\cdot}-W^\star_{i\cdot}\). Define the averaged prediction norm
\[  \norm{\Delta W}_{\Theta,n}^2  =  \frac1n\sum_{i=1}^n \norm{\Delta W_{i\cdot}}_{\Theta}^2.\]
For the nuisance errors \(\Delta m_{i}=m_{i}-m^\star_{i}\) and \(\Delta e_{ij}=e_{ij}-e^\star_{ij}\), define
\begin{equation}\label{eq:g-norm}  \norm{\Delta g_i}_{G}  =  \left[  \E\left\{    (\Delta m_i)^2+    \sum_{j\in\ol{\N}_1(i)}(\Delta e_{ij})^2  \right\}^2  \right]^{1/4},\end{equation}
and
\[  \cR_{g,n}^4  =  \frac1n\sum_{i=1}^n  k_i\norm{\wh g_i-g_i^\star}_{G}^4.\]

\subsection{Attention-based ReLU coefficient class}\label{app:attention-relu-class}
The coefficient class \(\cW_n\) is induced by two ReLU network classes and one scalar attention parameter. Let
\[  \cA=[-\theta,\theta],  \qquad \theta>0.\]
For \(f_1:[0,1]^d\to\R\), \(f_2:[0,1]^{2d}\to\R\), and \(a\in\cA\), define
\[  W_{ii}=f_1(X_i),\]
and for \(j\in\N_1(i)\),
\begin{equation}\label{eq:attention-W}  W_{ij}  =  \alpha_{ij}(f_2,a)f_2(X_i,X_j),  \qquad  \alpha_{ij}(f_2,a)  =  \frac{    \exp\{|a f_2(X_i,X_j)|\}  }{    \sum_{\ell\in\N_1(i)}    \exp\{|a f_2(X_i,X_\ell)|\}  }.\end{equation}
Thus
\[  W=W(f_1,f_2,a)\in \mathcal{W}_n,  \qquad  \mathcal{W}_n  :=  \{W(f_1,f_2,a):(f_1,f_2,a)\in\cF_1\times\cF_2\times\cA\}.\]
We use bounded-envelope fully connected ReLU classes. For input dimension \(p\), width \(M\), depth \(L\), weight bound \(\Lambda\ge 1\), and deterministic output envelope \(B>0\), define
\[  \cR_B(p,M,L,\Lambda;B)  =  \left\{  f=R(\Phi):
	\begin{array}{l}  \Phi \text{ is a fully connected ReLU network},\\  \text{input dimension }p,\text{ scalar output},\\  \text{width}\le M,\ \text{depth}\le L,\\  \text{all affine weights and biases bounded by }\Lambda,\\  \norm{f}_{L^\infty([0,1]^p)}\le B  \end{array}  \right\}.\]
We take
\begin{equation}\label{eq:relu-classes}  \cF_1=\cR_B(d,M_1,L_1,\Lambda_1;B),  \qquad  \cF_2=\cR_B(2d,M_2,L_2,\Lambda_2;B).\end{equation}
Consequently,
\[  \norm{f_1}_\infty\le B,\qquad  \norm{f_2}_\infty\le B.\]

\subsection{Auxiliary assumptions for the proof of Theorem~2}\label{app:auxiliary-assumptions}
Recall \textbf{Assumption 1}: for any two subsets of nodes on network \(\mathcal{S}_1\) and \(\mathcal{S}_2\) such that
\[  \ell(i,j)\geq K+2,  \qquad  \forall i\in\mathcal{S}_1,\ \forall j\in\mathcal{S}_2,\]
we have
\begin{align*}    \cup_{i\in \mathcal{S}_1} (Y_{\bar{\mathcal{N}}_1(i)}, T_{\bar{\mathcal{N}}_1(i)})    \perp    \cup_{i\in \mathcal{S}_2} (Y_{\bar{\mathcal{N}}_1(i)}, T_{\bar{\mathcal{N}}_1(i)}, X_{\bar{\mathcal{N}}_1(i)})    \mid    \cup_{i\in \mathcal{S}_1} \bm{X}_{\bar{\mathcal{N}}_{K}(i)},\end{align*}
where \(\cup\) above stands for the joint distribution among the components.
\noindent\textbf{Assumption 2}: \(\{X_i\}_{i\in \mathbf{V}}\) are independent.
\noindent\textbf{Assumption 3}: the residual \(\eps_i\) in the outcome model satisfies
\[    (Y_i,T_{\bar{\mathcal{N}}_1(i)})    \perp    \epsilon_i    \mid    \bm{X}_{\bar{\mathcal{N}}_{K}(i)} .\]

\noindent\textbf{Assumption 4}: for any \(j,k \in \bar{{\mathcal{N}}}_1(i)\), there exists constant \(0\leq\rho < 1\) such that
\[ |\text{corr}(T_j, T_k \mid \bm{X}_{\bar{\mathcal{N}}_i(K)})| \leq \rho.\]

\noindent\textbf{Assumption 5}: for constants \(M_Y,M_m<\infty\),
\[  |Y_i|\le M_Y,\qquad  |m_i^\star|\le M_m,\qquad  |\wh m_i|\le M_m,\]
and
\[  e_{ij}^\star,\wh e_{ij}\in[c,1-c],  \qquad  0<c<1/2 .\]

\section{Proof of Identification Results (Theorem 1)}\label{app:identification-proof}

\begin{proof}
Fix a unit \(i\). For notational simplicity, write
\[
  \bar N_i := \bar{\mathcal N}_1(i),
  \qquad
  N_i := \mathcal N_1(i),
  \qquad
  X_i^{(K)} := X_{\bar{\mathcal N}_K(i)} .
\]
For any local treatment vector \(s\in\{0,1\}^{|\bar N_i|}\), define the observed conditional mean
\[
  \mu_i(s)
  :=
  \mathbb{E}\!\left[
    Y_i
    \mid
    T_{\bar N_i}=s,\,
    X_i^{(K)},\, A
  \right].
\]
By local interference, the conditional potential outcome of unit \(i\) depends on the global
assignment \(t\) only through the closed one-hop treatment vector \(t_{\bar N_i}\):
\[
  \mathbb{E}\!\left[
    Y_i(t)
    \mid X,A
  \right]
  =
  \mathbb{E}\!\left[
    Y_i(t_{\bar N_i})
    \mid X_i^{(K)},A
  \right].
\]
By neighborhood unconfoundedness and overlap, for every
\(s\in\{0,1\}^{|\bar N_i|}\),
\[
  \mathbb{E}\!\left[
    Y_i(s)
    \mid X_i^{(K)},A
  \right]
  =
  \mathbb{E}\!\left[
    Y_i(s)
    \mid T_{\bar N_i}=s,\,
    X_i^{(K)},A
  \right]
  =
  \mathbb{E}\!\left[
    Y_i
    \mid T_{\bar N_i}=s,\,
    X_i^{(K)},A
  \right]
  =
  \mu_i(s),
\]
where the second equality uses consistency. Thus the local conditional potential-outcome
mean is identifiable from the observed law of
\[
  \left(Y_i,T_{\bar N_i},X_i^{(K)},A\right).
\]

By additive local interference, there exist functions
\[
  \{g_{ij}(t_j,X_i^{(K)},A):j\in\bar N_i\}
\]
such that
\[
  \mu_i(s)
  =
  \mathbb{E}\!\left[
    Y_i(s)
    \mid X_i^{(K)},A
  \right]
  =
  \sum_{j\in\bar N_i}
  g_{ij}(s_j,X_i^{(K)},A).
\]
For \(j=i\), define
\[
  \tau_i
  :=
  g_{ii}(1,X_i^{(K)},A)
  -
  g_{ii}(0,X_i^{(K)},A),
\]
and for \(j\in N_i\), define
\[
  \tau_{ij}
  :=
  g_{ij}(1,X_i^{(K)},A)
  -
  g_{ij}(0,X_i^{(K)},A).
\]
Equivalently, these contrasts are identifiable from observed conditional means. For example,
letting \(s^{(j,1)}\) and \(s^{(j,0)}\) be two local treatment vectors that differ only in the
\(j\)-th coordinate, with all other local treatments held fixed, we have
\[
  \mu_i(s^{(j,1)})-\mu_i(s^{(j,0)})
  =
  g_{ij}(1,X_i^{(K)},A)
  -
  g_{ij}(0,X_i^{(K)},A).
\]
Thus \(\tau_i\) and \(\tau_{ij}\) are identifiable, although the levels of the functions
\(g_{ij}\) themselves need not be uniquely identified without an additional normalization.

Now consider the direct effect. Let \(0_{\bar N_i}\) denote the all-zero treatment vector on
\(\bar N_i\), and let \(e_i\) denote the local vector with \(t_i=1\) and \(t_j=0\) for all
\(j\in N_i\). Then
\[
\begin{aligned}
  IDE_i
  &=
  \mathbb{E}\!\left[
    Y_i(t_i=1,t_{-i}=0)
    -
    Y_i(t_i=0,t_{-i}=0)
    \mid X,A
  \right]  \\
  &=
  \mu_i(e_i)-\mu_i(0_{\bar N_i}) \\
  &=
  g_{ii}(1,X_i^{(K)},A)-g_{ii}(0,X_i^{(K)},A) \\
  &=
  \tau_i .
\end{aligned}
\]
Similarly, for the spillover effect, let \(s^{\mathrm{sp}}\) denote the local vector with
\(t_i=0\) and \(t_j=1\) for all \(j\in N_i\). Then
\[
\begin{aligned}
  ISE_i
  &=
  \mathbb{E}\!\left[
    Y_i(t_i=0,t_{-i}=1)
    -
    Y_i(t_i=0,t_{-i}=0)
    \mid X,A
  \right] \\
  &=
  \mu_i(s^{\mathrm{sp}})-\mu_i(0_{\bar N_i}) \\
  &=
  \sum_{j\in N_i}
  \left\{
  g_{ij}(1,X_i^{(K)},A)
  -
  g_{ij}(0,X_i^{(K)},A)
  \right\} \\
  &=
  \sum_{j\in N_i}\tau_{ij}.
\end{aligned}
\]
For the total effect, let \(1_{\bar N_i}\) denote the all-one treatment vector on \(\bar N_i\).
Then
\[
\begin{aligned}
  ITE_i
  &=
  \mathbb{E}\!\left[
    Y_i(t_i=1,t_{-i}=1)
    -
    Y_i(t_i=0,t_{-i}=0)
    \mid X,A
  \right] \\
  &=
  \mu_i(1_{\bar N_i})-\mu_i(0_{\bar N_i}) \\
  &=
  \tau_i+\sum_{j\in N_i}\tau_{ij}.
\end{aligned}
\]

Finally, for any two global treatment assignments \(t\) and \(t'\), local interference and
additivity give
\[
\begin{aligned}
  \mathbb{E}\!\left[
    Y_i(t)-Y_i(t')
    \mid X,A
  \right]
  &=
  \mu_i(t_{\bar N_i})-\mu_i(t'_{\bar N_i}) \\
  &=
  \sum_{j\in\bar N_i}
  \left\{
  g_{ij}(t_j,X_i^{(K)},A)
  -
  g_{ij}(t'_j,X_i^{(K)},A)
  \right\} \\
  &=
  \tau_i(t_i-t'_i)
  +
  \sum_{j\in N_i}
  \tau_{ij}(t_j-t'_j),
\end{aligned}
\]
where the last equality uses binary treatments. This proves the stated identification results.
\end{proof}


\section{Proof of the Orthogonal Estimation Bound (Theorem 2)}\label{app:orthogonal-estimation-proof}
This appendix proves the main estimation result for the proposed orthogonal estimator. The proof has three components. First, Appendix~\ref{app:cnd-empirical-process} establishes empirical-process control under conditional neighborhood dependence. Second, Appendix~\ref{app:entropy-attention-relu} bounds the entropy of the attention-based ReLU interference class. Finally, we combine these ingredients with the orthogonal-learning perturbation bound to obtain the rate in Theorem~2.

\subsection{Cross-fitting and independence}\label{app:cross-fitting-independence}
Given Assumptions 1 and 2, when \(\ell(i,j)\geq 2K+1\), we have
\begin{equation}\label{cond_1}    (Y_{\bar{\mathcal{N}}_1(i)}, T_{\bar{\mathcal{N}}_1(i)}, X_{\bar{\mathcal{N}}_1(i)})    \perp    (Y_{\bar{\mathcal{N}}_1(j)}, T_{\bar{\mathcal{N}}_1(j)}, X_{\bar{\mathcal{N}}_1(j)}).\end{equation}
\begin{proof}    From Assumption 1, we have
	\[    (Y_{\bar{\mathcal{N}}_1(i)}, T_{\bar{\mathcal{N}}_1(i)})    \perp    (Y_{\bar{\mathcal{N}}_1(j)}, T_{\bar{\mathcal{N}}_1(j)}, X_{\bar{\mathcal{N}}_1(j)})    \mid    X_{\bar{\mathcal{N}}_K(i)}    \]    when \(\ell(i,j)\geq 2K+1\). Also,
	\[    (Y_{\bar{\mathcal{N}}_1(j)}, T_{\bar{\mathcal{N}}_1(j)})    \perp    X_{\bar{\mathcal{N}}_K(i)}    \mid    X_{\bar{\mathcal{N}}_K(j)},    \]    and
	\[    X_{\bar{\mathcal{N}}_K(j)}    \perp    X_{\bar{\mathcal{N}}_K(i)}.    \]    Therefore,
	\[    X_{\bar{\mathcal{N}}_K(i)}    \perp    (Y_{\bar{\mathcal{N}}_1(j)}, T_{\bar{\mathcal{N}}_1(j)}, X_{\bar{\mathcal{N}}_1(j)}).    \]    Hence,
	\[    (Y_{\bar{\mathcal{N}}_1(i)}, T_{\bar{\mathcal{N}}_1(i)}, X_{\bar{\mathcal{N}}_1(i)})    \perp    (Y_{\bar{\mathcal{N}}_1(j)}, T_{\bar{\mathcal{N}}_1(j)}, X_{\bar{\mathcal{N}}_1(j)}).    \]    We use the fact that if \(A \perp B \mid C\) and \(A \perp C\), then \(A \perp B\).\end{proof}
Based on \eqref{cond_1}, when we perform data splitting and estimate the nuisance \(\hat g\) and the target coefficient \( \hat W \) via cross-fitting, \(\hat g\) and \(\hat W\) become independent across separated folds. As a roadmap, we first verify the orthogonal-learning conditions in \citep{foster2023orthogonal} for each node \(i\) and each row \(W_{i\cdot}\) as the target parameter. We then follow the proof strategy of \citep{foster2023orthogonal} to derive convergence of \(\{\hat W_{ij}\}\) to \(\{W^\star_{ij}\}\).

\subsection{Verification of the orthogonal-learning conditions}\label{app:orthogonal-conditions}
Denote
\[  D_i = (X_{\bar{\mathcal{N}}_1(i)},T_{\bar{\mathcal{N}}_1(i)},Y_i)\]
from the joint distribution \(D\). Following the notation in Assumption 9 of \citep{foster2023orthogonal}, let
\[  \zeta=\Theta:=(W_{ij})_{j\in\bar{\mathcal{N}}_1(i)}\in\mathbb{R}^{k_i}\]
be the target functions to estimate, and let
\[  \gamma=g:=\{m_i,e_{ij}, j\in\bar{\mathcal{N}}_1(i)\}\]
be the nuisance function. Then we can write
\[  l_i=\frac12(Y_i-m_i-t)^2,\]
where
\[  t=\langle \bm{\Lambda}, \bm{W} \rangle,\qquad  \bm{\Lambda}=[T_j-e_j^\star]_{j\in\bar{\mathcal{N}}_1(i)},  \qquad  \bm{W}=[W_j]_{j\in\bar{\mathcal{N}}_1(i)}.\]
We check conditions (77)--(83) in Assumption 9 of \citep{foster2023orthogonal} as follows.

\paragraph{Condition (77).}
For \(j\in \bar{\mathcal{N}}_1(i)\),
\begin{align*}    &\frac{\partial l_i}{\partial W_j}    =    - (Y-m_i-\langle \bm{\Lambda}, \bm{W} \rangle)(T_j - e^{\star}_j),\\    &\frac{\partial^2 l_i}{\partial W_j \partial m_i}    =    T_j - e^{\star}_j,\\    &\frac{\partial^2 l_i}{\partial W_j \partial e^{\star}_j}    =    (Y-m_i-\langle \bm{\Lambda}, \bm{W} \rangle) + W_{ij}(T_j - e^{\star}_j),\\    &\frac{\partial^2 l_i}{\partial W_j \partial e^{\star}_k}    =    W_{ik}(T_j - e^{\star}_j),    \qquad    k\neq j.\end{align*}
Also, given
\[  \mathbb{E}_{D\mid X}(T_j-e_j^\star)=0\]
and
\[  \mathbb{E}_{D\mid X}(Y-m_i-\langle \bm{\Lambda}, \bm{W} \rangle)=0,\]
condition (77) holds.

\paragraph{Condition (78).}
Given the exogenous-error Assumption 3, we have
\begin{align*}    &\mathbb{E}_{D\mid X}\left[    (Y-m^{\star}_i-\langle \bm{\Lambda}, \bm{W} \rangle)(T_j - e^{\star}_j)    \right]  \\    &\qquad =    \mathbb{E}_{D\mid X}\left[    Y-m^{\star}_i-\langle \bm{\Lambda}, \bm{W} \rangle    \right]    \mathbb{E}_{D\mid X}\left[    T_j - e^{\star}_j    \right]    =    0.\end{align*}
Here
\[  \mathbb{E}_{D\mid X}  =  \mathbb{E}_{(Y_i,T_{\bar{\mathcal{N}}_1(i)})\mid  \bm{X}_{\bar{\mathcal{N}}_{K+1}(i)}}.\]

\paragraph{Condition (79).}
Notice that
\[  \frac{\partial^3 l_i}{\partial^2 g \partial W_{ij}}\]
is a \((d_i+2)\times(d_i+2)\) matrix with rows and columns indexed by
\[  [(e_{ij})_{j\in \bar{\mathcal{N}}_1(i)},m_i].\]
For each \(j \in \bar{\mathcal{N}}_1(i)\),
\begin{align*}\frac{\partial^3 l_i}{\partial W_{ij}\partial e_k \partial e_l } &= 0,\qquad l\neq j,\\\frac{\partial^3 l_i}{\partial W_{ij}\partial e_k \partial m_i } &= 0,\\\frac{\partial^3 l_i}{\partial W_{ij}\partial e_k \partial e_j } &= -W_{ik},\\\frac{\partial^3 l_i}{\partial W_{ij}\partial e_j \partial e_k } &= -W_{ik},\qquad k\neq j,\\\frac{\partial^3 l_i}{\partial W_{ij}\partial e_k \partial m_i } &= -1,\\\frac{\partial^3 l_i}{\partial W_{ij}\partial e_j \partial e_j } &= -2W_{ij}.\end{align*}
By the Gershgorin circle theorem,
\begin{align*}    \left\|    \mathbb{E}_{D}    \left[    \frac{\partial^3 l_i}{\partial^2 g \partial W_{ij}}    \right]    \right\|_{op}    \leq    2|W_{ij}| + 1 + \sum_{k\neq j,k\in \bar{\mathcal{N}}_1(i)}|W_{ik}|    \leq    1+3B.\end{align*}
Therefore, condition (79) holds with
\[  \mu_{si}=1+3B.\]

\paragraph{Condition (80).}
We have
\[  \phi(t)=t-\langle \bm{\Lambda}, \bm{W}\rangle,\]
and therefore
\[  T_{si}=\phi'(t)=\tau_{si}=1.\]

\paragraph{Condition (81).}
Notice that
\begin{align*}    \| W_{i\cdot} \|^2_{\Theta}    &=    \mathbb{E}_{D}\left[ \langle \bm{\Lambda}, \bm{W} \rangle^2\right] \\    &=    \mathbb{E}_{X}\mathbb{E}_{D|X}    \left[    \left( \sum_{j}(T_j-e^{\star}_j)W_{ij}\right)^2    \right] \\    &=    \mathbb{E}_{X}\left[    \sum_{j}\mathbb{E}_{D|X}(T_j-e^{\star}_j)^2 W^2_{ij}    +    \sum_{j\neq k}    \mathbb{E}_{D|X}(T_j-e^{\star}_j)(T_k-e^{\star}_k)W_{ij}W_{ik}    \right].\end{align*}
By the treatment-dependence Assumption 4,
\begin{align*}    &\sum_{j}\mathbb{E}_{D|X}(T_j-e^{\star}_j)^2 W^2_{ij}    +    \sum_{j\neq k}    \mathbb{E}_{D|X}(T_j-e^{\star}_j)(T_k-e^{\star}_k)W_{ij}W_{ik} \\    &\quad \geq    \sum_{j}\mathbb{E}_{D|X}(T_j-e^{\star}_j)^2 W^2_{ij}    -    \rho    \sum_{j\neq k}    \sqrt{\mathbb{E}_{D|X}(T_j-e^{\star}_j)^2}    \sqrt{\mathbb{E}_{D|X}(T_k-e^{\star}_k)^2}    W_{ij}W_{ik} \\    &\quad \geq    (1-\rho)    \sum_{j}    \mathbb{E}_{D|X}(T_j-e^{\star}_j)^2 W^2_{ij}.\end{align*}
With \(e^\star\in[c,1-c]\), we have
\[  c(1-c)\leq e^{\star}_j(1-e^{\star}_j)\leq \frac14.\]
Therefore,
\begin{align*}    \| W \|^2_{\Theta}    &\geq    (1-\rho)    \mathbb{E}_{X}    \left[    \sum_{j} e^{\star}_j(1-e^{\star}_j) W^2_{ij}    \right] \\    &\geq    (1-\rho)c(1-c)    \mathbb{E}_{X}    \left[    \sum_{j\in \bar{\mathcal{N}}_1(i)}W^2_{ij}    \right].\end{align*}
Thus,
\[    \| W \|_{\Theta}    \geq    \sqrt{(1-\rho)c(1-c)}    \| W_{i\cdot}\|_{L_2(D,l_2)}.\]
Condition (81) is valid with
\[  \lambda_{si}=(1-\rho)c(1-c).\]

\paragraph{Condition (82).}
Notice that
\[    \| \bm{\Lambda}-\bm{\Lambda}' \|^2_2    =    \sum_{j\in \bar{\mathcal{N}}_1(i)} (e^{\star}_{j}-e_{j'} )^2    \leq    \| g-g'\|^2_2.\]
Therefore, condition (82) is valid with
\[  L_{si}=1.\]

\paragraph{Condition (83).}
Condition (83) holds with
\[  R_{si}=B\sqrt{k_i}.\]
Then, following Lemma 5 of \citep{foster2023orthogonal}, we may take
\[  r=0,\qquad  \lambda=\frac14,\qquad  \kappa_i=  \frac{2B^2(d_i+1)}{(1-\rho)c(1-c)},\]
and
\[  \beta_1=1,\qquad  \beta_{2i}  =  \frac{(1+3B)\sqrt{d_i+1}}{\sqrt{(1-\rho)c(1-c)}}.\]

\subsection{Orthogonal perturbation bound}\label{app:orthogonal-perturbation}
Assume that with probability at least \(1-\delta\),
\[    L_n(\hat{W},g) - L_n({W}^{\star},g)    \leq    \operatorname{Rate}(D,W,\delta,g).\]
Notice that the population loss satisfies
\[  L_n  =  \mathbb{E}_D\left(\frac1n\sum_i l_i\right)  =  \frac1n\sum_i \mathbb{E}_D(l_i).\]
By linearity of the population loss and the orthogonal-learning perturbation argument, with probability at least \(1-\delta\),
\[    \frac{1}{n}\sum_i    \|\hat{W}_{i\cdot}- {W}^{\star}_{i\cdot}\|_{\Theta}^2    \leq    \frac{4}{\lambda}    \operatorname{Rate}_D(W,\delta,g)    +    \frac{2}{n}\sum_i    \left(    \frac{\beta^2_{2i}}{\lambda^2}    +    \frac{\kappa_i}{\lambda}    \right)    \| \hat{g}_i-g^{\star}_i\|^4_{\mathcal{G}}.\]
Therefore,
\begin{equation}\label{eq:ol-bound}  \norm{\wh W-W^\star}_{\Theta,n}^2  \le  \frac{4}{\lambda}  \operatorname{Rate}(D, W,\delta;\wh g)  +  2\left[  \left(\frac{\beta_2}{\lambda}\right)^2  +  \frac{\kappa}{\lambda}  \right]\cR_{g,n}^4 .\end{equation}
Here \(\operatorname{Rate}(D,W,\delta;\wh g)\) is the second-stage oracle excess-risk rate that would be obtained if the nuisance were fixed at \(\wh g\), and
\[  \beta_2=\frac{1+3B}{\sqrt{(1-\rho)c(1-c)}},  \qquad  \kappa=\frac{2B^2}{(1-\rho)c(1-c)}.\]
Let
\[  c_{\rho}=\sqrt{(1-\rho)c(1-c)}.\]
Then
\begin{align}\label{eq:l2-bound-rate}    \frac1n\sum_{i=1}^n    \norm{\wh W_{i\cdot}-W^\star_{i\cdot}}_2^2    &\leq    \frac{4}{\lambda c_{\rho}}    \operatorname{Rate}(D, W,\delta;\wh g) \notag\\    &\quad+    \frac{2}{c_{\rho}}    \left[    \left(\frac{\beta_2}{\lambda}\right)^2    +    \frac{\kappa}{\lambda}    \right]\cR_{g,n}^4.\end{align}

\subsection{Reduction to the empirical process}\label{app:reduction-empirical-process}
For plug-in ERM, define the empirical process
\[  Z_n  :=  \sqrt{n}\sup_{W\in\cW_n}  \left|  (L_n-\wh L_n)  \{\ell(W,\wh g)-\ell(W^\star,\wh g)\}  \right|.\]
Since \(\wh W\) minimizes the empirical risk,
\[ \wh L_n(\wh W,\wh g)-\wh L_n(W^\star,\wh g)  \le 0.\]
Therefore,
\[
	\begin{aligned}  L_n(\wh W,\wh g)-L_n(W^\star,\wh g)  &=  \left[  L_n(\wh W,\wh g)-L_n(W^\star,\wh g)  \right]  -  \left[  \wh L_n(\wh W,\wh g)-\wh L_n(W^\star,\wh g)  \right] \\  &\le \frac{Z_n}{\sqrt n}.\end{aligned}
\]
Thus, if \(\mathbb{E}(Z_n)\le \eta_n\), then by Markov's inequality, with probability at least \(1-\delta\),
\[  \operatorname{Rate}(D, W,\delta;\wh g)  \leq  \frac{\eta_n}{\delta\sqrt n}.\]
Appendix~\ref{app:cnd-empirical-process} bounds \(\mathbb{E}(Z_n)\) using the maximal inequality of \citep{lee2019stable}.

\subsection{Final proof of Theorem~2}\label{app:final-proof-theorem-2}
Combining the CND maximal inequality in Appendix~\ref{app:cnd-empirical-process} with the entropy bound in Appendix~\ref{app:entropy-attention-relu}, for every \(\delta\in(0,1)\), with probability at least \(1-\delta\),
\begin{align}\label{eq:final-nonasymptotic-bound}    \frac1n\sum_{i=1}^n    \norm{\wh W_{i\cdot}-W^\star_{i\cdot}}_2^2    &\leq    \frac{4C(1+d_{K,n})B_H}{\lambda\sqrt{n}\delta c_{\rho}}    \sqrt{    1+    V_{\mathrm{NN}}    \log\left(    \frac{eA_{\mathrm{NN}}}{B_H}    \right)    } \notag\\    &\quad+    \frac{2}{c_{\rho}}    \left[    \left(\frac{\beta_2}{\lambda}\right)^2    +    \frac{\kappa}{\lambda}    \right]\cR_{g,n}^4.\end{align}
Assume that
\[  \max\{W_1,W_2\}\leq \mathcal{W},  \qquad  \max\{L_1,L_2\}\leq \mathcal{L},  \qquad  \max\{\Lambda_1,\Lambda_2\}\leq \Lambda,\]
and denote
\[  \frac1n\sum_{i=1}^n  \E(\wh m_i-m_i^\star)^4  =  r_m^4,\]
and
\[  \frac1n\sum_{i=1}^n  \sum_{j\in\ol{\N}_1(i)}  \E(\wh e_{ij}-e_{ij}^\star)^4  =  r_e^4.\]
Then
\[  \cR_{g,n}^4  =  O_p\{k^2_{\max,n}(r_m^4+r_e^4)\}.\]
Therefore, the rate simplifies to
\begin{align}\label{eq:final-theorem-rate} \frac1n\sum_{i=1}^n  \norm{\wh W_{i\cdot}-W^\star_{i\cdot}}_2^2  =  \mathcal{O}\Bigg\{  &\frac{  d_{K,n}M_R^2\mathcal{W}\mathcal{L}^{1/2}  }{  \delta\sqrt{n(1-\rho)}  }  \sqrt{  \log\left(  \frac{  \theta B L_\ell \max\{B,(\mathcal{W}\Lambda)^{\mathcal{L}}\}  }{  M_R^2  }  \right)  } \notag\\  &\qquad+  \frac{B^2k^2_{\max,n}}{(1-\rho)^{3/2}}  (r_m^4+r_e^4)  \Bigg\}.\end{align}

\section{Empirical Process Control under Conditional Neighborhood Dependence}\label{app:cnd-empirical-process}
This appendix controls the second-stage empirical process under conditional neighborhood dependence. This is the part of the proof that yields the oracle excess-risk term in Theorem~2.

\subsection{Verification of conditional neighborhood dependence}\label{app:cnd-verification}
Following \citep{lee2019stable}, define the neighborhood system \(v_n\) by
\[  j\in v_n(i)  \quad \text{if} \quad  \ell(i,j)\leq K,\]
and
\[  v_n(\mathcal{S}_1)=\cup_{i\in \mathcal{S}_1} v_n(i).\]
Given the above two conditions, for any \(\mathcal{S}_1\) and \(\mathcal{S}_2\) such that
\[  v_n(\mathcal{S}_1)\cap v_n(\mathcal{S}_2)=\emptyset,\]
we have
\begin{align*}  &\cup_{i\in \mathcal{S}_1}  (Y_{\bar{\mathcal{N}}_1(i)}, T_{\bar{\mathcal{N}}_1(i)}, X_{\bar{\mathcal{N}}_1(i)})  \\  &\qquad\perp  \cup_{i\in \mathcal{S}_2}  (Y_{\bar{\mathcal{N}}_1(i)}, T_{\bar{\mathcal{N}}_1(i)}, X_{\bar{\mathcal{N}}_1(i)})  \mid  \cup_{i\in \mathcal{S}_1}  \bm{X}_{\bar{\mathcal{N}}_{K}(i)/\bar{\mathcal{N}}_1(i)}.\end{align*}
\begin{proof}    Denote
	\[    A =    \cup_{i\in \mathcal{S}_1}    (Y_{\bar{\mathcal{N}}_1(i)}, T_{\bar{\mathcal{N}}_1(i)}),    \]
	\[    B =    \cup_{i\in \mathcal{S}_2}    (Y_{\bar{\mathcal{N}}_1(i)}, T_{\bar{\mathcal{N}}_1(i)}, X_{\bar{\mathcal{N}}_1(i)}),    \]
	\[    C =    \cup_{i\in \mathcal{S}_1}    \bm{X}_{\bar{\mathcal{N}}_{K}(i)/\bar{\mathcal{N}}_1(i)},    \]
	\[    D =    \cup_{i\in \mathcal{S}_1}    \bm{X}_{\bar{\mathcal{N}}_1(i)},    \qquad    E =    \cup_{i\in \mathcal{S}_2}    \bm{X}_{\bar{\mathcal{N}}_{K}(i)}.    \]    Given Assumptions 1 and 2, we have
	\[      B\perp (C,D)\mid E    \]    and
	\[      E\perp (C,D),    \]    since \(\ell(i,j)\geq 2K+1\) for \(i\in\mathcal{S}_1\) and \(j\in\mathcal{S}_2\). Then
	\[      B\perp (C,D)      \qquad\text{and}\qquad      B\perp D\mid C.    \]    Also,
	\[      \mathbb{P}(AB\mid CD)      =      \mathbb{P}(A\mid CD)\mathbb{P}(B\mid CD)    \]    and
	\[      \mathbb{P}(BD\mid C)      =      \mathbb{P}(B\mid C)\mathbb{P}(D\mid C).    \]    Therefore,
	\begin{align*}        \mathbb{P}(ABD\mid C)        &=        \mathbb{P}(AB\mid CD)\mathbb{P}(D\mid C) \\        &=        \mathbb{P}(A\mid CD)\mathbb{P}(B\mid CD)\mathbb{P}(D\mid C)\\        &=        \mathbb{P}(AD\mid C)\mathbb{P}(B\mid C).    \end{align*}
\end{proof}
Then, by the definition in \citep{lee2019stable},
\[  \{T_{\bar{\mathcal{N}}_1(i)},Y_{\bar{\mathcal{N}}_1(i)}\}\]
are conditionally neighborhood dependent with respect to \((v_n,\bm{X})\).

\subsection{Centered empirical process}\label{app:centered-empirical-process}
Consider the centered empirical process
\[  \mathbb{G}_n(h)  =  \frac1{\sqrt n}  \sum_{i=1}^n  \left\{  h(D_i)-\E[h(D_i)\mid M_{v_n(i)}]  \right\},\]
where
\[  h(D_i):=\ell_i(W,g)-\ell_i(W^\star,g),\]
and \(\mathcal{M}_{v_n(i)}\) is the sigma-field generated by the CND neighborhood conditioning \(\bm{X}\). Let
\[  \cH_n  =  \left\{  h_W(D_i)=\ell_i(W,g)-\ell_i(W^\star,g):  W\in\cW_n  \right\}.\]
Then
\[  Z_n  =  \sup_{h\in\cH_n}  |\mathbb{G}_n(h)|.\]
Under Assumption 5, we have
\[  0\le \ell_i(W,g)\le \frac12 M_R^2,  \qquad  M_R=M_Y+M_m+2B.\]
Therefore, the loss class \(\cH_n\) has envelope
\[  B_H=\frac{M_R^2}{2}.\]

\subsection{CND maximal inequality}\label{app:cnd-maximal-inequality}
\begin{lemma}[CND maximal inequality]\label{lem:cnd-max}
	Given that
	\[  \{T_{\bar{\mathcal{N}}_1(i)}, Y_{\bar{\mathcal{N}}_1(i)}\}\]
	are conditionally neighborhood dependent with respect to \((v_n,\bm{X})\), Lemma 3.4 of \citep{lee2019stable} implies that there exists an absolute constant \(C>0\) such that
	\begin{equation}\label{eq:cnd-max}  \E[Z_n]  \le  C(1+d_{K,n}) \int_0^{B_H}  \sqrt{  1+  \log N_{[]}(\epsilon,\cH_n,\ol\rho_n)  }  \,d\epsilon,\end{equation}
	where
	\[  d_{K,n}:=\max_{1\leq i\leq n} |\mathcal{N}_K(i)|.\]
\end{lemma}
Next, Appendix~\ref{app:entropy-attention-relu} bounds the bracketing number
\[  N_{[]}(\epsilon,\cH_n,\ol\rho_n).\]

\section{Entropy Bound for the Attention-Based ReLU Interference Class}\label{app:entropy-attention-relu}
This appendix bounds the bracketing entropy of the second-stage loss class induced by the attention-based ReLU interference model.

\subsection{Induced loss class}\label{app:induced-loss-class}
Let the node-level covariate space be
\[    \X=[0,1]^d.\]
For a target node and its \(k-1\) neighbors, write
\[    x=(x_1,\ldots,x_k)\in\X^k,    \qquad    k\ge2.\]
The self component is modeled by a function with \(d\)-dimensional input, while the neighbor-pair component is modeled by a function with \(2d\)-dimensional input. For \(j=2,\ldots,k\), set
\[    z_j=(x_1,x_j)\in[0,1]^{2d}.\]
Let \(\Acal\subset[-\theta,\theta]\) for some \(\theta>0\). For \(f_1:[0,1]^d\to\R\), \(f_2:[0,1]^{2d}\to\R\), and \(a\in\Acal\), define
\[    G_{f_1,f_2,a}:\X^k\to\R^k\]
coordinatewise by
\begin{align}    G_{f_1,f_2,a}(x)_1 & = f_1(x_1), \notag \\    G_{f_1,f_2,a}(x)_j    &=    \frac{\exp\{\abs{a f_2(z_j)}\}}         {\sum_{\ell=2}^k \exp\{\abs{a f_2(z_\ell)}\}}    f_2(z_j),    \qquad j=2,\ldots,k .    \label{eq:G-abs-softmax}
\end{align}
Thus, the vector entering the softmax is not an additional function class. For each fixed \(x\in\X^k\), it is the vector of evaluations of the same function \(f_2\) at different pairwise inputs,
\[    q_{f_2}(x)    :=    \bigl(f_2(z_2),\ldots,f_2(z_k)\bigr)\in\R^{k-1}.\]
This point is important: a single \(L^\infty([0,1]^{2d})\) cover of \(\Ftwo\) controls all softmax coordinates simultaneously. Let \(\ell:\R^k\to\R\) be globally Lipschitz with respect to the sup norm: there exists \(L_\ell>0\) such that
\[    \abs{\ell(u)-\ell(v)}    \le    L_\ell\norm{u-v}_\infty,    \qquad    u,v\in\R^k.\]
The induced scalar loss class is
\[    \Hcal    :=    \{h_{f_1,f_2,a}=\ell\circ G_{f_1,f_2,a}:       f_1\in\Fone,\ f_2\in\Ftwo,\ a\in\Acal\}.\]
For \(\X^k\)-valued observations \(X_1,\ldots,X_n\), define the empirical \(L_2\)-type pseudometric by
\[    \bar{\rho}_n(h,h')    :=    \left\{    \frac1n\sum_{i=1}^n      \E_{X_i}\left[(h(X_i)-h'(X_i))^2\right]    \right\}^{1/2}.\]

\subsection{Bounded-envelope ReLU network classes}\label{app:bounded-relu-classes}
We use the fully connected ReLU network notation in \citep{ou2024covering}. A network configuration of depth \(L\) is a sequence
\[    \Phi=((A_i,b_i))_{i=1}^L,    \qquad    A_i\in\R^{N_i\times N_{i-1}},    \quad    b_i\in\R^{N_i},\]
with input dimension \(N_0=d\) and scalar output dimension \(N_L=1\). The realization \(R(\Phi)\) is the usual composition of affine maps and ReLU nonlinearities in the hidden layers. The width, depth, and weight magnitude are
\[    W(\Phi)=\max_{0\le i\le L}N_i,    \qquad    L(\Phi)=L,    \qquad    B_{\mathrm{wt}}(\Phi)=    \max_{1\le i\le L}    \max\{\norm{A_i}_\infty,\norm{b_i}_\infty\}.\]
For a weight-magnitude bound \(\Lambda\ge1\), define the fully connected bounded-weight ReLU realization class
\[    \Relu(p,W,L,\Lambda)    :=    \{R(\Phi):       N_0=d,\ N_L=1,       \ W(\Phi)\le W,       \ L(\Phi)\le L,       \ B_{\mathrm{wt}}(\Phi)\le\Lambda\}.\]
The deterministic envelope in the present proof is denoted by \(B>0\) and is distinct from the weight-magnitude bound \(\Lambda\). Without loss of generality, define the bounded-envelope subclass
\[    \ReluB(p,W,L,\Lambda;B)    :=    \{f\in\Relu(p,W,L,\Lambda):        \norm{f}_{L^\infty([0,1]^d)}\le B\}.\]
We take
\[    \Fone=\ReluB(d,W_1,L_1,\Lambda_1;B),    \qquad    \Ftwo=\ReluB(2d,W_2,L_2,\Lambda_2;B),\]
where
\[    W_1\ge d,    \qquad    W_2\ge 2d,    \qquad    \Lambda_1,\Lambda_2\ge1.\]
Consequently,
\[    \norm{f_1}_{L^\infty([0,1]^d)}\le B    \quad\text{for all }f_1\in\Fone,\]
and
\[    \norm{f_2}_{L^\infty([0,1]^{2d})}\le B    \quad\text{for all }f_2\in\Ftwo.\]
By \citep{ou2024covering}, there exists an absolute constant \(C>0\) such that, for every input dimension \(s\in\N\), \(W,L\in\N\), \(\Lambda\ge1\), and \(\eta\in(0,1/2)\),
\begin{equation}    \log N\left(\eta,        \Relu(s,W,L,\Lambda),        L^\infty([0,1]^s)\right)    \le    C W^2L      \log\left(        \frac{(W+1)^L\Lambda^L}{\eta}      \right).    \label{eq:OB-cover}
\end{equation}
The same upper bound applies to bounded-envelope subclasses.

\subsection{Lipschitz comparison for the absolute-score softmax map}\label{app:softmax-lipschitz}
Let \(m=k-1\). For \(q=(q_2,\ldots,q_k)\in[-B,B]^m\) and \(a\in[-\theta,\theta]\), define
\[    \pi_j^a(q)    :=    \frac{\exp\{\abs{a q_j}\}}            {\sum_{\ell=2}^k\exp\{\abs{a q_\ell}\}},    \qquad    T_a(q)_j:=\pi_j^a(q)q_j,    \qquad    j=2,\ldots,k.\]
We first record a basic Lipschitz fact for the softmax map. If
\[    \sigma_j(u)=\frac{e^{u_j}}{\sum_{\ell=2}^k e^{u_\ell}},    \qquad u\in\R^m,\]
then
\[    \norm{\sigma(u)-\sigma(v)}_\infty    \le    \frac12\norm{u-v}_\infty .\]
Indeed, for each row \(j\) of the Jacobian of \(\sigma\),
\[    \sum_{\ell=2}^k    \abs{\frac{\partial \sigma_j(u)}{\partial u_\ell}}    =    \sigma_j(u)(1-\sigma_j(u))      +      \sum_{\ell\ne j}\sigma_j(u)\sigma_\ell(u)    =    2\sigma_j(u)(1-\sigma_j(u))    \le\frac12.\]
The mean value theorem gives the stated \(\ell_\infty\to\ell_\infty\) Lipschitz bound.For fixed \(a\), using
\[    \bigl\|\abs{a q}-\abs{a r}\bigr\|_\infty    \le    \abs{a}\norm{q-r}_\infty    \le    \theta\norm{q-r}_\infty,\]
one obtains, for each coordinate \(j\),
\begin{align*}    \abs{T_a(q)_j-T_a(r)_j}    &\le    \abs{q_j-r_j}    +    B\abs{\pi_j^a(q)-\pi_j^a(r)}  \\    &\le    \left(1+\frac{\theta B}{2}\right)\norm{q-r}_\infty .\end{align*}
For fixed \(q\) and two parameters \(a,b\in[-\theta,\theta]\),
\[    \bigl\|\abs{a q}-\abs{b q}\bigr\|_\infty    \le    B\abs{a-b},\]
and therefore
\[    \abs{T_a(q)_j-T_b(q)_j}    \le    B\abs{\pi_j^a(q)-\pi_j^b(q)}    \le    \frac{B^2}{2}\abs{a-b}.\]
Combining the two estimates gives the uniform Lipschitz bound
\[    \norm{T_a(q)-T_b(r)}_\infty    \le    L_w\norm{q-r}_\infty+L_a\abs{a-b},\]
where
\[    L_w:=1+\frac{\theta B}{2},    \qquad    L_a:=\frac{B^2}{2}.\]
Now take
\[  (f_1,f_2,a),(g_1,g_2,b)\in\Fone\times\Ftwo\times\Acal.\]
Since
\[    \norm{q_{f_2}(x)-q_{g_2}(x)}_\infty    \le    \norm{f_2-g_2}_{L^\infty([0,1]^{2d})}    \qquad\text{for all }x\in\X^k,\]
we have
\begin{align*}    \norm{G_{f_1,f_2,a}(x)-G_{g_1,g_2,b}(x)}_\infty    \le    \max\Bigl\{       &\norm{f_1-g_1}_{L^\infty([0,1]^d)}, \\       &L_w\norm{f_2-g_2}_{L^\infty([0,1]^{2d})}       +       L_a\abs{a-b}    \Bigr\}.\end{align*}
By the \(L_\ell\)-Lipschitz property of \(\ell\),
\begin{align}\label{eq:H-lipschitz-bound}    \norm{h_{f_1,f_2,a}-h_{g_1,g_2,b}}_\infty    \le    L_\ell    \max\Bigl\{       &\norm{f_1-g_1}_{L^\infty([0,1]^d)}, \notag\\       &L_w\norm{f_2-g_2}_{L^\infty([0,1]^{2d})}       +       L_a\abs{a-b}    \Bigr\}.\end{align}

\subsection{Reduction from bracketing to sup-norm covering}\label{app:bracketing-covering-reduction}
For any \(h,h'\in\Hcal\), we have
\[  \bar{\rho}_n(h,h') \le \norm{h-h'}_\infty.\]
Therefore, every bracket with width at most \(\varepsilon\) in the sup norm is also a bracket with width at most \(\varepsilon\) under \(\bar{\rho}_n\). For any real-valued function class \(\mathcal G\),
\begin{equation}    N_{[]}(\varepsilon,\mathcal G,\norm{\cdot}_\infty)    \le    N(\varepsilon/2,\mathcal G,\norm{\cdot}_\infty).    \label{eq:bracket-cover-explained}
\end{equation}
Indeed, let \(g_1,\ldots,g_M\) be an \(\varepsilon/2\)-cover of \(\mathcal G\) in the sup norm. For every \(g\in\mathcal G\), there exists \(m\in\{1,\ldots,M\}\) such that
\[    \norm{g-g_m}_\infty\le\varepsilon/2.\]
Equivalently,
\[    g_m(x)-\varepsilon/2\le g(x)\le g_m(x)+\varepsilon/2    \qquad\text{for all }x.\]
Thus \(g\) is contained in the bracket
\[    [g_m-\varepsilon/2,\ g_m+\varepsilon/2].\]
The width of this bracket is exactly
\[    \norm{(g_m+\varepsilon/2)-(g_m-\varepsilon/2)}_\infty=\varepsilon.\]
The bracket endpoints need not belong to \(\mathcal G\), which is allowed in the standard definition of bracketing numbers. Since each cover center generates one valid bracket, the number of required \(\varepsilon\)-brackets is no larger than the number of \(\varepsilon/2\)-covering balls. Combining the domination \(\bar{\rho}_n\le\norm{\cdot}_\infty\) with \eqref{eq:bracket-cover-explained} yields
\begin{equation}    N_{[]}(\varepsilon,\Hcal,\bar{\rho}_n)    \le    N(\varepsilon/2,\Hcal,\norm{\cdot}_\infty).    \label{eq:reduce-H-cover}
\end{equation}

\subsection{Product-cover bracketing bound}\label{app:product-cover-bound}
Set
\[    \delta_1:=\frac{\varepsilon}{2L_\ell},    \qquad    \delta_2:=\frac{\varepsilon}{4L_\ell L_w},    \qquad    \delta_a:=\frac{\varepsilon}{4L_\ell L_a}             =             \frac{\varepsilon}{2L_\ell B^2}.\]
Assume that
\[    \norm{f_1-g_1}_{L^\infty([0,1]^d)}\le\delta_1,\]
\[    \norm{f_2-g_2}_{L^\infty([0,1]^{2d})}\le\delta_2,\]
and
\[    \abs{a-b}\le\delta_a.\]
Then the Lipschitz comparison in \eqref{eq:H-lipschitz-bound} gives
\[    \norm{h_{f_1,f_2,a}-h_{g_1,g_2,b}}_\infty    \le    L_\ell\max\{\delta_1,L_w\delta_2+L_a\delta_a\}    =    \frac{\varepsilon}{2}.\]
Consequently,
\begin{equation}
	\begin{aligned}    N(\varepsilon/2,\Hcal,\norm{\cdot}_\infty)    &\le    N\left(\delta_1,\Fone,L^\infty([0,1]^d)\right) \\    &\quad\times    N\left(\delta_2,\Ftwo,L^\infty([0,1]^{2d})\right)    N\left(\delta_a,\Acal,\abs{\cdot}\right).\end{aligned}    \label{eq:product-cover-explained}
\end{equation}
To justify \eqref{eq:product-cover-explained}, take a \(\delta_1\)-cover of \(\Fone\), a \(\delta_2\)-cover of \(\Ftwo\), and a \(\delta_a\)-cover of \(\Acal\). For each triple \((f_1,f_2,a)\), choose cover centers \((g_1,g_2,b)\) satisfying the three displayed approximation inequalities. The preceding Lipschitz comparison then shows that the induced loss \(h_{g_1,g_2,b}\) is within \(\varepsilon/2\) of \(h_{f_1,f_2,a}\) in sup norm. Hence the Cartesian product of the three component covers induces an \(\varepsilon/2\)-cover of \(\Hcal\). The cardinality of this product cover is the product of the three component covering cardinalities. If two different triples induce the same loss function, the actual cardinality only decreases.Because
\[  \Fone\subseteq\Relu(d,W_1,L_1,\Lambda_1)  \qquad\text{and}\qquad  \Ftwo\subseteq\Relu(2d,W_2,L_2,\Lambda_2),\]
we have
\begin{align*}    \log N\left(\delta_1,\Fone,L^\infty([0,1]^d)\right)    &\le    C_1 W_1^2L_1      \log\left(        \frac{(W_1+1)^{L_1}\Lambda_1^{L_1}}{\varepsilon}      \right), \\    \log N\left(\delta_2,\Ftwo,L^\infty([0,1]^{2d})\right)    &\le    C_2 W_2^2L_2      \log\left(        \frac{(W_2+1)^{L_2}\Lambda_2^{L_2}}{\varepsilon}      \right).\end{align*}
For the scalar parameter set, since \(\Acal\subset[-\theta,\theta]\),
\[    N(\delta_a,\Acal,\abs{\cdot})    \le    1+\frac{2\theta}{\delta_a}    =    1+\frac{4\theta L_\ell B^2}{\varepsilon}.\]
Combining these three estimates with \eqref{eq:reduce-H-cover} and \eqref{eq:product-cover-explained} gives the desired bound.
\begin{proposition}\label{prop:attention-relu-bracketing}
	Let
	\[    \Fone=\ReluB(d,W_1,L_1,\Lambda_1;B),    \qquad    \Ftwo=\ReluB(2d,W_2,L_2,\Lambda_2;B),\]
	where \(B>0\), \(W_1\ge d\), \(W_2\ge2d\), and \(\Lambda_1,\Lambda_2\ge1\). Let \(\Acal\subset[-\theta,\theta]\), and let \(\ell\) be \(L_\ell\)-Lipschitz with respect to \(\norm{\cdot}_\infty\) on \(\R^k\). Define \(\Hcal\) using the absolute-score softmax map \eqref{eq:G-abs-softmax}. Then, for every \(0<\varepsilon<L_\ell\),
	\begin{align}    \log N_{[]}(\varepsilon,\Hcal,\bar{\rho}_n) & \le    C_1 W_1^2L_1      \log\left(        \frac{2L_\ell (W_1+1)^{L_1}\Lambda_1^{L_1}}{\varepsilon}      \right)  \notag \\    &\quad    +    C_2 W_2^2L_2      \log\left(        \frac{4L_\ell L_w              (W_2+1)^{L_2}\Lambda_2^{L_2}}{\varepsilon}      \right) \notag\\    &\quad    +    \log\left(1+\frac{4\theta L_\ell B^2}{\varepsilon}\right).    \label{eq:final-log-bound}
	\end{align}
\end{proposition}
\begin{proof}Equation \eqref{eq:reduce-H-cover} converts \(\bar{\rho}_n\)-bracketing of \(\Hcal\) into sup-norm covering of \(\Hcal\). Equation \eqref{eq:product-cover-explained} bounds the latter by the product of the component covering numbers for \(\Fone\), \(\Ftwo\), and \(\Acal\). The \(L^\infty\) covering theorem of \citep{ou2024covering}, applied to the ambient classes \(\Relu(d,W_1,L_1,\Lambda_1)\) and \(\Relu(2d,W_2,L_2,\Lambda_2)\), yields the first two logarithmic terms in \eqref{eq:final-log-bound}; passing to bounded-envelope subclasses can only decrease the covering numbers. The elementary one-dimensional covering bound for \(\Acal\subset[-\theta,\theta]\) yields the final logarithmic term. The absolute value in the softmax exponent changes only the Lipschitz constants in Appendix~\ref{app:softmax-lipschitz}, giving
	\[  L_w=1+\theta B/2  \qquad  \text{and}  \qquad  L_a=B^2/2,\]
	and hence the scalar-parameter factor
	\[  1+4\theta L_\ell B^2/\varepsilon.\]
	This proves \eqref{eq:final-log-bound}.\end{proof}

\subsection{Entropy integral consequence}\label{app:entropy-integral}
Consequently, with
\[  V_{\mathrm{NN}}  :=  W_1^2L_1+W_2^2L_2+1\]
and
\[  A_{\mathrm{NN}}  :=  e\left[  2L_\ell(W_1+1)^{L_1}\Lambda_1^{L_1}  +  4L_\ell L_w(W_2+1)^{L_2}\Lambda_2^{L_2}  +  4\theta L_\ell B^2  \right],\]
there is a constant \(C>0\) such that
\begin{equation}\label{eq:simplified-bracket}  \log N_{[]}(\epsilon,\cH_n,\ol\rho_n)  \le  C V_{\mathrm{NN}}  \log\left(\frac{A_{\mathrm{NN}}}{\epsilon}\right).\end{equation}
For the entropy integral
\[  \mathfrak{J}_n  :=  \int_0^{B_H}  \sqrt{  1+  \log N_{[]}(\epsilon,\cH_n,\ol\rho_n)  }  \,d\epsilon,\]
we have
\[  \mathfrak{J}_n  \le  C B_H  \sqrt{  1+  V_{\mathrm{NN}}  \log\left(  \frac{eA_{\mathrm{NN}}}{B_H}  \right)  }.\]
\begin{proof}Using \eqref{eq:simplified-bracket},
	\[  \mathfrak{J}_n  \le  \int_0^{B_H}  \sqrt{  1+  C V_{\mathrm{NN}}  \log\left(\frac{A_{\mathrm{NN}}}{\epsilon}\right)  }  \,d\epsilon.\]
	For \(0<\epsilon\le B_H\),
	\[  \log\left(\frac{A_{\mathrm{NN}}}{\epsilon}\right)  \le  \log\left(\frac{eA_{\mathrm{NN}}}{B_H}\right)  +  \log\left(\frac{B_H}{e\epsilon}\right).\]
	The integral of the square-root logarithmic singularity is finite:
	\[  \int_0^{B_H}  \sqrt{1+\log(B_H/\epsilon)}\,d\epsilon  \le  C B_H .\]
	Therefore,
	\[  \mathfrak{J}_n  \le  C B_H  \sqrt{  1+  V_{\mathrm{NN}}  \log\left(  \frac{eA_{\mathrm{NN}}}{B_H}  \right)  }.\]
\end{proof}

\end{document}